\documentclass{article} 
\usepackage{Styles/PRIMEarxiv}

\usepackage{natbib}
\usepackage{hyperref}
\usepackage{url}
\usepackage{amsmath}
\usepackage{graphicx}
\usepackage{subfig}
\usepackage{svg}
\usepackage{wrapfig}
\usepackage{amssymb}
\usepackage[linesnumbered,ruled]{algorithm2e}

\title{Overcoming Slow Decision Frequencies in Continuous Control: Model-Based Sequence Reinforcement Learning for Model-Free\\Control}

\author{
 Devdhar Patel \\
  Manning College of Information \\and Computer Sciences\\
  University of Massachusetts Amherst\\
  Amherst, MA 01002 \\
  \texttt{devdharpatel@cs.umass.edu} \\
   \And
 Hava Siegelmann \\
  Manning College of Information and \\Computer Sciences\\
   University of Massachusetts Amherst\\
  Amherst, MA 01002 \\
  \texttt{hava@umass.edu} \\
}

\begin{document}

\maketitle

\begin{abstract}
Reinforcement learning (RL) is rapidly reaching and surpassing human-level control capabilities. However, state-of-the-art RL algorithms often require timesteps and reaction times significantly faster than human capabilities, which is impractical in real-world settings and typically necessitates specialized hardware. We introduce Sequence Reinforcement Learning (SRL), an RL algorithm designed to produce a sequence of actions for a given input state, enabling effective control at lower decision frequencies. SRL addresses the challenges of learning action sequences by employing both a model and an actor-critic architecture operating at different temporal scales. We propose a "temporal recall" mechanism, where the critic uses the model to estimate intermediate states between primitive actions, providing a learning signal for each individual action within the sequence. Once training is complete, the actor can generate action sequences independently of the model, achieving model-free control at a slower frequency. We evaluate SRL on a suite of continuous control tasks, demonstrating that it achieves performance comparable to state-of-the-art algorithms while significantly reducing actor sample complexity. To better assess performance across varying decision frequencies, we introduce the Frequency-Averaged Score (FAS) metric. Our results show that SRL significantly outperforms traditional RL algorithms in terms of FAS, making it particularly suitable for applications requiring variable decision frequencies. Furthermore, we compare SRL with model-based online planning, showing that SRL achieves comparable FAS while leveraging the same model during training that online planners use for planning. 
\end{abstract}
\section{Introduction}

Biological and artificial agents must learn behaviors that maximize rewards to thrive in complex environments. Reinforcement learning (RL), a class of algorithms inspired by animal behavior, facilitates this learning process \citep{sutton2018reinforcement}. The connection between neuroscience and RL is profound. The Temporal Difference (TD) error, a key concept in RL, effectively models the firing patterns of dopamine neurons in the midbrain \citep{Schultz1997, Schultz2015, Cohen2012}. Additionally, a longstanding goal of RL algorithms is to match and surpass human performance in control tasks \citep{berner2019dota, Schrittwieser2020, Kaufmann2023, Wurman2022, Vinyals2019, Mnih2015}.

However, most of these successes are achieved by leveraging large amounts of data in simulated environments and operating at speeds orders of magnitude faster than biological neurons. For example, the default timestep for the Humanoid task in the MuJoCo environment \citep{todorov2012mujoco} in OpenAI Gym \citep{towers_gymnasium_2023} is 15 milliseconds. In contrast, human reaction times range from 150 milliseconds \citep{jain2015comparative} to several seconds for complex tasks \citep{limpert2011brake}. Table \ref{table:reaction_times} shows the significant gap between AI and humans in terms of timestep and reaction times. When RL agents are constrained to human-like decision frequencies, even state-of-the-art algorithms struggle to perform in simple environments (\cite{dulac2020empirical}, Figure \ref{fig:SAC_ASL} in Appendix).

\begin{table}[ht]
\centering
\begin{tabular}{|l|c|}
\hline
\textbf{Environment / Task} & \textbf{Timestep / Reaction Time} \\ \hline
Inverted Pendulum           & 40ms                                  \\ \hline
Walker 2d                   & 8ms                                   \\ \hline
Hopper                      & 8ms                                   \\ \hline
Ant                         & 50ms                                  \\ \hline
Half Cheetah                & 50ms                                  \\ \hline
Dota 2 1v1 \citep{berner2019dota}        & 67ms                                  \\ \hline
Dota 2 5v5 \citep{berner2019dota}        & 80ms                                  \\ \hline
GT Sophy \citep{wurman2022outracing}          & 23-30ms                               \\ \hline
Drone Racing \citep{kaufmann2023champion} & 10ms                            \\ \hline
 Humans        & $\geq150
$ms                                \\ \hline
\end{tabular}
\caption{Timestep / reaction times for various benchmark environments and popular works that pit humans vs. AI.}
\label{table:reaction_times}
\end{table}

The primary reason for this difficulty is the implicit assumption in RL that the environment and the agent operate at a constant timestep. Consequently, in embodied agents that implement RL algorithms, all components: sensors, compute units, and actuators—are synchronized to the same frequency at the algorithmic level. Typically, this frequency is limited by the speed of computation in artificial agents \citep{Katz2019}. As a result, robots often require fast onboard computing hardware (CPU or GPU) to achieve higher control frequencies \citep{Margolis2024, li2022quadruped, haarnoja2023learning}. 


To allow the RL agent to observe and react to changes in the environment quickly, RL algorithms are forced to set a high frequency. Even in completely predictable environments, when the agent learns to walk or move, a small timestep is required to account for the actuation frequency required for the task, but it is not necessary to observe the environment as often or compute new actions as frequently. RL algorithms suffer from catastrophic failure due to missing inputs (also referred to as observational dropout). This behavior level gap between RL and humans can be bridged by bridging the gap in the underlying process.

 Towards that end, we propose Sequence Reinforcement Learning (SRL), a model for action sequence learning based on the role of the basal ganglia (BG) and the prefrontal cortex (PFC). Our model learns open-loop control utilizing a low decision frequency. Additionally, the algorithm utilizes a simultaneously learned model of the environment during its training but can act without it for fast and cheap inference. We demonstrate the algorithm achieves competitive performance on difficult continuous control tasks while utilizing a fraction of observations and calls to the policy. To the best of our knowledge, SRL is the first to achieve this feat. To further quantify this result and set a benchmark for control at slow frequencies, we introduce the Frequency Averaged Score (FAS) and demonstrate that SRL achieves significantly higher FAS than Soft-Actor-Critic (SAC) \citep{haarnoja2018soft} and Generative-Planning-Method (GPM) \citep{zhanggenerative}. Additionally, we demonstrate that on complex environments (with high state and action dimensions), SRL also beats model-based online planning on FAS. Finally, in the appendix, we discuss the available evidence in neuroscience that has inspired our algorithm and also present promising initial results in the proposed future work of generative replay in latent space.

\section{Necessity of Sequence Learning: Frequency, Delay and Response Time}

To perform any control task, the agent requires the following three components: Sensor, Processor/Computer, Actuator. In the traditional RL framework, all three components act at the same frequency due to the common timestep. However, this is not the case in biological agents that have different sensors of varying frequencies that are often faster than the compute frequency or the speed at which the brain can process the information \citep{borghuis2019temporal}. Additionally, in order to afford fast and precise control, the actuator frequency is also much faster than the compute frequency (see Figure \ref{fig:combined_image} in Appendix).

Low-compute hardware faces two primary challenges for real-time control: delay and throughput. The high inference times associated with low-compute devices result in a delay between receiving observations and performing corresponding actions in the environment. Additionally, they lead to low decision frequencies in sequential decision-making tasks.

While many prior works have focused on addressing delay by designing delay-aware algorithms \citep{chen2020delay, chen2021delay, derman2021acting}, mitigating delay alone does not resolve the performance issues caused by low decision frequency. Adapting RL algorithms to operate effectively in low-frequency compute settings remains an open challenge \citep{dulac2020empirical}.

The Sequence Reinforcement Learning (SRL) algorithm offers a promising solution to these low-decision frequency scenarios. To address the complete set of challenges posed by low-compute environments, SRL can be integrated with delay-aware algorithms to simultaneously manage delays while achieving higher action frequencies. Moreover, SRL inherently addresses delays by producing sequences of actions that can bridge the gap caused by processing latency. For example, if output arrives with a delay of $n$ timesteps, the first $n$ actions of the new sequence can be ignored, as they were already executed as part of the previous sequence. This mechanism ensures smooth and continuous action execution despite processing delays.


\textbf{Why low-frequency compute?}

Recent advancements in reinforcement learning (RL) algorithms, combined with high-speed computing, have led to two common approaches for addressing the speed-accuracy trade-off:

\begin{enumerate}
    
\item \textbf{Faster hardware:} The use of GPUs has become standard for enabling rapid inference in autonomous agents \citep{long2024learning, csomay2024dynamically, lazcano2024depth}. However, GPUs are often impractical in many real-world applications due to their high cost, energy demands, and large physical size. As a result, recent research has also focused on developing specialized embedded deep learning accelerators to address these challenges \citep{akkad2023embedded}.

\item \textbf{Software optimization:} Techniques such as quantization \citep{jafarpourmarzouni2024towards}, multi-exit networks \citep{rahmath2022early}, and model compression \citep{neill2020overview} are commonly employed to reduce inference times without requiring additional hardware.

\end{enumerate}

In essence, these approaches focus on either accelerating hardware or optimizing software. In this work, we propose an alternative paradigm: enhancing accuracy at low operating frequencies instead of striving for high frequencies. By advancing research in this direction, we aim to relax the dependency on high-performance hardware, enabling RL algorithms to operate effectively on low-compute devices while also making ultra-high-frequency control feasible on current hardware platforms.

\section{Related Work}

\subsection{Model-Based Reinforcement Learning}
Model-Based Reinforcement Learning (MBRL) algorithms leverage a model of the environment, which can be either learned or known, to enhance RL performance \citep{moerland2023model}. Broadly, MBRL algorithms have been utilized to:

\begin{enumerate}
    \item Improve Data Efficiency: By augmenting real-world data with model-generated data, MBRL can significantly enhance data efficiency \citep{yarats2021improving, janner2019trust, wang2021offline}.

    \item Enhance Exploration: MBRL aids in exploration by using models to identify potential or unexplored states \citep{pathak2017curiosity, stadie2015incentivizing, savinov2018episodic}.

    \item Boost Performance: Better learned representations from MBRL can lead to improved asymptotic performance \citep{silver2017mastering, levine2013guided}.
    \item Transfer Learning: MBRL supports transfer learning, enabling knowledge transfer across different tasks or environments \citep{zhang2018decoupling, sasso2022multi}.

    \item Online Planning: Models can be used for online planning with a single-step policy \citep{fickinger2021scalable}. However, this approach increases model complexity, as each online planning step necessitates an additional call to the model. This makes it unsuitable for applications with limited computational budgets and strict requirements for fast inference.
    
\end{enumerate}
Compared to online planning, our algorithm maintains a model complexity of zero after training, eliminating the need for any model calls post-training for generating a sequence of actions. This significantly reduces the computational and energy requirements, making it more suitable for practical applications in constrained environments. Additionally, model-based online planning is less biologically plausible than SRL. \citet{wiestler2013skill} demonstrated that the activations in the motor cortex reduce after skill learning, suggesting that the brain gets more efficient at performing the task after learning. In contrast, model-based online planning does not reduce in the compute and model complexity, but rather might increase in complexity as we perform longer sequences. SRL, on the other hand, has a model complexity of zero after training and thus is biologically plausible based on this observed phenomenon.

\subsection{Model Predictive Control}

Similar to model-based reinforcement learning, Model Predictive Control (MPC) utilizes a model of the system to predict and optimize future behavior. In the context of modern robotics, MPC has been effectively applied to trajectory planning and real-time control for both ground and aerial vehicles. MPC has been applied to problems like autonomous driving \citep{gray2013robust}  and bipedal control \citep{galliker2022planar}. Similar to online planning, MPC often requires access to a model of the system after training.

Additionally, similar to current RL, MPC requires very fast operational timesteps for practical applications. For example, \citet{galliker2022planar} implemented a walker at 10 ms, \citet{farshidian2017efficient} implemented a four-legged robot at 4 ms, and \citet{di2018dynamic} implemented the MIT Cheetah 3 at 33.33 ms.

\subsection{Macro-Actions, Action Repetition, and Frame-skipping}
Reinforcement Learning (RL) algorithms that utilize macro-actions demonstrate many benefits, including improved exploration and faster learning \citep{McGovern1997RolesOM}. However, identifying effective macro-actions is a challenging problem due to the curse of dimensionality, which arises from large action spaces. To address this issue, some approaches have employed genetic algorithms \citep{chang2022reusability} or relied on expert demonstrations to extract macro-actions \citep{kim2020reinforcement}. However, these methods are not scalable and lack biological plausibility. In contrast, our approach learns macro-actions using the principles of RL, thus requiring little overhead while combining the flexibility of primitive actions with the efficiency of macro-actions.

To overcome the curse of dimensionality while gaining the benefits of macro-actions, many approaches utilize frame-skipping and action repetition, where macro-actions are restricted to a single primitive action that is repeated. Frame-skipping and action repetition serve as a form of partial open-loop control, where the agent selects a sequence of actions to be executed without considering the intermediate states. Consequently, the number of actions is linear in the number of time steps \citep{Kalyanakrishnan2021AnAO, Srinivas2017DynamicAR, Biedenkapp2021TempoRLLW, Sharma2017LearningTR, yu2021taac}.

For instance, FiGaR \citep{Sharma2017LearningTR} shifts the problem of macro-action learning to predicting the number of steps that the outputted action can be repeated. TempoRL \citep{Biedenkapp2021TempoRLLW} improved upon FiGaR by conditioning the number of repetitions on the selected actions. However, none of these algorithms can scale to continuous control tasks with multiple action dimensions, as action repetition forces all actuators and joints to be synchronized in their repetitions, leading to poor performance for longer action sequences.

TLA \citep{10.1162/neco_a_01718} has recently shown an enhancement of TempoRL through the implementation of two hierarchical policies functioning at varying timesteps, coordinated by a third policy. Although TLA exhibits commendable results in environments characterized by a single action dimension, its advantages are constrained in multi-dimensional environments. This limitation arises as extended timesteps necessitate synchronization across all degrees of freedom, thereby diminishing performance during prolonged timesteps. In contrast, SRL is capable of executing distinct actions at each timestep without an increase in decision count.


\subsection{Temporally Correlated Exploration}
Recent advancements in reinforcement learning have extended the concepts of macro-actions and action-repetition to improve exploration by incorporating temporally correlated exploration, where successive actions during exploration exhibit temporal dependencies. For instance, \citet{dabney2020temporally} proposed temporally extended $\epsilon$-greedy exploration, which involves repeating actions for random durations during exploration. Building on this foundation, subsequent works have investigated approaches such as state-dependent exploration \cite{raffin2022smooth}, episodic reinforcement learning \cite{li2024open}, and temporally correlated latent noise \cite{chiappa2024latent} to enhance exploration efficiency and improve the smoothness of resulting policies. However, these methods are limited in their adaptability to challenges such as observational dropout, low decision or observational frequency, as the trained policy requires state input at each timestep.
To address long-horizon temporally correlated exploration, \citet{zhanggenerative} introduced the Generative Planning Method (GPM), which employs a recurrent actor network similar to the architecture used in SRL to generate sequences of actions from a single state. We provide an empirical comparison to GPM in Section \ref{Experiments}.



\section{Sequence Reinforcement Learning}
\begin{figure}[htbp]
    \centering
    \includegraphics[width=0.4\textwidth]{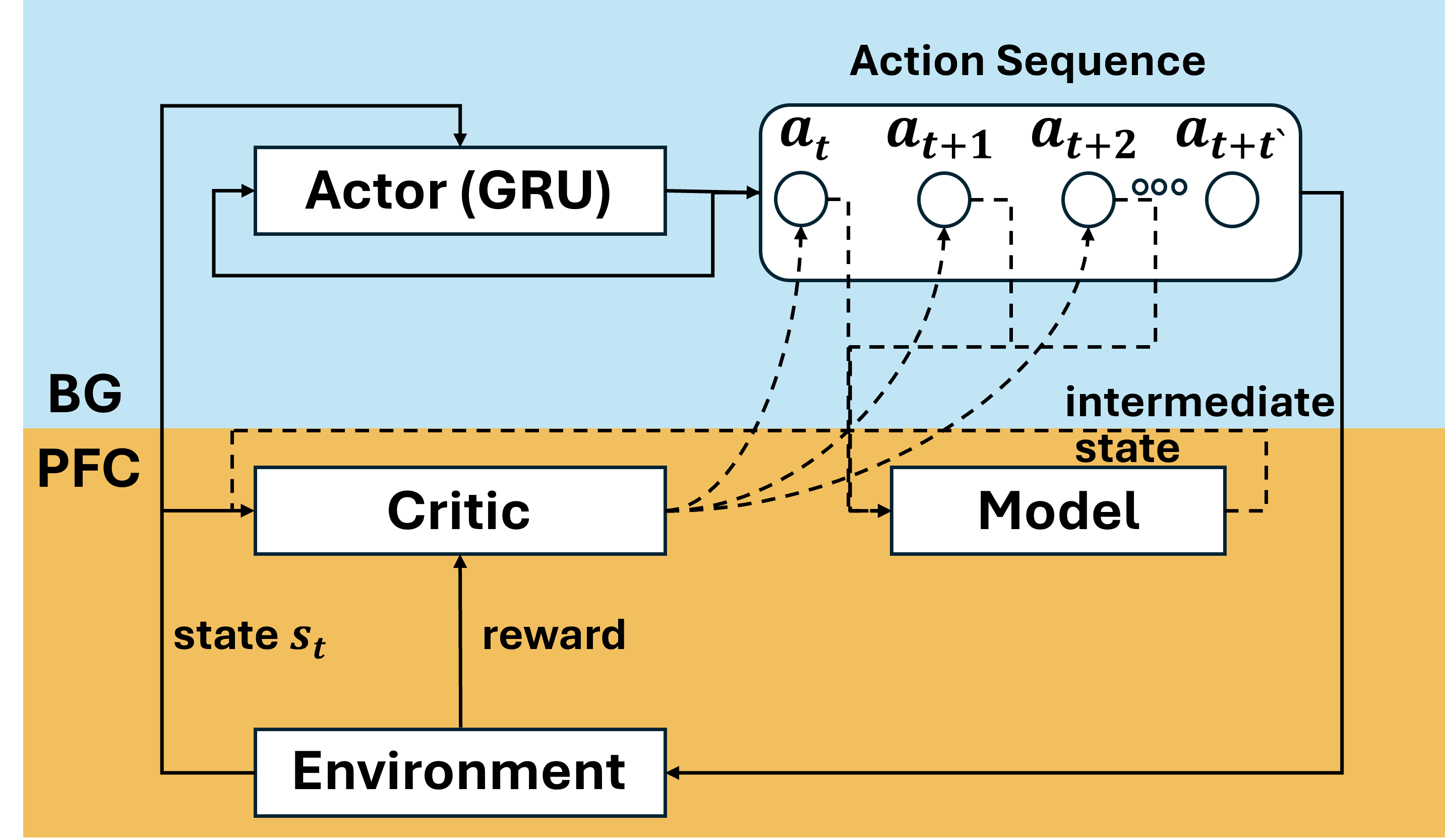}
    \caption{The Sequence Reinforcement Learning (SRL) model. The SRL takes inspiration from the function of the basal ganglia (BG) (Top/Blue) and the prefrontal cortex (PFC) (Bottom/Yellow). We train an actor with a gated  recurrent unit that can produce sequences of arbitrary lengths given a single state.  This is achieved by utilizing a critic and a model that acts at a finer temporal resolution during training/replay to provide an error signal to each primitive action of the action sequence.}
    \label{fig:combined_image}
\end{figure}

We introduce a novel reinforcement learning model capable of learning sequences of actions (macro-actions) by replaying memories at a finer temporal resolution than the action generation, utilizing a model of the environment during training. We provide the neural basis for our algorithm in the Appendix (\ref{Neural})

\subsubsection*{Components}
The Sequence Reinforcement Learning (SRL) algorithm learns to plan "in-the-mind" using a model during training, allowing the learned action-sequences to be executed without the need for model-based online planning. This is achieved using an actor-critic setting where the actor and critic operate at different frequencies, representing the observation/computation and actuation frequencies, respectively. Essentially, the critic is only used during training/replay and can operate at any temporal resolution, while the actor is constrained to the temporal resolution of the slowest component in the sensing-compute-actuation loop. Denoting the actor's timestep as $t'$ and the critic's timestep as $t$, our algorithm includes three components:
\begin{equation}
\begin{split}
    \text{Model }&: s_{t+1} = \textbf{m}_\phi(s_{t}, a_{t}) \\
    \text{Critic }&: q_{t} = \textbf{q}_\psi(s_{t}, a_{t}) \\
    \text{Actor }&: m_{t':t'+J-1 } = a_{t'}, a_{t'+t}, a_{t'+2t} .. \sim \pi_\omega(s_{t'})
\end{split}
\end{equation}
We denote individual actions in the action sequence generated by the actor using the notation $\pi_\omega(s_{t'})_t$

We denote individual actions in the action sequence $m_{t':t'+J-1 } = a_{t'}, a_{t'+t}, a_{t'+2t} .. $ generated by the actor using the notation $\pi_\omega(s_{t'})_t$ to represent the action $a_{t'+t}$.

\begin{enumerate}
    \item \textbf{Model}: Learns the dynamics of the environment, predicting the next state $s_{t+1}$ given the current state $s_t$ and primitive action $a_t$.
    \item \textbf{Critic}: Takes the same input as the model but predicts the Q-value of the state-action pair.
    \item \textbf{Actor}: Produces a sequence of actions given an observation at time $t'$. Observations from the environment can occur at any timestep $t$ or $t'$, where we assume $t' > t$. Specifically, in our algorithm, $t' = Jt$ where $J > 1; J \in \mathbb{Z}$.
\end{enumerate}

Each component of our algorithm is trained in parallel, demonstrating competitive learning speeds.

We follow the Soft-Actor-Critic (SAC) algorithm \citep{haarnoja2018soft} for learning the actor-critic. Exploration and uncertainty are critical factors heavily influenced by timestep size and planning horizon. Many model-free algorithms like DDPG \citep{lillicrap2015continuous} and TD3 \citep{fujimoto2018addressing} explore by adding random noise to each action during training. However, planning a sequence of actions over a longer timestep can result in additive noise, leading to poor performance during training and exploration if the noise parameter is not tuned properly. The SAC algorithm addresses this by automatically maximizing the entropy while also maximizing the expected return, allowing our algorithm to automatically tune its exploration based on the selected sequence length parameter ($J$).

\subsubsection*{Learning the Model}

The model is trained to minimize the Mean Squared Error of the predicted states. For a trajectory $\tau = (s_t, a_t, s_{t+1})$ drawn from the replay buffer $\mathcal{D}$, the predicted state is taken from  $\Tilde{s}_{t+1}\sim  \textbf{m}_\phi(s_{t}, a_{t})$. The loss function is:
\begin{equation}
    \mathcal{L}_{\phi} = \mathbb{E}_{\tau\sim\mathcal{D}}(\Tilde{s}_{t+1} - s_{t+1})^2
\end{equation}
For this work, the model is a feed-forward neural network with two hidden layers. In addition to the current model $\textbf{m}_\phi$, we also maintain a target model $\textbf{m}_{\phi^-}$ that is the exponential moving average of the current model.

\subsubsection*{Learning the Critic}

The critic is trained to predict the Q-value of a given state-action pair $\Tilde{q}_t = \textbf{q}_\psi(s_t, a_t)$ using the target value from the modified Bellman equation:
\begin{equation}
\hat{q}_t = r_t+\gamma \mathbb{E}_{a_{t+1}\sim\pi_\omega(s_{t+1})_0}[\textbf{q}_{\psi^-}(s_{t+1},a_{t+1}) - \alpha \log \pi_\omega(a_{t+1}|s_{t+1})]
\end{equation}
Here, $\textbf{q}_{\psi^-}$ is the target critic, which is the exponential moving average of the critic and $\alpha$ is the temperature parameter that controls the relative importance of the entropy term. Following the SAC algorithm, we train two critics and use the minimum of the two $\textbf{q}_{\psi^-}$ values to train the current critics. The loss function is:
\begin{equation}
    \mathcal{L_\psi} = \mathbb{E}_{\tau\sim\mathcal{D}}[(\Tilde{q_t}_k - \hat{q}_t)^2]\forall k \in 1, 2
\end{equation}
Both critics are feed-forward neural networks with two hidden layers. It should be noted that while the actor utilizes the model during training, the critic does not train on any data generated by the model, thus the critic training is model-free and grounded in the real environment states. 

\subsubsection*{Learning the Policy}
The SRL policy utilizes two hidden layers followed by a Gated-Recurrent-Unit (GRU) \citep{cho2014learning} that takes as input the previous action in the action sequence, followed by two linear layers that output the mean and standard deviation of the Gaussian distribution of the action. This design allows the policy to produce action sequences of arbitrary length given a single state and the last action.

A naive approach to training a sequence of actions would be to augment the action space to include all possible actions of the sequence length. However, this quickly leads to the curse of dimensionality, as each sequence is considered a unique action, dramatically increasing the policy's complexity. Additionally, such an approach ignores the temporal information of the action sequence and faces the difficult problem of credit assignment, with only a single scalar reward for the entire action sequence.

To address these problems, we use different temporal scales for the actor and critic. The critic assigns value to each primitive action of the action sequence, bypassing the credit assignment problem caused by the single scalar reward. However, using collected state-action transitions to train the action sequence is impractical, as changing the first action in the sequence would render all future states inaccurate. Thus, the model populates intermediate states, which the critic then uses to assign value to each primitive action in the sequence.

Therefore, given a trajectory $\tau=(a_{t-1},s_t,a_t,s_{t+1})$, we first produce the $J$-step action sequence using the policy: $\Tilde{m}_{t:t+J-1} \sim \pi_\omega(s_t)$. We then iteratively apply the target model to get the intermediate states $\tilde{s}_{t+1:t+J-1}$. Finally, we use the critic to calculate the loss for the actor as follows:
\begin{equation}
    \mathcal{L}_\omega =  \mathbb{E}_{\tau\sim\mathcal{D}}\bigg[\alpha \log \pi_\omega(\tilde{a}_{t}|s_{t}) - \textbf{q}_\psi(s_t,\tilde{a}_t) + \sum_{j=1}^{J-1} \alpha \log \pi_\omega(\tilde{a}_{t+j}|\tilde{s}_{t+j}) - \textbf{q}_\psi(\tilde{s}_{t+j},\tilde{a}_{t+j})\bigg]
\end{equation}

\section{Experiments}\label{Experiments}
\subsection*{Overview}
We evaluate our SRL approach on 11 continuous control tasks, comparing it against SAC \citep{haarnoja2018soft} and GPM \citep{zhanggenerative}. We utilize the OpenAI Gym \citep{1606.01540} implementation of the MuJoCo environments \citep{todorov2012mujoco}.

\subsection*{Experimental Setup}

We train SRL with four different action sequence lengths (ASL), $J = {2, 4, 8, 16}$, referred to as SRL-$J$. During training, SRL is evaluated based on its $J$ value, processing states only after every $J$ actions. All hyperparameters are identical between SRL and SAC, except for the actor update frequency: SRL updates the actor every 4 steps, while SAC updates every step. Thus, SAC has four more actor update steps compared to SRL. Additionally, SRL learns a model in parallel with the actor and critic. Additionally, we also train SAC at different step sizes that correspond to SRL, forming SAC-$J$ where $J={1, 2, 4, 8, 16}$. Note that we do not provide SRL-1 since for sequences of length 1, SRL is the same algorithm as SAC.

We present the learning curves of SRL and SAC across 11 continuous control tasks in the appendix. We find that on all environments except Swimmer, SAC-1 demonstrates optimal performance and often significantly outperforms the longer timesteps. Thus, the default environments are picked to maximize performance under the standard RL setting where the observation, decision, and the action frequency are the same. It should be noted that the learning curves presented for SRL-$J$ and SAC-$J$ take in states every $J$ steps.

\subsection*{Frequency-Averaged Score}


Transitioning from simulation to real-world implementation (Sim2Real) in control systems is challenging because deployment introduces computational stochasticity, leading to variable sensor sampling rates (throughput) and inconsistent end-to-end delays from sensing to actuation \citep{sandha2021sim2real}. This gap is not captured by the mean reward or return that is the norm in current RL literature. To address this, we introduce Frequency-Averaged Score (FAS) that is the normalized area under the curve (AUC) of the performance vs. decision frequency plot.
We provide plots for all environments in the Appendix. We note that this experimental setup is similar to the challenge 7 introduced in by \citet{dulac2020empirical} and SRL addresses the challenge of low throughput that is introduced in that work. The FAS captures the overall performance of the policy at different decision frequencies, timesteps or macro-action lengths. A High FAS indicates that the policy performance generalizes across decision frequencies, observation frequencies and timestep sizes. 

\begin{table}[ht]
\centering
\begin{tabular}{|l|c|c|c|c|c|}
\hline
\textbf{Environment} & \textbf{SAC-1} & \textbf{SAC-2} & \textbf{SAC-4} & \textbf{SAC-8} & \textbf{SAC-16} \\ \hline
Pendulum                & 0.44 $\pm$ 0.03 & 0.42 $\pm$ 0.03 & \textbf{0.50} $\pm$ 0.03 & 0.49 $\pm$ 0.04 & 0.33 $\pm$ 0.05 \\ \hline
Lunar Lander   & 0.18 $\pm$ 0.23 & 0.23 $\pm$ 0.02 & 0.33 $\pm$ 0.02 & 0.45 $\pm$ 0.03 & \textbf{0.56} $\pm$ 0.09 \\ \hline
Hopper                  & 0.05 $\pm$ 0.03 & 0.09 $\pm$ 0.01 & 0.14 $\pm$ 0.03 & 0.14 $\pm$ 0.04 & \textbf{0.26} $\pm$ 0.08 \\ \hline
Walker2d                & 0.07 $\pm$ 0.01 & 0.08 $\pm$ 0.03 & 0.14 $\pm$ 0.04 & \textbf{0.23} $\pm$ 0.07 & 0.15 $\pm$ 0.04 \\ \hline
Ant                     & -0.05 $\pm$ 0.04 & 0.11 $\pm$ 0.01 & 0.16 $\pm$ 0.02 & \textbf{0.16} $\pm$ 0.01 & 0.13 $\pm$ 0.01 \\ \hline
HalfCheetah             & 0.01 $\pm$ 0.01 & \textbf{0.04} $\pm$ 0.01 & 0.03 $\pm$ 0.00 & 0.02 $\pm$ 0.01 & 0.01 $\pm$ 0.01 \\ \hline
Humanoid                & 0.06 $\pm$ 0.01 & 0.06 $\pm$ 0.01 & 0.08 $\pm$ 0.03 & 0.17 $\pm$ 0.02 & \textbf{0.18} $\pm$ 0.04 \\ \hline
InvertedPendulum        & 0.05 $\pm$ 0.02 & 0.07 $\pm$ 0.00 & 0.14 $\pm$ 0.00 & 0.31 $\pm$ 0.02 & \textbf{0.34} $\pm$ 0.20 \\ \hline
InvertedDPendulum  & 0.02 $\pm$ 0.00 & 0.07 $\pm$ 0.00 & \textbf{0.09} $\pm$ 0.01 & 0.01 $\pm$ 0.00 & 0.01 $\pm$ 0.00 \\ \hline
Reacher                 & 0.72 $\pm$ 0.04 & 0.78 $\pm$ 0.01 & 0.84 $\pm$ 0.03 & 0.86 $\pm$ 0.02 & \textbf{0.87} $\pm$ 0.02 \\ \hline
Swimmer                 & 0.08 $\pm$ 0.02 & 0.28 $\pm$ 0.04 & 0.46 $\pm$ 0.03 & 0.53 $\pm$ 0.03 & \textbf{0.54} $\pm$ 0.06 \\ \hline
\end{tabular}
\caption{Mean Frequency-Averaged Score (FAS) and standard deviation for different environments for SAC-$J$ configurations ($J=1,2,4,8,16$. $J$ is the action sequence length during training). Each value is averaged over 5 trials (rounded to two decimals, highest value highlighted).}
\label{Table:FAS_SAC}
\end{table}

\begin{table}[ht]
\centering
\begin{tabular}{|l|c|c|c|c|}
\hline
\textbf{Environment} & \textbf{SRL-2} & \textbf{SRL-4} & \textbf{SRL-8} & \textbf{SRL-16} \\ \hline
Pendulum             & 0.49 $\pm$ 0.04 & 0.68 $\pm$ 0.02 & 0.78 $\pm$ 0.04 & \textbf{0.88} $\pm$ 0.02 \\ \hline
Lunar Lander         & 0.14 $\pm$ 0.06 & 0.52 $\pm$ 0.03 & 0.73 $\pm$ 0.04 & \textbf{0.84} $\pm$ 0.03 \\ \hline
Hopper               & 0.10 $\pm$ 0.02 & 0.23 $\pm$ 0.03 & 0.42 $\pm$ 0.04 & \textbf{0.57} $\pm$ 0.02 \\ \hline
Walker2d             & 0.12 $\pm$ 0.03 & 0.25 $\pm$ 0.06 & \textbf{0.28} $\pm$ 0.06 & 0.24 $\pm$ 0.11 \\ \hline
Ant                  & 0.04 $\pm$ 0.01 & 0.29 $\pm$ 0.09 & 0.45 $\pm$ 0.14 & \textbf{0.54} $\pm$ 0.13 \\ \hline
HalfCheetah          & 0.06 $\pm$ 0.01 & 0.13 $\pm$ 0.02 & 0.22 $\pm$ 0.01 & \textbf{0.28} $\pm$ 0.01 \\ \hline
Humanoid             & 0.07 $\pm$ 0.00 & 0.18 $\pm$ 0.02 & 0.37 $\pm$ 0.04 & \textbf{0.46} $\pm$ 0.04 \\ \hline
InvertedPendulum          & 0.09 $\pm$ 0.03 & 0.16 $\pm$ 0.03 & 0.27 $\pm$ 0.02 & \textbf{0.44} $\pm$ 0.04 \\ \hline
InvertedDPendulum         & 0.07 $\pm$ 0.00 & \textbf{0.13} $\pm$ 0.02 & 0.03 $\pm$ 0.02 & 0.02 $\pm$ 0.00 \\ \hline
Reacher              & 0.90 $\pm$ 0.01 & 0.93 $\pm$ 0.00 & 0.95 $\pm$ 0.00 & \textbf{0.96} $\pm$ 0.00 \\ \hline
Swimmer              & 0.32 $\pm$ 0.05 & 0.38 $\pm$ 0.17 & 0.31 $\pm$ 0.02 & \textbf{0.42} $\pm$ 0.15 \\ \hline
\end{tabular}
\caption{Mean Frequency-Averaged Score (FAS) and standard deviation for different environments for SRL-$J$ configurations ($J=2,4,8,16$. $J$ is the action sequence length during training). Each value is averaged over 5 trials (rounded to two decimals, highest value highlighted).}
\label{Table:FAS}
\end{table}

Tables \ref{Table:FAS_SAC} and \ref{Table:FAS} present the Frequency Averaged Score (FAS) for SAC and SRL across varying action sequence lengths. Overall, SRL-16 demonstrates strong and consistent performance across most environments and a wide range of frequencies. However, in the Walker2d-v2 and InvertedDoublePendulum-v2 environments, SRL faces challenges when learning longer action sequences. We hypothesize that these difficulties stem from higher modeling errors in these environments. Future work aimed at improving environmental models could potentially address these issues.

SAC, in contrast, performed poorly across all environments, highlighting the limitations of traditional RL methods in adapting to changes in frequency. Although training SAC with larger timesteps ($J$) improves FAS, this approach compromises performance at shorter timesteps, ultimately reducing the overall score (see Appendix Fig. \ref{fig:SAC_ASL}).

An exception to this trend is the Swimmer environment, where SAC benefits from improved exploration due to extended actions. SRL, which does not use action repetition, does not perform as well in this specific case. However, this limitation could be addressed by incorporating action repetition or action correlation during exploration—an enhancement that lies beyond the scope of the current work. 

In order to further validate the utility of FAS, we test all the policies (SAC and SRL-$J$) in a stochastic timestep environment. The timestep (time until next input) is randomly chosen from a uniform distribution of integers in [1,16] after each decision. This is a more realistic setting as it tests the performance of the policy when the frequency is not constant. Each policy is evaluated over 10 episodes with stochastic timesteps. 


In all tested environments, except for the Inverted Double Pendulum, there is a strong Pearson correlation coefficient (greater than or equal to 0.82) between FAS and performance in stochastic conditions. This high correlation confirms the effectiveness of FAS as a metric for measuring a policy's generalized performance across various timesteps and frequencies. The Inverted Double Pendulum, however, presents a unique challenge due to its requirement for high precision at low decision frequencies, leading to significantly lower FAS scores for all algorithms and thus it is an outlier. Comprehensive plots for all nine environments are included in the appendix (Fig. \ref{fig:Stochastic}).

\subsection*{Comparison to Generative Planning Method}

The Generative Planning Method (GPM) \citep{zhanggenerative} uses a recurrent actor, like SRL, to generate actions for improved exploration. Originally designed for a different context and evaluated in the standard RL setting, GPM optimizes plan actions to maximize Q-value, potentially exceeding SAC in FAS score. We compare SRL and GPM in four environments to test this.

In the original work, GPM was trained with a plan length of 3, similar to the $J$ parameter in our study. Though shorter plans may restrict generalization to longer sequences, GPM is robust to plan length variations. For fair comparison, we use the best-performing $J$ values for SRL in each environment.

\begin{figure}
    \centering
    \includegraphics[width=0.22\linewidth]{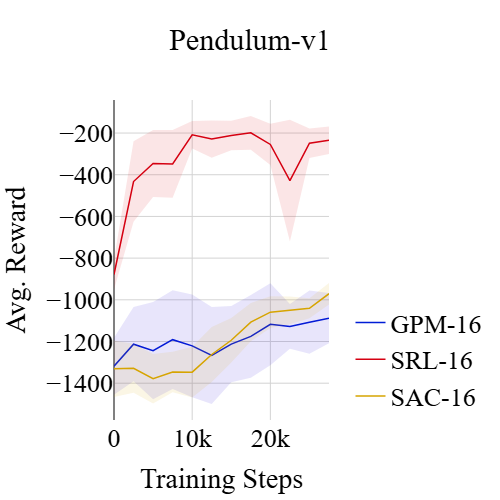}
    \includegraphics[width=0.22\linewidth]{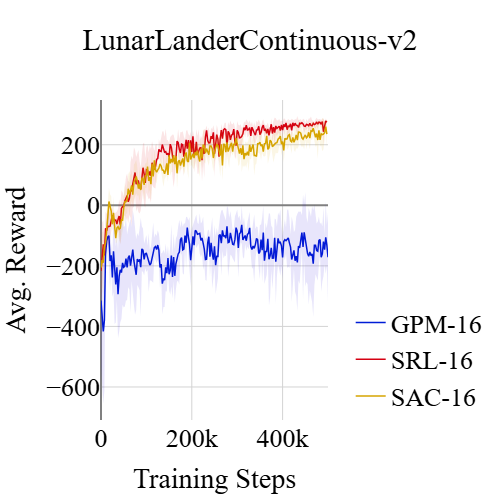}
    \includegraphics[width=0.22\linewidth]{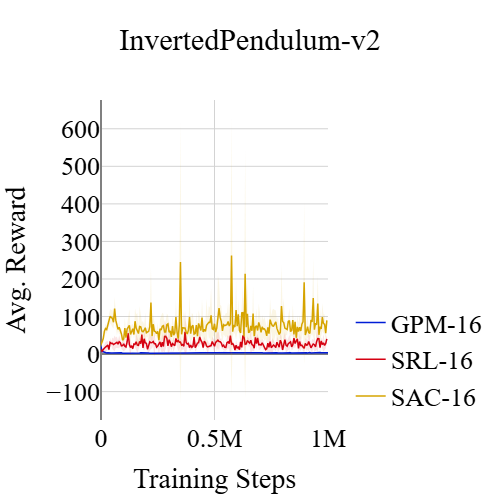}
    \includegraphics[width=0.22\linewidth]{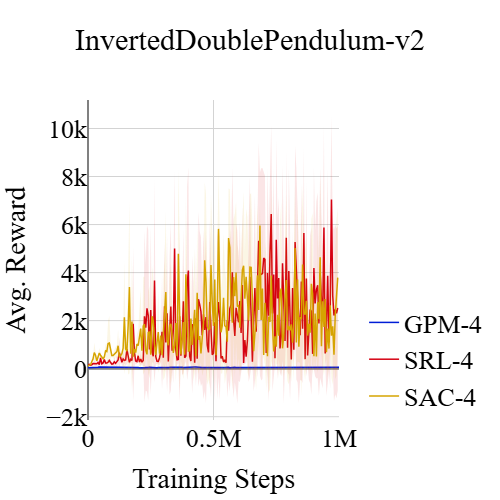}

    \includegraphics[width=0.22\linewidth]{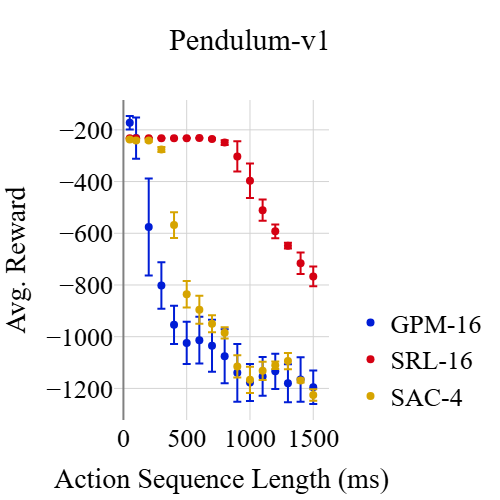}
    \includegraphics[width=0.22\linewidth]{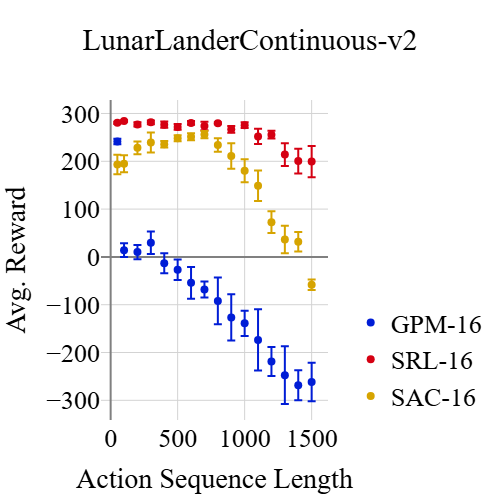}
    \includegraphics[width=0.22\linewidth]{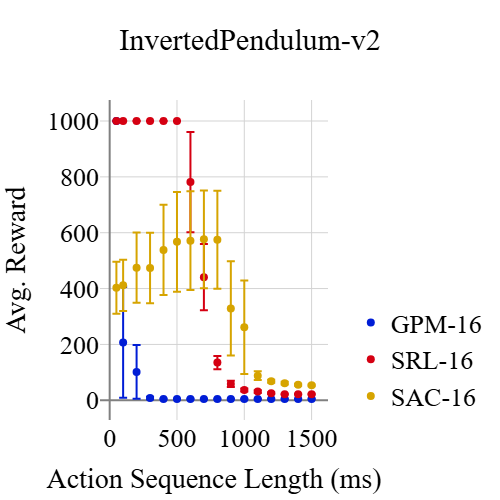}
    \includegraphics[width=0.22\linewidth]{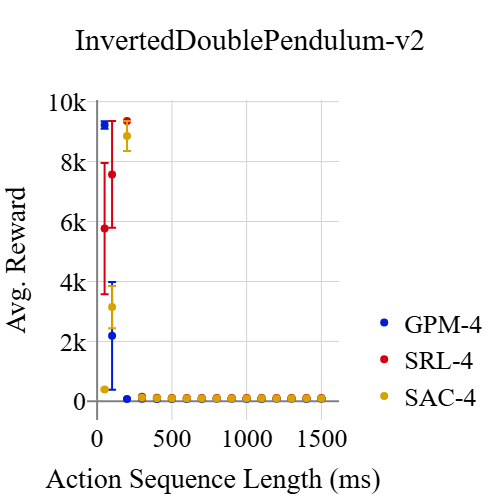}
    \caption{Comparison of SAC and SRL to GPM. Top: Learning curves. Bottom: Performance of the trained policies at different action sequence lengths. The action sequences for SRL and GPM are generated using the recurrent actor while SAC utilizes action repetition. GPM achieves FAS of 0.41, 0.04, 0.04, 0.04 on the environments from left to right respectively. }
    \label{fig:GPM}
\end{figure}
Figure \ref{fig:GPM} shows the learning curves and FAS evaluation plots for GPM compared to SAC and SRL. While GPM generates a plan by optimizing a sequence of actions, it achieves optimal performance only at sequence lengths of one. As a result, its FAS score is even lower than that of SAC-$J$.

Notably, on the InvertedDoublePendulum-v2 environment, both SAC and SRL exhibit high performance at action sequence lengths (ASL) of 4, which aligns with their training at $J=4$. However, their performance decreases at shorter ASLs. In contrast, GPM shows a similar FAS profile to SAC-1, indicating that its performance does not generalize well to longer action sequences.



\subsection*{Comparison to Model-based Online Planning}
Model-based online planning is another approach that allows the RL agent to reduce its observational frequency. However, it often requires a highly accurate model of the environment and incurs increased model complexity due to the use of the model during control.

Since SRL incorporates a model of the environment that is learned in parallel, we compare the performance of the SRL actor utilizing the actor-generated action sequences against model-based online planning, where the actor produces only a single action between each simulated state.

\begin{table}[ht]
\centering
\begin{tabular}{|l|c|c|c|c|}
\hline
\textbf{Environment}       & \textbf{SRL} & \textbf{Online Planning} & \textbf{State Space} & \textbf{Action Space} \\ \hline
Lunar Lander               & \textbf{0.84 $\pm$ 0.03}            & 0.79 $\pm$ 0.08  & 8 & 2                         \\ \hline
Hopper                     & 0.57 $\pm$ 0.02            & \textbf{0.59 $\pm$ 0.19}  &  11  & 3                     \\ \hline
Walker2d                   & \textbf{0.28 $\pm$ 0.06}            & 0.20 $\pm$ 0.05 &  17 & 6                        \\ \hline
Ant                        & \textbf{0.54 $\pm$ 0.13}            & 0.34 $\pm$ 0.08  & 27 & 8                        \\ \hline
HalfCheetah                & \textbf{0.28 $\pm$ 0.01}            & 0.19 $\pm$ 0.02  & 17 & 6                        \\ \hline
Humanoid                   & \textbf{0.46 $\pm$ 0.04}            & 0.18 $\pm$ 0.03  & 376 & 17                        \\ \hline
InvPendulum                & 0.44 $\pm$ 0.04            & \textbf{0.63 $\pm$ 0.10}  & 4 & 1                    \\ \hline
InvDPendulum               & \textbf{0.13 $\pm$ 0.02}            & 0.10 $\pm$ 0.07 & 11 &  1                      \\ \hline
Reacher                    & \textbf{0.96 $\pm$ 0.00 }           & 0.95 $\pm$ 0.00 & 11 & 2                         \\ \hline
Swimmer                    & 0.42 $\pm$ 0.15            & \textbf{0.43 $\pm$ 0.14} & 8 &  2                       \\ \hline
\end{tabular}
\caption{Comparison of the FAS of SRL and corresponding model-based online planning policies across different environments.}
\label{table:Model}
\end{table}


Table \ref{table:Model} compares the FAS score SRL to online planning using the same model in online planning versus the action sequences generated by the SRL policy. We see that SRL can learn action sequences and is competitive to model-based online planning. Notably, SRL performs better in environments with larger action and state space dimensions. Such environments are harder to model. Thus, SRL can leverage inaccurate models to learn accurate action sequences, further reducing the required computational complexity during training. We hypothesize that this superior performance is due to the fact that the actor learns a $J$-step action sequence concurrently, while online planning only produces one action at a time. Consequently, SRL is able to learn and produce long, coherent action sequences, whereas single-step predictions tend to drift, similar to the 'hallucination' phenomenon observed in transformer-based language models.

\section{Discussion and Future Work} 

SRL bridges the gap between RL and real-world applications by enabling robust control at low decision frequencies. Its ability to learn long action sequences expands the potential for deploying RL in resource-constrained environments, such as robotics and autonomous systems. Additionally, it shows promise for applications where obtaining observations is costly, such as in medical diagnostics and treatment planning. Future work will explore hierarchical policies and biologically inspired attention mechanisms.

The current RL framework encourages synchrony between the environment and the components of the agent. However, the brain utilizes components that act at different frequencies and yet is capable of robust and accurate control. SRL provides an approach to reconcile this difference between neuroscience and RL, while remaining competitive on current RL benchmarks. SRL offers substantial benefits over traditional RL algorithms, particularly in the context of autonomous agents in constrained settings. By enabling operation at slower observational frequencies and providing a gradual decay in performance with reduced input frequency, SRL addresses critical issues related to sensor failure and occlusion, and energy consumption. Additionally, SRL generates long sequences of actions from a single state, which can enhance the explainability of the policy and provide opportunities to override the policy early in case of safety concerns. SRL also learns a latent representation of the action sequence, which could be used in the future to interface with large language models for multimodal explainability and even hierarchical reinforcement learning and transfer learning.

\section{Conclusion}

In this paper, we introduced Sequence Reinforcement Learning (SRL): a model-based action sequence learning algorithm for model-free control. We demonstrated the improvement of SRL over the existing framework by testing it over various control frequencies. Furthermore, we introduce the Frequency-Averaged-Score (FAS) metric to measure the robustness of a policy across different frequencies. Our work is the first to achieve competitive results on continuous control environments at low control frequencies and serves as a benchmark for future work in this direction. Finally, we demonstrated directions for future work, including comparison to model-based planning, generative replay, and connections to neuroscience.


\subsubsection*{Acknowledgments}
We would like to thank Dr. Terrence Sejnowski for his valuable discussions, insightful feedback, and guidance throughout this work. His expertise and support have been instrumental in refining the ideas presented in this paper.

\newpage\bibliographystyle{unsrtnat}

\bibliography{main}

\begin{thebibliography}{90}
\providecommand{\natexlab}[1]{#1}
\providecommand{\url}[1]{\texttt{#1}}
\expandafter\ifx\csname urlstyle\endcsname\relax
  \providecommand{\doi}[1]{doi: #1}\else
  \providecommand{\doi}{doi: \begingroup \urlstyle{rm}\Url}\fi

\bibitem[Sutton and Barto(2018)]{sutton2018reinforcement}
Richard~S Sutton and Andrew~G Barto.
\newblock \emph{Reinforcement learning: An introduction}.
\newblock MIT press, 2018.

\bibitem[Schultz et~al.(1997)Schultz, Dayan, and Montague]{Schultz1997}
W.~Schultz, P.~Dayan, and P.~R. Montague.
\newblock A neural substrate of prediction and reward.
\newblock \emph{Science}, 275:\penalty0 1593--1599, 1997.
\newblock ISSN 00368075.
\newblock \doi{10.1126/SCIENCE.275.5306.1593/ASSET/6CC77AD2-EA4B-4861-A4ED-A076123F94E0/ASSETS/GRAPHIC/SE1174905004.JPEG}.
\newblock URL \url{https://www.science.org/doi/10.1126/science.275.5306.1593}.

\bibitem[Schultz(2015)]{Schultz2015}
Wolfram Schultz.
\newblock Neuronal reward and decision signals: From theories to data.
\newblock \emph{Physiological Reviews}, 95:\penalty0 853--951, 7 2015.
\newblock ISSN 15221210.
\newblock \doi{10.1152/PHYSREV.00023.2014/ASSET/IMAGES/LARGE/Z9J0031527320049.JPEG}.
\newblock URL \url{https://journals.physiology.org/doi/10.1152/physrev.00023.2014}.

\bibitem[Cohen et~al.(2012)Cohen, Haesler, Vong, Lowell, and Uchida]{Cohen2012}
Jeremiah~Y. Cohen, Sebastian Haesler, Linh Vong, Bradford~B. Lowell, and Naoshige Uchida.
\newblock Neuron-type-specific signals for reward and punishment in the ventral tegmental area.
\newblock \emph{Nature 2012 482:7383}, 482:\penalty0 85--88, 1 2012.
\newblock ISSN 1476-4687.
\newblock \doi{10.1038/nature10754}.
\newblock URL \url{https://www.nature.com/articles/nature10754}.

\bibitem[OpenAI et~al.(2019)OpenAI, :, Berner, Brockman, Chan, Cheung, Debiak, Dennison, Farhi, Fischer, Hashme, Hesse, Józefowicz, Gray, Olsson, Pachocki, Petrov, d.~O.~Pinto, Raiman, Salimans, Schlatter, Schneider, Sidor, Sutskever, Tang, Wolski, and Zhang]{berner2019dota}
OpenAI, :, Christopher Berner, Greg Brockman, Brooke Chan, Vicki Cheung, Przemyslaw Debiak, Christy Dennison, David Farhi, Quirin Fischer, Shariq Hashme, Chris Hesse, Rafal Józefowicz, Scott Gray, Catherine Olsson, Jakub Pachocki, Michael Petrov, Henrique~P. d.~O.~Pinto, Jonathan Raiman, Tim Salimans, Jeremy Schlatter, Jonas Schneider, Szymon Sidor, Ilya Sutskever, Jie Tang, Filip Wolski, and Susan Zhang.
\newblock Dota 2 with large scale deep reinforcement learning, 2019.
\newblock URL \url{https://arxiv.org/abs/1912.06680}.

\bibitem[Schrittwieser et~al.(2020)Schrittwieser, Antonoglou, Hubert, Simonyan, Sifre, Schmitt, Guez, Lockhart, Hassabis, Graepel, Lillicrap, and Silver]{Schrittwieser2020}
Julian Schrittwieser, Ioannis Antonoglou, Thomas Hubert, Karen Simonyan, Laurent Sifre, Simon Schmitt, Arthur Guez, Edward Lockhart, Demis Hassabis, Thore Graepel, Timothy Lillicrap, and David Silver.
\newblock Mastering atari, go, chess and shogi by planning with a learned model.
\newblock \emph{Nature 2020 588:7839}, 588:\penalty0 604--609, 12 2020.
\newblock ISSN 1476-4687.
\newblock \doi{10.1038/s41586-020-03051-4}.
\newblock URL \url{https://www.nature.com/articles/s41586-020-03051-4}.

\bibitem[Kaufmann et~al.(2023{\natexlab{a}})Kaufmann, Bauersfeld, Loquercio, Müller, Koltun, and Scaramuzza]{Kaufmann2023}
Elia Kaufmann, Leonard Bauersfeld, Antonio Loquercio, Matthias Müller, Vladlen Koltun, and Davide Scaramuzza.
\newblock Champion-level drone racing using deep reinforcement learning.
\newblock \emph{Nature 2023 620:7976}, 620:\penalty0 982--987, 8 2023{\natexlab{a}}.
\newblock ISSN 1476-4687.
\newblock \doi{10.1038/s41586-023-06419-4}.
\newblock URL \url{https://www.nature.com/articles/s41586-023-06419-4}.

\bibitem[Wurman et~al.(2022{\natexlab{a}})Wurman, Barrett, Kawamoto, MacGlashan, Subramanian, Walsh, Capobianco, Devlic, Eckert, Fuchs, Gilpin, Khandelwal, Kompella, Lin, MacAlpine, Oller, Seno, Sherstan, Thomure, Aghabozorgi, Barrett, Douglas, Whitehead, Dürr, Stone, Spranger, and Kitano]{Wurman2022}
Peter~R. Wurman, Samuel Barrett, Kenta Kawamoto, James MacGlashan, Kaushik Subramanian, Thomas~J. Walsh, Roberto Capobianco, Alisa Devlic, Franziska Eckert, Florian Fuchs, Leilani Gilpin, Piyush Khandelwal, Varun Kompella, Hao~Chih Lin, Patrick MacAlpine, Declan Oller, Takuma Seno, Craig Sherstan, Michael~D. Thomure, Houmehr Aghabozorgi, Leon Barrett, Rory Douglas, Dion Whitehead, Peter Dürr, Peter Stone, Michael Spranger, and Hiroaki Kitano.
\newblock Outracing champion gran turismo drivers with deep reinforcement learning.
\newblock \emph{Nature 2022 602:7896}, 602:\penalty0 223--228, 2 2022{\natexlab{a}}.
\newblock ISSN 1476-4687.
\newblock \doi{10.1038/s41586-021-04357-7}.
\newblock URL \url{https://www.nature.com/articles/s41586-021-04357-7}.

\bibitem[Vinyals et~al.(2019)Vinyals, Babuschkin, Czarnecki, Mathieu, Dudzik, Chung, Choi, Powell, Ewalds, Georgiev, Oh, Horgan, Kroiss, Danihelka, Huang, Sifre, Cai, Agapiou, Jaderberg, Vezhnevets, Leblond, Pohlen, Dalibard, Budden, Sulsky, Molloy, Paine, Gulcehre, Wang, Pfaff, Wu, Ring, Yogatama, Wünsch, McKinney, Smith, Schaul, Lillicrap, Kavukcuoglu, Hassabis, Apps, and Silver]{Vinyals2019}
Oriol Vinyals, Igor Babuschkin, Wojciech~M. Czarnecki, Michaël Mathieu, Andrew Dudzik, Junyoung Chung, David~H. Choi, Richard Powell, Timo Ewalds, Petko Georgiev, Junhyuk Oh, Dan Horgan, Manuel Kroiss, Ivo Danihelka, Aja Huang, Laurent Sifre, Trevor Cai, John~P. Agapiou, Max Jaderberg, Alexander~S. Vezhnevets, Rémi Leblond, Tobias Pohlen, Valentin Dalibard, David Budden, Yury Sulsky, James Molloy, Tom~L. Paine, Caglar Gulcehre, Ziyu Wang, Tobias Pfaff, Yuhuai Wu, Roman Ring, Dani Yogatama, Dario Wünsch, Katrina McKinney, Oliver Smith, Tom Schaul, Timothy Lillicrap, Koray Kavukcuoglu, Demis Hassabis, Chris Apps, and David Silver.
\newblock Grandmaster level in starcraft ii using multi-agent reinforcement learning.
\newblock \emph{Nature 2019 575:7782}, 575:\penalty0 350--354, 10 2019.
\newblock ISSN 1476-4687.
\newblock \doi{10.1038/s41586-019-1724-z}.
\newblock URL \url{https://www.nature.com/articles/s41586-019-1724-z}.

\bibitem[Mnih et~al.(2015)Mnih, Kavukcuoglu, Silver, Rusu, Veness, Bellemare, Graves, Riedmiller, Fidjeland, Ostrovski, Petersen, Beattie, Sadik, Antonoglou, King, Kumaran, Wierstra, Legg, and Hassabis]{Mnih2015}
Volodymyr Mnih, Koray Kavukcuoglu, David Silver, Andrei~A. Rusu, Joel Veness, Marc~G. Bellemare, Alex Graves, Martin Riedmiller, Andreas~K. Fidjeland, Georg Ostrovski, Stig Petersen, Charles Beattie, Amir Sadik, Ioannis Antonoglou, Helen King, Dharshan Kumaran, Daan Wierstra, Shane Legg, and Demis Hassabis.
\newblock Human-level control through deep reinforcement learning.
\newblock \emph{Nature 2015 518:7540}, 518:\penalty0 529--533, 2 2015.
\newblock ISSN 1476-4687.
\newblock \doi{10.1038/nature14236}.
\newblock URL \url{https://www.nature.com/articles/nature14236}.

\bibitem[Todorov et~al.(2012)Todorov, Erez, and Tassa]{todorov2012mujoco}
Emanuel Todorov, Tom Erez, and Yuval Tassa.
\newblock Mujoco: A physics engine for model-based control.
\newblock In \emph{2012 IEEE/RSJ International Conference on Intelligent Robots and Systems}, pages 5026--5033. IEEE, 2012.
\newblock \doi{10.1109/IROS.2012.6386109}.

\bibitem[Towers et~al.(2023)Towers, Terry, Kwiatkowski, Balis, Cola, Deleu, Goulão, Kallinteris, KG, Krimmel, Perez-Vicente, Pierré, Schulhoff, Tai, Shen, and Younis]{towers_gymnasium_2023}
Mark Towers, Jordan~K. Terry, Ariel Kwiatkowski, John~U. Balis, Gianluca~de Cola, Tristan Deleu, Manuel Goulão, Andreas Kallinteris, Arjun KG, Markus Krimmel, Rodrigo Perez-Vicente, Andrea Pierré, Sander Schulhoff, Jun~Jet Tai, Andrew Tan~Jin Shen, and Omar~G. Younis.
\newblock Gymnasium, March 2023.
\newblock URL \url{https://zenodo.org/record/8127025}.

\bibitem[Jain et~al.(2015)Jain, Bansal, Kumar, and Singh]{jain2015comparative}
Aditya Jain, Ramta Bansal, Avnish Kumar, and KD~Singh.
\newblock A comparative study of visual and auditory reaction times on the basis of gender and physical activity levels of medical first year students.
\newblock \emph{International journal of applied and basic medical research}, 5\penalty0 (2):\penalty0 124--127, 2015.

\bibitem[Limpert(2011)]{limpert2011brake}
Rudolf Limpert.
\newblock \emph{Brake design and safety}.
\newblock SAE international, 2011.

\bibitem[Dulac-Arnold et~al.(2021)Dulac-Arnold, Levine, Mankowitz, Li, Paduraru, Gowal, and Hester]{dulac2020empirical}
Gabriel Dulac-Arnold, Nir Levine, Daniel~J. Mankowitz, Jerry Li, Cosmin Paduraru, Sven Gowal, and Todd Hester.
\newblock An empirical investigation of the challenges of real-world reinforcement learning, 2021.
\newblock URL \url{https://arxiv.org/abs/2003.11881}.

\bibitem[Wurman et~al.(2022{\natexlab{b}})Wurman, Barrett, Kawamoto, MacGlashan, Subramanian, Walsh, Capobianco, Devlic, Eckert, Fuchs, et~al.]{wurman2022outracing}
Peter~R Wurman, Samuel Barrett, Kenta Kawamoto, James MacGlashan, Kaushik Subramanian, Thomas~J Walsh, Roberto Capobianco, Alisa Devlic, Franziska Eckert, Florian Fuchs, et~al.
\newblock Outracing champion gran turismo drivers with deep reinforcement learning.
\newblock \emph{Nature}, 602\penalty0 (7896):\penalty0 223--228, 2022{\natexlab{b}}.

\bibitem[Kaufmann et~al.(2023{\natexlab{b}})Kaufmann, Bauersfeld, Loquercio, M{\"u}ller, Koltun, and Scaramuzza]{kaufmann2023champion}
Elia Kaufmann, Leonard Bauersfeld, Antonio Loquercio, Matthias M{\"u}ller, Vladlen Koltun, and Davide Scaramuzza.
\newblock Champion-level drone racing using deep reinforcement learning.
\newblock \emph{Nature}, 620\penalty0 (7976):\penalty0 982--987, 2023{\natexlab{b}}.

\bibitem[Katz et~al.(2019)Katz, Carlo, and Kim]{Katz2019}
Benjamin Katz, Jared~DI Carlo, and Sangbae Kim.
\newblock Mini cheetah: A platform for pushing the limits of dynamic quadruped control.
\newblock \emph{Proceedings - IEEE International Conference on Robotics and Automation}, 2019-May:\penalty0 6295--6301, 5 2019.
\newblock ISSN 10504729.
\newblock \doi{10.1109/ICRA.2019.8793865}.

\bibitem[Margolis et~al.(2024)Margolis, Yang, Paigwar, Chen, and Agrawal]{Margolis2024}
Gabriel~B. Margolis, Ge~Yang, Kartik Paigwar, Tao Chen, and Pulkit Agrawal.
\newblock Rapid locomotion via reinforcement learning.
\newblock \emph{International Journal of Robotics Research}, 43:\penalty0 572--587, 4 2024.
\newblock ISSN 17413176.
\newblock \doi{10.1177/02783649231224053/ASSET/IMAGES/LARGE/10.1177_02783649231224053-FIG10.JPEG}.
\newblock URL \url{https://journals.sagepub.com/doi/full/10.1177/02783649231224053}.

\bibitem[Li et~al.(2022)Li, Dong, Qin, Chen, Xu, and Ding]{li2022quadruped}
Qikai Li, Guiyu Dong, Ripeng Qin, Jiawei Chen, Kun Xu, and Xilun Ding.
\newblock Quadruped reinforcement learning without explicit state estimation.
\newblock In \emph{2022 IEEE International Conference on Robotics and Biomimetics (ROBIO)}, pages 1989--1994. IEEE, 2022.

\bibitem[Haarnoja et~al.(2024)Haarnoja, Moran, Lever, Huang, Tirumala, Humplik, Wulfmeier, Tunyasuvunakool, Siegel, Hafner, et~al.]{haarnoja2023learning}
Tuomas Haarnoja, Ben Moran, Guy Lever, Sandy~H Huang, Dhruva Tirumala, Jan Humplik, Markus Wulfmeier, Saran Tunyasuvunakool, Noah~Y Siegel, Roland Hafner, et~al.
\newblock Learning agile soccer skills for a bipedal robot with deep reinforcement learning.
\newblock \emph{Science Robotics}, 9\penalty0 (89):\penalty0 eadi8022, 2024.

\bibitem[Haarnoja et~al.(2019)Haarnoja, Zhou, Hartikainen, Tucker, Ha, Tan, Kumar, Zhu, Gupta, Abbeel, and Levine]{haarnoja2018soft}
Tuomas Haarnoja, Aurick Zhou, Kristian Hartikainen, George Tucker, Sehoon Ha, Jie Tan, Vikash Kumar, Henry Zhu, Abhishek Gupta, Pieter Abbeel, and Sergey Levine.
\newblock Soft actor-critic algorithms and applications, 2019.
\newblock URL \url{https://arxiv.org/abs/1812.05905}.

\bibitem[Zhang et~al.(2022)Zhang, Xu, and Yu]{zhanggenerative}
Haichao Zhang, Wei Xu, and Haonan Yu.
\newblock Generative planning for temporally coordinated exploration in reinforcement learning.
\newblock In \emph{The Tenth International Conference on Learning Representations, {ICLR} 2022, Virtual Event, April 25-29, 2022}. OpenReview.net, 2022.
\newblock URL \url{https://openreview.net/forum?id=YZHES8wIdE}.

\bibitem[Borghuis et~al.(2019)Borghuis, Tadin, Lankheet, Lappin, and van~de Grind]{borghuis2019temporal}
Bart~G Borghuis, Duje Tadin, Martin~JM Lankheet, Joseph~S Lappin, and Wim~A van~de Grind.
\newblock Temporal limits of visual motion processing: psychophysics and neurophysiology.
\newblock \emph{Vision}, 3\penalty0 (1):\penalty0 5, 2019.

\bibitem[Chen et~al.(2020)Chen, Xu, Liu, Li, and Zhao]{chen2020delay}
Baiming Chen, Mengdi Xu, Zuxin Liu, Liang Li, and Ding Zhao.
\newblock Delay-aware multi-agent reinforcement learning for cooperative and competitive environments, 2020.
\newblock URL \url{https://arxiv.org/abs/2005.05441}.

\bibitem[Chen et~al.(2021)Chen, Xu, Li, and Zhao]{chen2021delay}
Baiming Chen, Mengdi Xu, Liang Li, and Ding Zhao.
\newblock Delay-aware model-based reinforcement learning for continuous control.
\newblock \emph{Neurocomputing}, 450:\penalty0 119--128, 2021.

\bibitem[Derman et~al.(2021)Derman, Dalal, and Mannor]{derman2021acting}
Esther Derman, Gal Dalal, and Shie Mannor.
\newblock Acting in delayed environments with non-stationary markov policies.
\newblock In \emph{9th International Conference on Learning Representations, {ICLR} 2021, Virtual Event, Austria, May 3-7, 2021}. OpenReview.net, 2021.
\newblock URL \url{https://openreview.net/forum?id=j1RMMKeP2gR}.

\bibitem[Long et~al.(2024)Long, Ren, Shi, Wang, Huang, Luo, and Pang]{long2024learning}
Junfeng Long, Junli Ren, Moji Shi, Zirui Wang, Tao Huang, Ping Luo, and Jiangmiao Pang.
\newblock Learning humanoid locomotion with perceptive internal model, 2024.
\newblock URL \url{https://arxiv.org/abs/2411.14386}.

\bibitem[Csomay-Shanklin et~al.(2024)Csomay-Shanklin, Compton, and Ames]{csomay2024dynamically}
Noel Csomay-Shanklin, William~D. Compton, and Aaron~D. Ames.
\newblock Dynamically feasible path planning in cluttered environments via reachable bezier polytopes, 2024.
\newblock URL \url{https://arxiv.org/abs/2411.13507}.

\bibitem[Lazcano(2024)]{lazcano2024depth}
Vanel Lazcano.
\newblock Depth map completion using a specific graph metric and balanced infinity laplacian for autonomous vehicles.
\newblock In \emph{Iberoamerican Congress on Pattern Recognition}, pages 187--197. Springer, 2024.

\bibitem[Akkad et~al.(2023)Akkad, Mansour, and Inaty]{akkad2023embedded}
Ghattas Akkad, Ali Mansour, and Elie Inaty.
\newblock Embedded deep learning accelerators: A survey on recent advances.
\newblock \emph{IEEE Transactions on Artificial Intelligence}, 2023.

\bibitem[Jafarpourmarzouni et~al.(2024)Jafarpourmarzouni, Luo, Lu, Dong, et~al.]{jafarpourmarzouni2024towards}
Reza Jafarpourmarzouni, Yichen Luo, Sidi Lu, Zheng Dong, et~al.
\newblock Towards real-time and efficient perception workflows in software-defined vehicles.
\newblock \emph{IEEE Internet of Things Journal}, 2024.

\bibitem[Rahmath~P et~al.(2022)Rahmath~P, Srivastava, Chaurasia, Pacheco, and Couto]{rahmath2022early}
Haseena Rahmath~P, Vishal Srivastava, Kuldeep Chaurasia, Roberto~G Pacheco, and Rodrigo~S Couto.
\newblock Early-exit deep neural network-a comprehensive survey.
\newblock \emph{ACM Computing Surveys}, 2022.

\bibitem[Neill(2020)]{neill2020overview}
James~O' Neill.
\newblock An overview of neural network compression, 2020.
\newblock URL \url{https://arxiv.org/abs/2006.03669}.

\bibitem[Moerland et~al.(2023)Moerland, Broekens, Plaat, Jonker, et~al.]{moerland2023model}
Thomas~M Moerland, Joost Broekens, Aske Plaat, Catholijn~M Jonker, et~al.
\newblock Model-based reinforcement learning: A survey.
\newblock \emph{Foundations and Trends{\textregistered} in Machine Learning}, 16\penalty0 (1):\penalty0 1--118, 2023.

\bibitem[Yarats et~al.(2021)Yarats, Zhang, Kostrikov, Amos, Pineau, and Fergus]{yarats2021improving}
Denis Yarats, Amy Zhang, Ilya Kostrikov, Brandon Amos, Joelle Pineau, and Rob Fergus.
\newblock Improving sample efficiency in model-free reinforcement learning from images.
\newblock In \emph{Proceedings of the AAAI Conference on Artificial Intelligence}, volume~35, pages 10674--10681, 2021.

\bibitem[Janner et~al.(2019)Janner, Fu, Zhang, and Levine]{janner2019trust}
Michael Janner, Justin Fu, Marvin Zhang, and Sergey Levine.
\newblock When to trust your model: Model-based policy optimization.
\newblock \emph{Advances in neural information processing systems}, 32, 2019.

\bibitem[Wang et~al.(2021)Wang, Li, Jiang, Zhu, Li, and Zhang]{wang2021offline}
Jianhao Wang, Wenzhe Li, Haozhe Jiang, Guangxiang Zhu, Siyuan Li, and Chongjie Zhang.
\newblock Offline reinforcement learning with reverse model-based imagination.
\newblock \emph{Advances in Neural Information Processing Systems}, 34:\penalty0 29420--29432, 2021.

\bibitem[Pathak et~al.(2017)Pathak, Agrawal, Efros, and Darrell]{pathak2017curiosity}
Deepak Pathak, Pulkit Agrawal, Alexei~A Efros, and Trevor Darrell.
\newblock Curiosity-driven exploration by self-supervised prediction.
\newblock In \emph{International conference on machine learning}, pages 2778--2787. PMLR, 2017.

\bibitem[Stadie et~al.(2015)Stadie, Levine, and Abbeel]{stadie2015incentivizing}
Bradly~C. Stadie, Sergey Levine, and Pieter Abbeel.
\newblock Incentivizing exploration in reinforcement learning with deep predictive models, 2015.
\newblock URL \url{https://arxiv.org/abs/1507.00814}.

\bibitem[Savinov et~al.(2019)Savinov, Raichuk, Vincent, Marinier, Pollefeys, Lillicrap, and Gelly]{savinov2018episodic}
Nikolay Savinov, Anton Raichuk, Damien Vincent, Rapha{\"{e}}l Marinier, Marc Pollefeys, Timothy~P. Lillicrap, and Sylvain Gelly.
\newblock Episodic curiosity through reachability.
\newblock In \emph{7th International Conference on Learning Representations, {ICLR} 2019, New Orleans, LA, USA, May 6-9, 2019}. OpenReview.net, 2019.
\newblock URL \url{https://openreview.net/forum?id=SkeK3s0qKQ}.

\bibitem[Silver et~al.(2017)Silver, Schrittwieser, Simonyan, Antonoglou, Huang, Guez, Hubert, Baker, Lai, Bolton, et~al.]{silver2017mastering}
David Silver, Julian Schrittwieser, Karen Simonyan, Ioannis Antonoglou, Aja Huang, Arthur Guez, Thomas Hubert, Lucas Baker, Matthew Lai, Adrian Bolton, et~al.
\newblock Mastering the game of go without human knowledge.
\newblock \emph{nature}, 550\penalty0 (7676):\penalty0 354--359, 2017.

\bibitem[Levine and Koltun(2013)]{levine2013guided}
Sergey Levine and Vladlen Koltun.
\newblock Guided policy search.
\newblock In \emph{International conference on machine learning}, pages 1--9. PMLR, 2013.

\bibitem[Zhang et~al.(2018)Zhang, Satija, and Pineau]{zhang2018decoupling}
Amy Zhang, Harsh Satija, and Joelle Pineau.
\newblock Decoupling dynamics and reward for transfer learning.
\newblock In \emph{6th International Conference on Learning Representations, {ICLR} 2018, Vancouver, BC, Canada, April 30 - May 3, 2018, Workshop Track Proceedings}. OpenReview.net, 2018.
\newblock URL \url{https://openreview.net/forum?id=H1aoddyvM}.

\bibitem[Sasso et~al.(2023)Sasso, Sabatelli, and Wiering]{sasso2022multi}
Remo Sasso, Matthia Sabatelli, and Marco~A. Wiering.
\newblock Multi-source transfer learning for deep model-based reinforcement learning.
\newblock \emph{Transactions on Machine Learning Research}, 2023.
\newblock ISSN 2835-8856.
\newblock URL \url{https://openreview.net/forum?id=1nhTDzxxMA}.

\bibitem[Fickinger et~al.(2021)Fickinger, Hu, Amos, Russell, and Brown]{fickinger2021scalable}
Arnaud Fickinger, Hengyuan Hu, Brandon Amos, Stuart Russell, and Noam Brown.
\newblock Scalable online planning via reinforcement learning fine-tuning.
\newblock \emph{Advances in Neural Information Processing Systems}, 34:\penalty0 16951--16963, 2021.

\bibitem[Wiestler and Diedrichsen(2013)]{wiestler2013skill}
Tobias Wiestler and J{\"o}rn Diedrichsen.
\newblock Skill learning strengthens cortical representations of motor sequences.
\newblock \emph{Elife}, 2:\penalty0 e00801, 2013.

\bibitem[Gray et~al.(2013)Gray, Gao, Hedrick, and Borrelli]{gray2013robust}
Andrew Gray, Yiqi Gao, J~Karl Hedrick, and Francesco Borrelli.
\newblock Robust predictive control for semi-autonomous vehicles with an uncertain driver model.
\newblock In \emph{2013 IEEE intelligent vehicles symposium (IV)}, pages 208--213. IEEE, 2013.

\bibitem[Galliker et~al.(2022)Galliker, Csomay-Shanklin, Grandia, Taylor, Farshidian, Hutter, and Ames]{galliker2022planar}
Manuel~Y Galliker, Noel Csomay-Shanklin, Ruben Grandia, Andrew~J Taylor, Farbod Farshidian, Marco Hutter, and Aaron~D Ames.
\newblock Planar bipedal locomotion with nonlinear model predictive control: Online gait generation using whole-body dynamics.
\newblock In \emph{2022 IEEE-RAS 21st International Conference on Humanoid Robots (Humanoids)}, pages 622--629. IEEE, 2022.

\bibitem[Farshidian et~al.(2017)Farshidian, Neunert, Winkler, Rey, and Buchli]{farshidian2017efficient}
Farbod Farshidian, Michael Neunert, Alexander~W Winkler, Gonzalo Rey, and Jonas Buchli.
\newblock An efficient optimal planning and control framework for quadrupedal locomotion.
\newblock In \emph{2017 IEEE International Conference on Robotics and Automation (ICRA)}, pages 93--100. IEEE, 2017.

\bibitem[Di~Carlo et~al.(2018)Di~Carlo, Wensing, Katz, Bledt, and Kim]{di2018dynamic}
Jared Di~Carlo, Patrick~M Wensing, Benjamin Katz, Gerardo Bledt, and Sangbae Kim.
\newblock Dynamic locomotion in the mit cheetah 3 through convex model-predictive control.
\newblock In \emph{2018 IEEE/RSJ international conference on intelligent robots and systems (IROS)}, pages 1--9. IEEE, 2018.

\bibitem[McGovern et~al.(1997)McGovern, Sutton, and Fagg]{McGovern1997RolesOM}
Amy McGovern, Richard~S. Sutton, and Andrew~H. Fagg.
\newblock Roles of macro-actions in accelerating reinforcement learning.
\newblock 1997.

\bibitem[Chang et~al.(2022)Chang, Chang, Kuo, and Lee]{chang2022reusability}
Yi-Hsiang Chang, Kuan-Yu Chang, Henry Kuo, and Chun-Yi Lee.
\newblock Reusability and transferability of macro actions for reinforcement learning.
\newblock \emph{ACM Transactions on Evolutionary Learning and Optimization}, 2\penalty0 (1):\penalty0 1--16, 2022.

\bibitem[Kim et~al.(2020)Kim, Yamada, Miyoshi, Iwata, and Yamakawa]{kim2020reinforcement}
Heecheol Kim, Masanori Yamada, Kosuke Miyoshi, Tomoharu Iwata, and Hiroshi Yamakawa.
\newblock Reinforcement learning in latent action sequence space.
\newblock In \emph{2020 IEEE/RSJ International Conference on Intelligent Robots and Systems (IROS)}, pages 5497--5503. IEEE, 2020.

\bibitem[Kalyanakrishnan et~al.(2021)Kalyanakrishnan, Aravindan, Bagdawat, Bhatt, Goka, Gupta, Krishna, and Piratla]{Kalyanakrishnan2021AnAO}
Shivaram Kalyanakrishnan, Siddharth Aravindan, Vishwajeet Bagdawat, Varun Bhatt, Harshith Goka, Archit Gupta, Kalpesh Krishna, and Vihari Piratla.
\newblock An analysis of frame-skipping in reinforcement learning, 2021.
\newblock URL \url{https://arxiv.org/abs/2102.03718}.

\bibitem[Srinivas et~al.(2017)Srinivas, Sharma, and Ravindran]{Srinivas2017DynamicAR}
A.~Srinivas, Sahil Sharma, and Balaraman Ravindran.
\newblock Dynamic action repetition for deep reinforcement learning.
\newblock In \emph{AAAI}, 2017.

\bibitem[Biedenkapp et~al.(2021)Biedenkapp, Rajan, Hutter, and Lindauer]{Biedenkapp2021TempoRLLW}
Andr{\'e} Biedenkapp, Raghu Rajan, Frank Hutter, and Marius Lindauer.
\newblock Temporl: Learning when to act.
\newblock In \emph{International Conference on Machine Learning}, pages 914--924. PMLR, 2021.

\bibitem[Sharma et~al.(2017)Sharma, Lakshminarayanan, and Ravindran]{Sharma2017LearningTR}
Sahil Sharma, Aravind~S. Lakshminarayanan, and Balaraman Ravindran.
\newblock Learning to repeat: Fine grained action repetition for deep reinforcement learning.
\newblock In \emph{5th International Conference on Learning Representations, {ICLR} 2017, Toulon, France, April 24-26, 2017, Conference Track Proceedings}. OpenReview.net, 2017.
\newblock URL \url{https://openreview.net/forum?id=B1GOWV5eg}.

\bibitem[Yu et~al.(2021)Yu, Xu, and Zhang]{yu2021taac}
Haonan Yu, Wei Xu, and Haichao Zhang.
\newblock Taac: Temporally abstract actor-critic for continuous control.
\newblock \emph{Advances in Neural Information Processing Systems}, 34:\penalty0 29021--29033, 2021.

\bibitem[Patel et~al.(2024)Patel, Sejnowski, and Siegelmann]{10.1162/neco_a_01718}
Devdhar Patel, Terrence Sejnowski, and Hava Siegelmann.
\newblock Optimizing attention and cognitive control costs using temporally layered architectures.
\newblock \emph{Neural Computation}, 36\penalty0 (12):\penalty0 2734--2763, 11 2024.
\newblock ISSN 0899-7667.
\newblock \doi{10.1162/neco_a_01718}.
\newblock URL \url{https://doi.org/10.1162/neco\_a\_01718}.

\bibitem[Dabney et~al.(2021)Dabney, Ostrovski, and Barreto]{dabney2020temporally}
Will Dabney, Georg Ostrovski, and Andr{\'{e}} Barreto.
\newblock Temporally-extended {\(\epsilon\)}-greedy exploration.
\newblock In \emph{9th International Conference on Learning Representations, {ICLR} 2021, Virtual Event, Austria, May 3-7, 2021}. OpenReview.net, 2021.
\newblock URL \url{https://openreview.net/forum?id=ONBPHFZ7zG4}.

\bibitem[Raffin et~al.(2022)Raffin, Kober, and Stulp]{raffin2022smooth}
Antonin Raffin, Jens Kober, and Freek Stulp.
\newblock Smooth exploration for robotic reinforcement learning.
\newblock In \emph{Conference on robot learning}, pages 1634--1644. PMLR, 2022.

\bibitem[Li et~al.(2024)Li, Zhou, Roth, Thilges, Otto, Lioutikov, and Neumann]{li2024open}
Ge~Li, Hongyi Zhou, Dominik Roth, Serge Thilges, Fabian Otto, Rudolf Lioutikov, and Gerhard Neumann.
\newblock Open the black box: Step-based policy updates for temporally-correlated episodic reinforcement learning.
\newblock In \emph{The Twelfth International Conference on Learning Representations, {ICLR} 2024, Vienna, Austria, May 7-11, 2024}. OpenReview.net, 2024.
\newblock URL \url{https://openreview.net/forum?id=mnipav175N}.

\bibitem[Chiappa et~al.(2024)Chiappa, Marin~Vargas, Huang, and Mathis]{chiappa2024latent}
Alberto~Silvio Chiappa, Alessandro Marin~Vargas, Ann Huang, and Alexander Mathis.
\newblock Latent exploration for reinforcement learning.
\newblock \emph{Advances in Neural Information Processing Systems}, 36, 2024.

\bibitem[Lillicrap et~al.(2019)Lillicrap, Hunt, Pritzel, Heess, Erez, Tassa, Silver, and Wierstra]{lillicrap2015continuous}
Timothy~P. Lillicrap, Jonathan~J. Hunt, Alexander Pritzel, Nicolas Heess, Tom Erez, Yuval Tassa, David Silver, and Daan Wierstra.
\newblock Continuous control with deep reinforcement learning, 2019.
\newblock URL \url{https://arxiv.org/abs/1509.02971}.

\bibitem[Fujimoto et~al.(2018)Fujimoto, Hoof, and Meger]{fujimoto2018addressing}
Scott Fujimoto, Herke Hoof, and David Meger.
\newblock Addressing function approximation error in actor-critic methods.
\newblock In \emph{International conference on machine learning}, pages 1587--1596. PMLR, 2018.

\bibitem[Cho et~al.(2014)Cho, van Merrienboer, G{\"{u}}l{\c{c}}ehre, Bahdanau, Bougares, Schwenk, and Bengio]{cho2014learning}
Kyunghyun Cho, Bart van Merrienboer, {\c{C}}aglar G{\"{u}}l{\c{c}}ehre, Dzmitry Bahdanau, Fethi Bougares, Holger Schwenk, and Yoshua Bengio.
\newblock Learning phrase representations using {RNN} encoder-decoder for statistical machine translation.
\newblock In Alessandro Moschitti, Bo~Pang, and Walter Daelemans, editors, \emph{Proceedings of the 2014 Conference on Empirical Methods in Natural Language Processing, {EMNLP} 2014, October 25-29, 2014, Doha, Qatar, {A} meeting of SIGDAT, a Special Interest Group of the {ACL}}, pages 1724--1734. {ACL}, 2014.
\newblock \doi{10.3115/V1/D14-1179}.
\newblock URL \url{https://doi.org/10.3115/v1/d14-1179}.

\bibitem[Brockman et~al.(2016)Brockman, Cheung, Pettersson, Schneider, Schulman, Tang, and Zaremba]{1606.01540}
Greg Brockman, Vicki Cheung, Ludwig Pettersson, Jonas Schneider, John Schulman, Jie Tang, and Wojciech Zaremba.
\newblock Openai gym, 2016.
\newblock URL \url{https://arxiv.org/abs/1606.01540}.

\bibitem[Sandha et~al.(2021)Sandha, Garcia, Balaji, Anwar, and Srivastava]{sandha2021sim2real}
Sandeep~Singh Sandha, Luis Garcia, Bharathan Balaji, Fatima Anwar, and Mani Srivastava.
\newblock Sim2real transfer for deep reinforcement learning with stochastic state transition delays.
\newblock In \emph{Conference on Robot Learning}, pages 1066--1083. PMLR, 2021.

\bibitem[Yarats and Kostrikov(2020)]{pytorch_sac}
Denis Yarats and Ilya Kostrikov.
\newblock Soft actor-critic (sac) implementation in pytorch.
\newblock \url{https://github.com/denisyarats/pytorch_sac}, 2020.

\bibitem[Van~de Ven et~al.(2020)Van~de Ven, Siegelmann, and Tolias]{van2020brain}
Gido~M Van~de Ven, Hava~T Siegelmann, and Andreas~S Tolias.
\newblock Brain-inspired replay for continual learning with artificial neural networks.
\newblock \emph{Nature communications}, 11\penalty0 (1):\penalty0 4069, 2020.

\bibitem[Zhao et~al.(2023)Zhao, Zhao, Boney, Kannala, and Pajarinen]{zhao2023simplified}
Yi~Zhao, Wenshuai Zhao, Rinu Boney, Juho Kannala, and Joni Pajarinen.
\newblock Simplified temporal consistency reinforcement learning.
\newblock In \emph{International Conference on Machine Learning}, pages 42227--42246. PMLR, 2023.

\bibitem[Wymbs and Grafton(2015)]{wymbs2015human}
Nicholas~F Wymbs and Scott~T Grafton.
\newblock The human motor system supports sequence-specific representations over multiple training-dependent timescales.
\newblock \emph{Cerebral cortex}, 25\penalty0 (11):\penalty0 4213--4225, 2015.

\bibitem[Favila et~al.(2024)Favila, Gurney, and Overton]{favila2024role}
Natalia Favila, Kevin Gurney, and Paul~G Overton.
\newblock Role of the basal ganglia in innate and learned behavioural sequences.
\newblock \emph{Reviews in the Neurosciences}, 35\penalty0 (1):\penalty0 35--55, 2024.

\bibitem[Jin et~al.(2014)Jin, Tecuapetla, and Costa]{jin2014basal}
Xin Jin, Fatuel Tecuapetla, and Rui~M Costa.
\newblock Basal ganglia subcircuits distinctively encode the parsing and concatenation of action sequences.
\newblock \emph{Nature neuroscience}, 17\penalty0 (3):\penalty0 423--430, 2014.

\bibitem[Jin and Costa(2015)]{jin2015shaping}
Xin Jin and Rui~M Costa.
\newblock Shaping action sequences in basal ganglia circuits.
\newblock \emph{Current opinion in neurobiology}, 33:\penalty0 188--196, 2015.

\bibitem[Berns and Sejnowski(1996)]{berns1996basal}
Gregory~S Berns and Terrence~J Sejnowski.
\newblock How the basal ganglia make decisions.
\newblock In \emph{Neurobiology of decision-making}, pages 101--113. Springer, 1996.

\bibitem[Berns and Sejnowski(1998)]{berns1998computational}
Gregory~S Berns and Terrence~J Sejnowski.
\newblock A computational model of how the basal ganglia produce sequences.
\newblock \emph{Journal of cognitive neuroscience}, 10\penalty0 (1):\penalty0 108--121, 1998.

\bibitem[Garr(2019)]{garr2019contributions}
Eric Garr.
\newblock Contributions of the basal ganglia to action sequence learning and performance.
\newblock \emph{Neuroscience \& Biobehavioral Reviews}, 107:\penalty0 279--295, 2019.

\bibitem[Doupe et~al.(2005)Doupe, Perkel, Reiner, and Stern]{doupe2005birdbrains}
Allison~J Doupe, David~J Perkel, Anton Reiner, and Edward~A Stern.
\newblock Birdbrains could teach basal ganglia research a new song.
\newblock \emph{Trends in neurosciences}, 28\penalty0 (7):\penalty0 353--363, 2005.

\bibitem[Jin and Costa(2010)]{jin2010start}
Xin Jin and Rui~M Costa.
\newblock Start/stop signals emerge in nigrostriatal circuits during sequence learning.
\newblock \emph{Nature}, 466\penalty0 (7305):\penalty0 457--462, 2010.

\bibitem[Matamales et~al.(2017)Matamales, Skrbis, Bailey, Balsam, Balleine, G{\"o}tz, and Bertran-Gonzalez]{matamales2017corticostriatal}
Miriam Matamales, Zala Skrbis, Matthew~R Bailey, Peter~D Balsam, Bernard~W Balleine, J{\"u}rgen G{\"o}tz, and Jesus Bertran-Gonzalez.
\newblock A corticostriatal deficit promotes temporal distortion of automatic action in ageing.
\newblock \emph{ELife}, 6:\penalty0 e29908, 2017.

\bibitem[Phillips et~al.(1995)Phillips, Chiu, Bradshaw, and Iansek]{phillips1995impaired}
James~G Phillips, Ed~Chiu, John~L Bradshaw, and Robert Iansek.
\newblock Impaired movement sequencing in patients with huntington's disease: a kinematic analysis.
\newblock \emph{Neuropsychologia}, 33\penalty0 (3):\penalty0 365--369, 1995.

\bibitem[Boyd et~al.(2009)Boyd, Edwards, Siengsukon, Vidoni, Wessel, and Linsdell]{boyd2009motor}
LA~Boyd, JD~Edwards, CS~Siengsukon, ED~Vidoni, BD~Wessel, and MA~Linsdell.
\newblock Motor sequence chunking is impaired by basal ganglia stroke.
\newblock \emph{Neurobiology of learning and memory}, 92\penalty0 (1):\penalty0 35--44, 2009.

\bibitem[Geissler et~al.(2021)Geissler, Frings, and Moeller]{geissler2021illuminating}
Christoph~F Geissler, Christian Frings, and Birte Moeller.
\newblock Illuminating the prefrontal neural correlates of action sequence disassembling in response--response binding.
\newblock \emph{Scientific Reports}, 11\penalty0 (1):\penalty0 22856, 2021.

\bibitem[Immink et~al.(2021)Immink, Pointon, Wright, and Marino]{immink2021prefrontal}
Maarten~A Immink, Monique Pointon, David~L Wright, and Frank~E Marino.
\newblock Prefrontal cortex activation during motor sequence learning under interleaved and repetitive practice: a two-channel near-infrared spectroscopy study.
\newblock \emph{Frontiers in Human Neuroscience}, 15:\penalty0 644968, 2021.

\bibitem[Shahnazian et~al.(2022)Shahnazian, Senoussi, Krebs, Verguts, and Holroyd]{shahnazian2022neural}
Danesh Shahnazian, Mehdi Senoussi, Ruth~M Krebs, Tom Verguts, and Clay~B Holroyd.
\newblock Neural representations of task context and temporal order during action sequence execution.
\newblock \emph{Topics in Cognitive Science}, 14\penalty0 (2):\penalty0 223--240, 2022.

\bibitem[Zielinski et~al.(2020)Zielinski, Tang, and Jadhav]{zielinski2020role}
Mark~C Zielinski, Wenbo Tang, and Shantanu~P Jadhav.
\newblock The role of replay and theta sequences in mediating hippocampal-prefrontal interactions for memory and cognition.
\newblock \emph{Hippocampus}, 30\penalty0 (1):\penalty0 60--72, 2020.

\bibitem[Malerba et~al.(2018)Malerba, Tsimring, and Bazhenov]{malerba2018learning}
Paola Malerba, Katya Tsimring, and Maxim Bazhenov.
\newblock Learning-induced sequence reactivation during sharp-wave ripples: a computational study.
\newblock In \emph{Advances in the Mathematical Sciences: AWM Research Symposium, Los Angeles, CA, April 2017}, pages 173--204. Springer, 2018.

\bibitem[Rubin et~al.(2022)Rubin, Hosman, Kelemen, Kapitonava, Willett, Coughlin, Halgren, Kimchi, Williams, Simeral, et~al.]{rubin2022learned}
Daniel~B Rubin, Tommy Hosman, Jessica~N Kelemen, Anastasia Kapitonava, Francis~R Willett, Brian~F Coughlin, Eric Halgren, Eyal~Y Kimchi, Ziv~M Williams, John~D Simeral, et~al.
\newblock Learned motor patterns are replayed in human motor cortex during sleep.
\newblock \emph{Journal of Neuroscience}, 42\penalty0 (25):\penalty0 5007--5020, 2022.

\end{thebibliography}

\newpage 
\appendix
\section{Appendix}
\addcontentsline{toc}{section}{Appendix Table of Contents}

\subsection*{Table of Contents}
\hrule
\begin{description}
    \item[A. 1] SRL Algorithm \dotfill \pageref{A1}
    \item[A. 2] Hyperparameters \dotfill \pageref{A2}
    \item[A. 3] Implementation Details \dotfill \pageref{A3}
    \item[A. 4] Practical Considerations on Low-Compute Hardware \dotfill \pageref{A4}
    \item[A. 5] Learning Curves \dotfill \pageref{A5}
    \item[A. 6] Plots for Frequency Averaged Scores \dotfill \pageref{A6}
    \item[A. 7] Plots for FAS vs. Performance for Stochastic Timestep \dotfill \pageref{A7}
    \item[A. 8] Generative Replay in Latent Space \dotfill \pageref{A8}
    \item[A. 9] Neural Basis for Sequence Learning \dotfill \pageref{Neural}
    \item[A. 10] Clarification Figure \dotfill \pageref{A10}
    \item[A. 11] Learning Curves by J \dotfill \pageref{A11}
    \item [A. 12] Randomized frame-skipping \dotfill \pageref{A12}
    \item [A. 13] Results for TempoRL Algorithm \dotfill \pageref{A13}
\end{description}
\hrule

\subsection{SRL Algorithm}\label{A1}

\begin{algorithm}[H]
\SetAlgoLined
\caption{Sequence Reinforcement Learning}
\KwIn{$\phi, \psi_1, \psi_2, \omega$. Initial parameters}
$\bar{\phi} \leftarrow \phi, \bar{\psi}_1 \leftarrow \psi_1$, $\bar{\psi}_2 \leftarrow \psi_2$ \tcp*{Initialize target network weights}
$D \leftarrow \emptyset$ \tcp*{Initialize an empty replay pool}
\For{each iteration}{
$\{a_t, a_{t+1}, \ldots, a_{t+J-1}\} \sim \pi_\omega(\{a_t, a_{t+1}, \ldots, a_{t+J-1}\}|s_t)$ \tcp*{Sample action sequence from the policy}

  \For{each action $a_t$ in the sequence}{
    $s_{t+1} \sim p(s_{t+1}|s_t, a_t)$ \tcp*{Sample transition from the environment}
    $D \leftarrow D \cup \{(s_t, a_t, r(s_t, a_t), s_{t+1})\}$ \tcp*{Store transition in the replay pool}
  }
  \For{each gradient step}{
    $\phi \leftarrow \phi - \lambda_\textbf{m} \nabla_{\phi} \mathcal{L}_{\phi}$ \tcp*{Update the model parameters}
  
  \For{$i \in \{1, 2\}$}{
      $\psi_i \leftarrow \psi_i - \lambda_Q \nabla_{\psi_i} \mathcal{L}_{\psi_i}$ \tcp*{Update the Q-function parameters}
       }
    $\{a_t, a_{t+1}, \ldots, a_{t+J-1}\} \sim \pi_\omega(\{a_t, a_{t+1}, \ldots, a_{t+J-1}\}|s_t)$ \tcp*{Sample action sequence from the policy}

    \If{iteration mod actor\_update\_frequency == 0}{
      \For{$j \in \{1, \ldots, J\}$}{
      
      $s_{j+1} \sim\textbf{m}_{\bar{\phi}}(s_{j+1}|s_j, a_j)$ \tcp*{Sample transition from the target model}
    }
      $\phi \leftarrow \omega - \lambda_\pi \nabla_{\omega} L_\omega$ \tcp*{Update policy weights}
    }

    $\alpha \leftarrow \alpha - \lambda \nabla_{\hat{\alpha}} \mathcal{L}(\alpha)$ \tcp*{Adjust temperature}
     \For{$i \in \{1, 2\}$}{
    $\bar{\psi}_i \leftarrow \tau \psi_i + (1 - \tau) \bar{\psi}_i$ \tcp*{Update target network weights}
    }
     $\bar{\phi} \leftarrow \tau \phi + (1 - \tau) \bar{\phi}$  \tcp*{Update target model weights}
  }
}
\KwOut{$\phi, \psi_1, \psi_2, \omega$\tcp*{Optimized parameters}}
\end{algorithm}

\subsection{Hyperparameters}\label{A2}
The table below lists the hyperparameters that are common between every environment used for all our experiments for the SAC and SRL algorithms:
\begin{table}[h]
\centering
\begin{tabular}{|l|l|p{6.5cm}|}
    \hline
    Hyperparameter  &  Value & description \\
    \hline
    Hidden Layer Size & 256 & Size of the hidden layers in the feed forward networks of Actor, Critic, Model and Encoder networks \\
    \hline 
    Updates per step & 1 & Number of learning updates per one step in the environment \\
    \hline 
    Target Update Interval & 1 & Inverval between each target update \\
    \hline
    $\gamma$ & 0.99 & Discount Factor \\
    \hline
    $\tau$ & 0.005 & Update rate for the target networks (Critic and Model) \\
    \hline
    Learning Rate & 0.0003 & Learning rate for all neural networks \\
    \hline
    Replay Buffer Size & $10^6$ & Size of the replay buffer \\
    \hline
    Batch Size & 256 & Batch size for learning\\ 
    \hline 
    Start Time-steps & 10000 & Initial number of steps where random policy is followed \\
    \hline
\end{tabular}
\caption{List of Common hyperparameters }
\label{appendixtable1}
\end{table}

\begin{table}[h]
\centering
\begin{tabular}{|l|l|l|l|l|l|l|}
    \hline
    Environment  &max Timestep & Eval frequency \\
    \hline 
    LunarLanderContinuous-v2& 500000 & 2500 \\
    \hline 
    Hopper-v2 & 1000000 & 5000 \\
    \hline 
    Walker2d-v2 & 1000000 & 5000 \\
    \hline 
    Ant-v2 & 5000000 & 5000 \\
    \hline 
    HalfCheetah-v2 & 5000000 & 5000 \\
    \hline
    Humanoid-v2 & 10000000 & 5000 \\
    \hline
\end{tabular}
\caption{List of environment-specific hyperparameters }
\label{appendixtable2}
\end{table}

\subsection{Implementation Details}\label{A3}

Due to its added complexity during training, SRL requires longer wall clock time for training when compared to SAC. We performed a minimal hyperparameter search over the actor update frequency parameter on the Hopper environment (tested values: 1, 2, 4, 8, 16). All the other hyperparamters were picked to be equal to the SAC implementation. We also did not perform a hyerparameter search over the size of GRU for the actor. It was picked to have the same size as the hidden layers of the feed forward network of the actor in SAC. The neural network for the model was also picked to have the same architecture as the actor from SAC, thus it has two hidden layers with 256 neurons. Similarly the encoder for the latent SRL implementation was also picked to have the same architecture. For the latent SRL implementation we also add an additional replay buffer to store transitions of length 5, to implement the temporal consistency training for the model. This was done for simplicity of the implementation, and it can be removed since it is redundant to save memory.

All experiments were performed on a GPU cluster the Nvidia 1080ti GPUs. Each run was performed using a single GPU, utilizing 8 CPU  cores of Intel(R) Xeon(R) Silver 4116 (24 core) and 16GB of memory.

We utilize the pytorch implementation of SAC (\url{https://github.com/denisyarats/pytorch_sac}) \citep{pytorch_sac}. The official github repository for SRL is: \url{https://github.com/dee0512/Sequence-Reinforcement-Learning}.

\subsection{Practical Considerations on Low-Compute Hardware}\label{A4}
In this work, we utilize a GRU for action generation. However, we did not test the performance of other recurrent architectures or transformers. Depending on the hardware constraints and the application, a more complicated or simple architecture could be utilized. Furthermore, we also leave the exploration of actor complexity to generalization to larger action sequences to future work.

Autonomous agents often have observation processing before it is fed into the RL algorithm. It should be noted that observation processing often forms a significant portion of the latency while the recurrent portion of the actor for SRL governs the actuation frequency. Furthermore, as mentioned before, SRL can also inherently handle delays by acting in a predictive manner where the sequence of actions performed in anticipation of the next state that is being processed. Furthermore, in such cases, where there is an overlap between two consecutive action sequences, additional MSE loss can be utilize to align two action sequences. We also leave this exploration to future work.

\newpage
\subsection{Learning Curves}\label{A5}

\begin{figure}[htbp]
    \centering
    \subfloat{
        \includegraphics[width=0.3\textwidth]{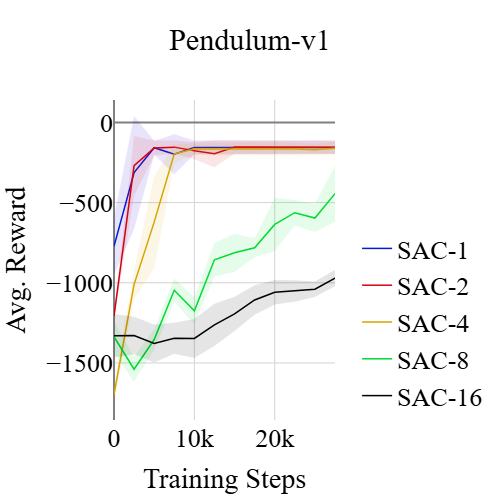}
        \label{fig:image1}
    }
    \hfill
    \subfloat{
        \includegraphics[width=0.3\textwidth]{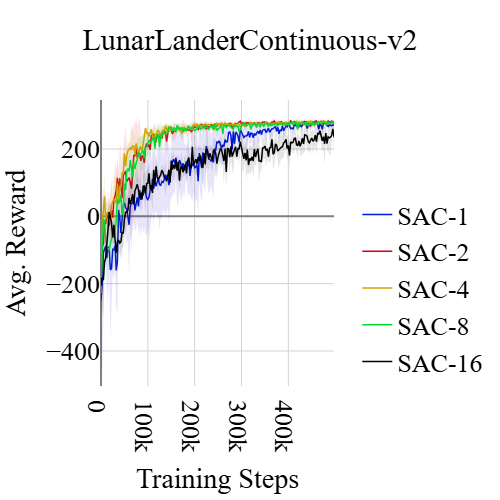}
        \label{fig:image2}
    }
    \hfill
    \subfloat{
        \includegraphics[width=0.3\textwidth]{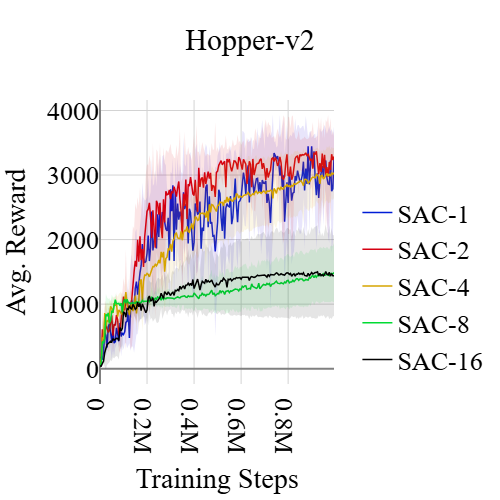}
        \label{fig:image3}
    } \\
    \subfloat{
        \includegraphics[width=0.3\textwidth]{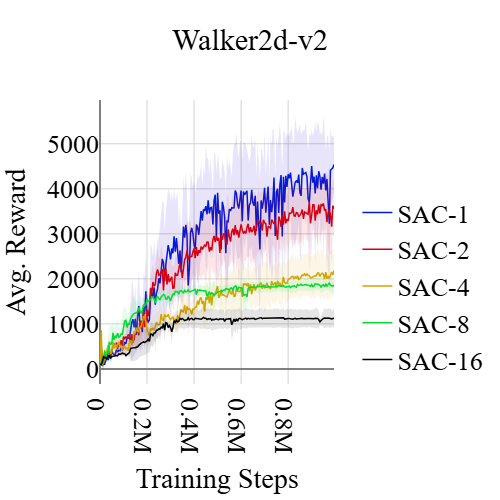}
        \label{fig:image4}
    }
    \hfill
    \subfloat{
        \includegraphics[width=0.3\textwidth]{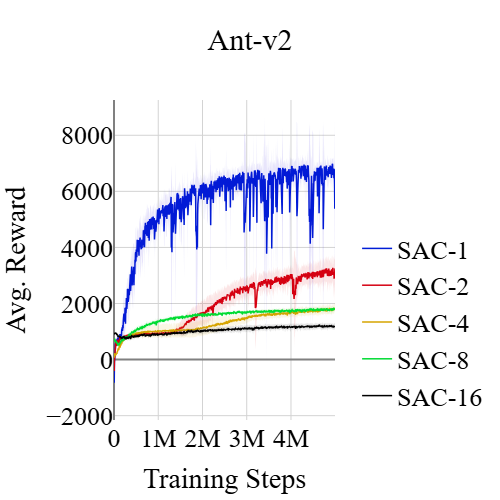}
        \label{fig:image5}
    }
    \hfill
    \subfloat{
        \includegraphics[width=0.3\textwidth]{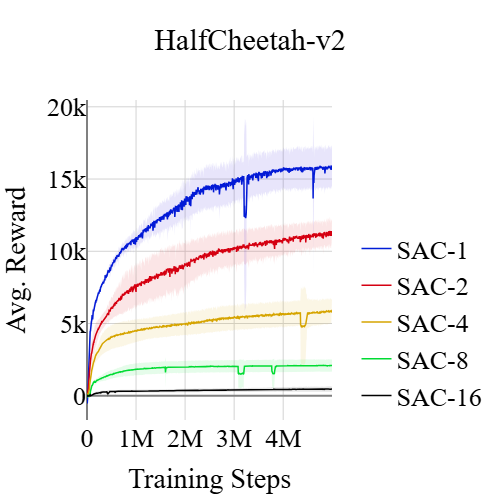}
        \label{fig:image6}
    }
      \hfill
    \subfloat{
        \includegraphics[width=0.3\textwidth]{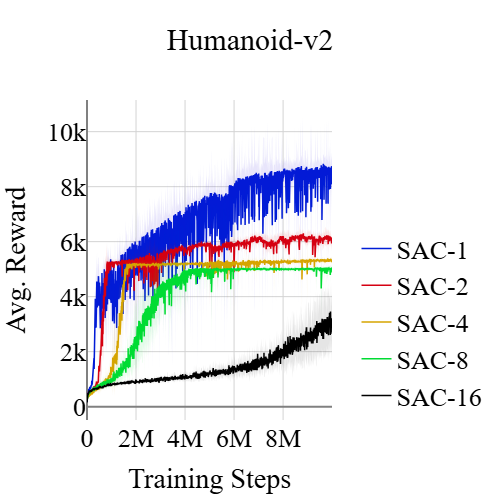}
        \label{fig:image7}
    }  \hfill
    \subfloat{
        \includegraphics[width=0.3\textwidth]{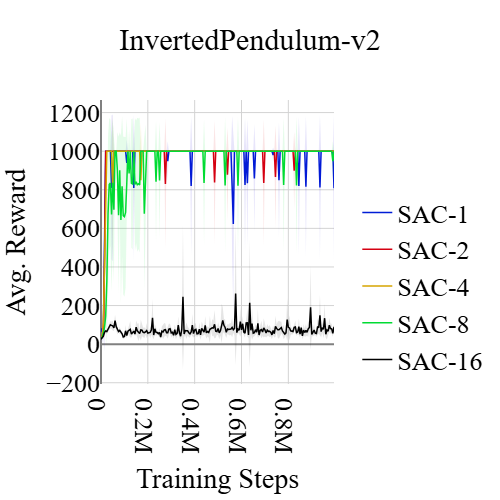}
        \label{fig:image8}
    }  \hfill
    \subfloat{
        \includegraphics[width=0.3\textwidth]{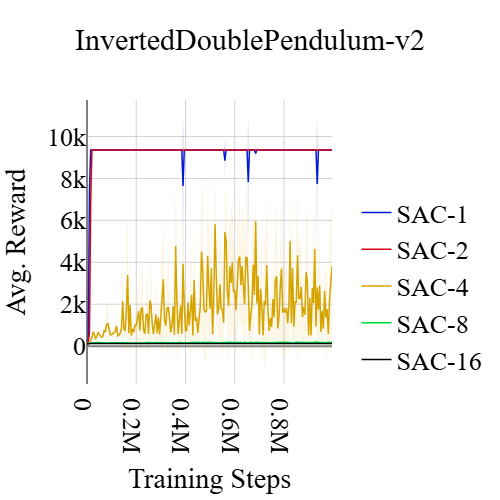}
        \label{fig:image9}
    }  \hfill
    \subfloat{
        \includegraphics[width=0.3\textwidth]{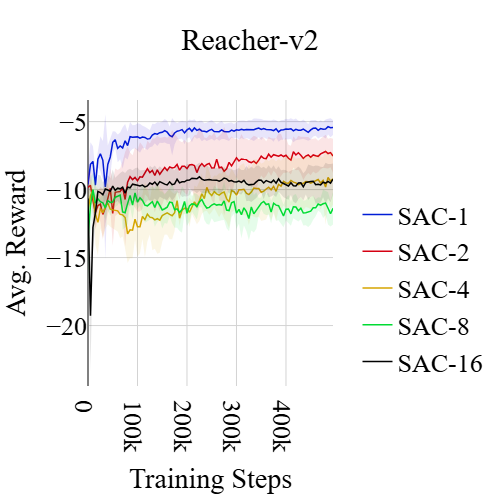}
        \label{fig:image10}
    }
    \subfloat{
        \includegraphics[width=0.3\textwidth]{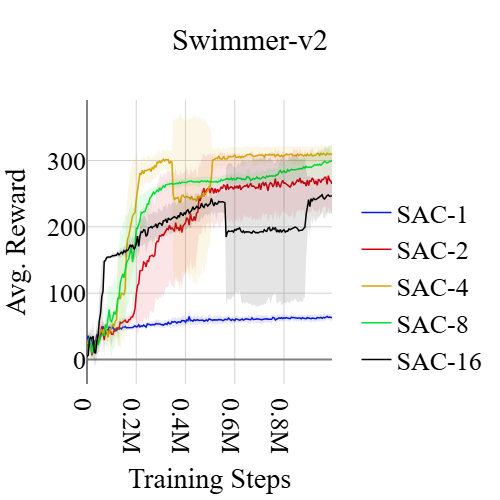}
        \label{fig:image11}
    }
    \caption{Learning curves for extended action Soft-Actor Critic (SAC-$J$) \citep{haarnoja2018soft} over continuous control tasks. The default timestep $J=1$ is the optimal for all environments except the swimmer and lunar lander. Larger timesteps support better exploration but also result in worse performance. These results demonstrate that on all environments except swimmer and lunar-lander, the default timestep is picked to optimize for the sweet-spot between better exploration and better performance. }
    \label{fig:LC}
\end{figure}

\begin{figure}[htbp]
    \centering
    \subfloat{
        \includegraphics[width=0.3\textwidth]{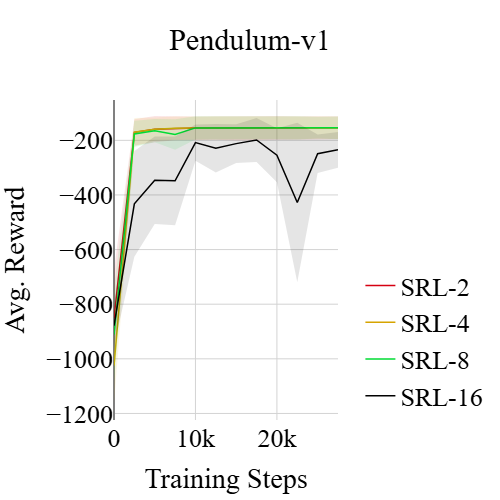}
        \label{fig:image1}
    }
    \hfill
    \subfloat{
        \includegraphics[width=0.3\textwidth]{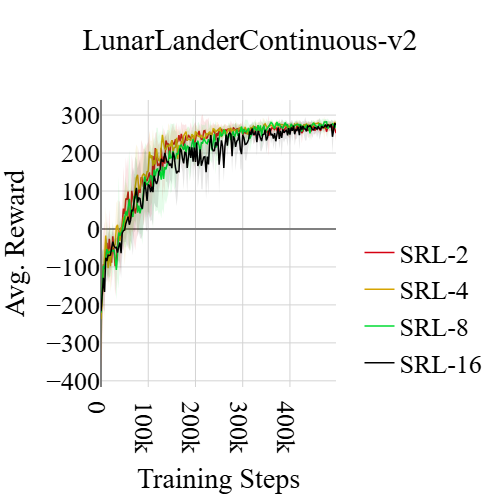}
        \label{fig:image2}
    }
    \hfill
    \subfloat{
        \includegraphics[width=0.3\textwidth]{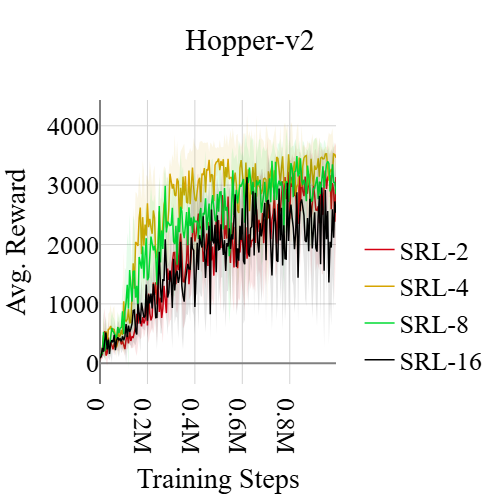}
        \label{fig:image3}
    } \\
    \subfloat{
        \includegraphics[width=0.3\textwidth]{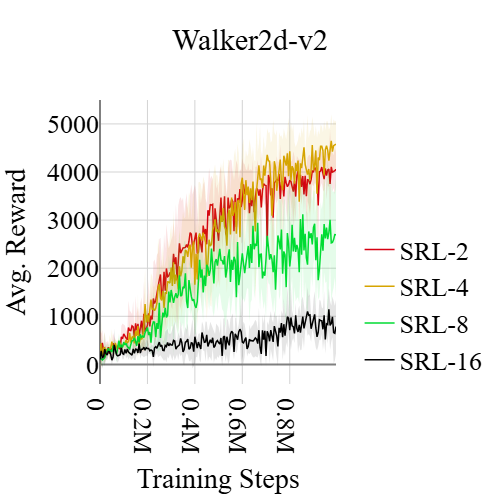}
        \label{fig:image4}
    }
    \hfill
    \subfloat{
        \includegraphics[width=0.3\textwidth]{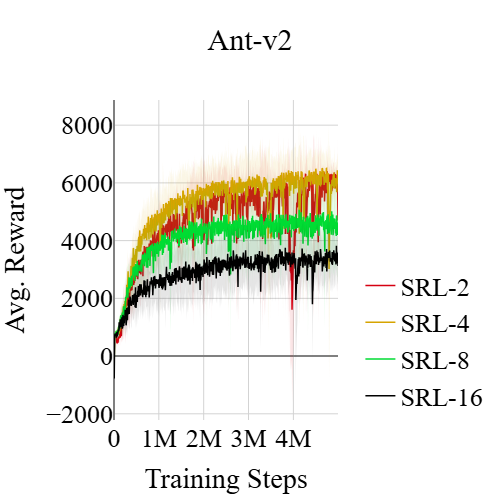}
        \label{fig:image5}
    }
    \hfill
    \subfloat{
        \includegraphics[width=0.3\textwidth]{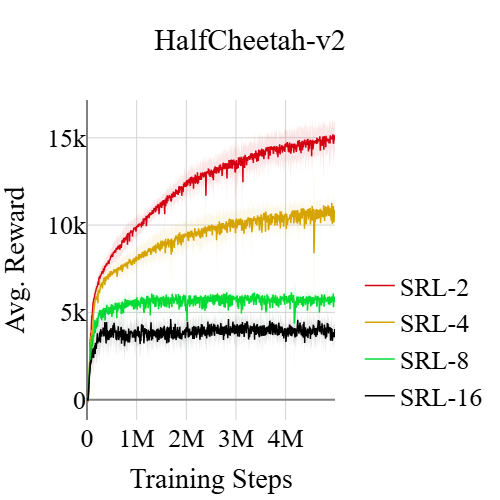}
        \label{fig:image6}
    }
      \hfill
    \subfloat{
        \includegraphics[width=0.3\textwidth]{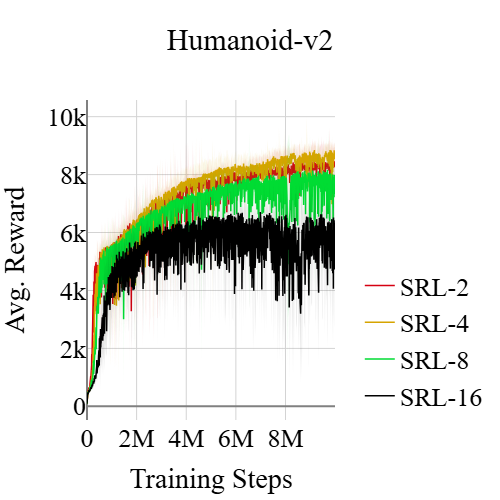}
        \label{fig:image6}
    }  \hfill
    \subfloat{
        \includegraphics[width=0.3\textwidth]{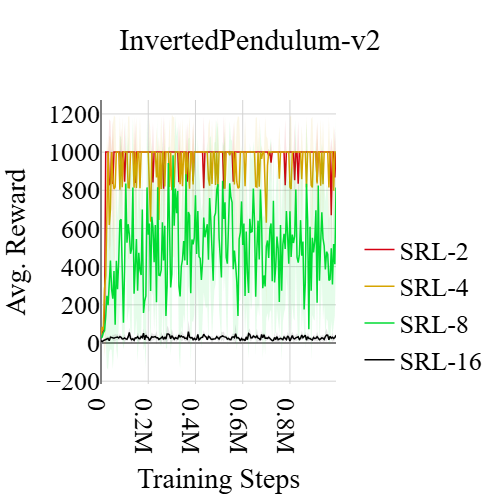}
        \label{fig:image6}
    }  \hfill
    \subfloat{
        \includegraphics[width=0.3\textwidth]{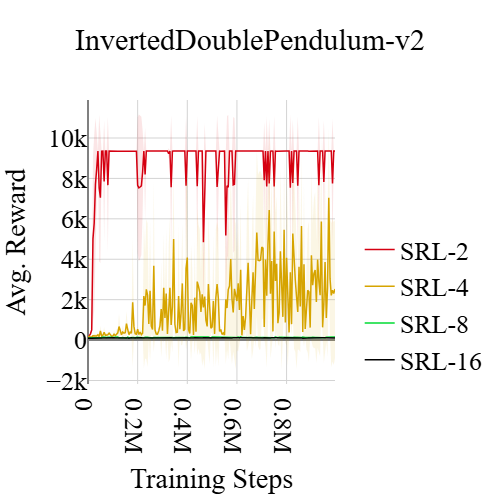}
        \label{fig:image6}
    }  \hfill
    \subfloat{
        \includegraphics[width=0.3\textwidth]{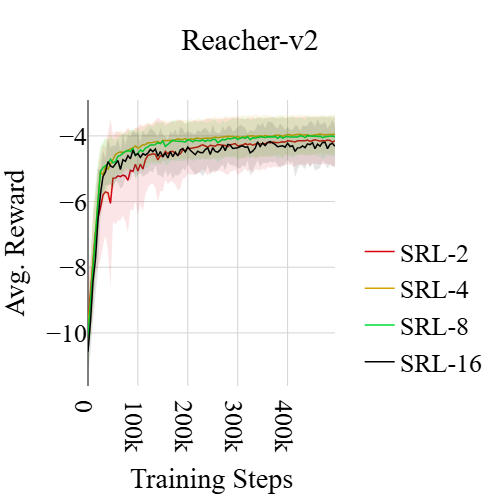}
        \label{fig:image6}
    }
     \subfloat{
        \includegraphics[width=0.3\textwidth]{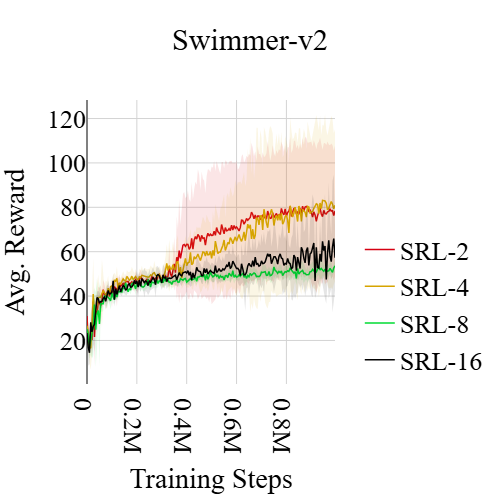}
        \label{fig:image6}
    }
    \caption{Learning curves of SRL-$J$ \citep{haarnoja2018soft} over continuous control tasks. During evaluation, SRL receives input after $J$ primitive actions. All curves are averaged over 5 trials, with shaded regions representing standard deviation. }
    \label{fig:LC}
\end{figure}

\subsection{Plots for Frequency Averaged Scores}\label{A6}

Figure \ref{fig:ASL} shows the plots for FAS. The ASL of 1 in the figure represents the performance of each policy in the standard reinforcement learning setting. We can see that SRL is competitive with SAC on ASL of 1 on all environments tested. Larger $H$ results in better robustness at longer ASLs but it often comes at the cost of lower performance at shorter ASLs. 

Additionally, as the FAS reflects, SRL is also significantly more robust across different frequencies than standard RL (SAC).  

\begin{figure}[htbp]
    \centering
    \subfloat{
        \includegraphics[width=0.3\textwidth]{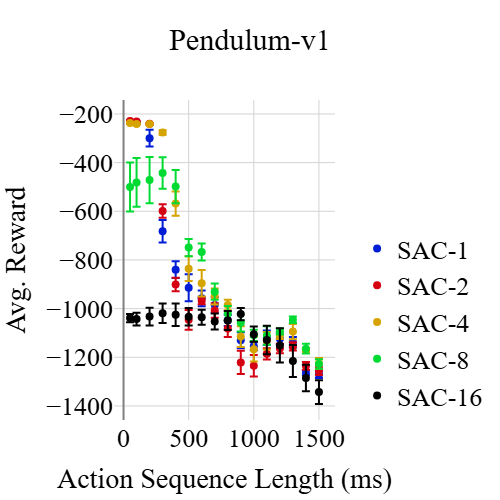}
        \label{fig:image1}
    }
    \hfill
    \subfloat{
        \includegraphics[width=0.3\textwidth]{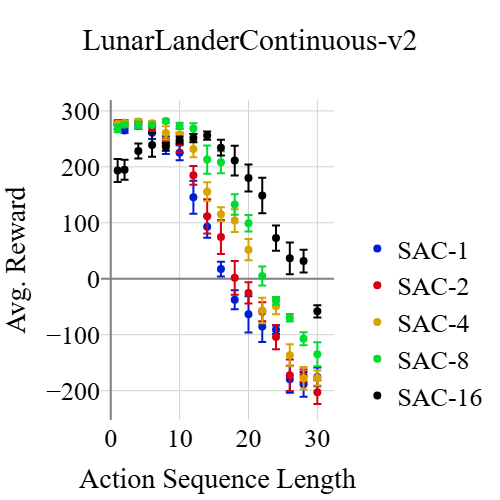}
        \label{fig:image2}
    }
    \hfill
    \subfloat{
        \includegraphics[width=0.3\textwidth]{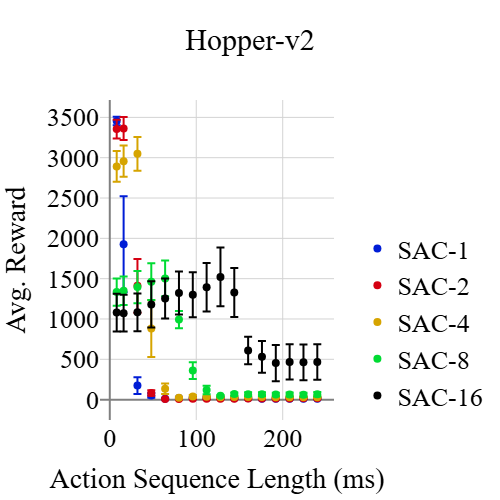}
        \label{fig:image3}
    } \\
    \subfloat{
        \includegraphics[width=0.3\textwidth]{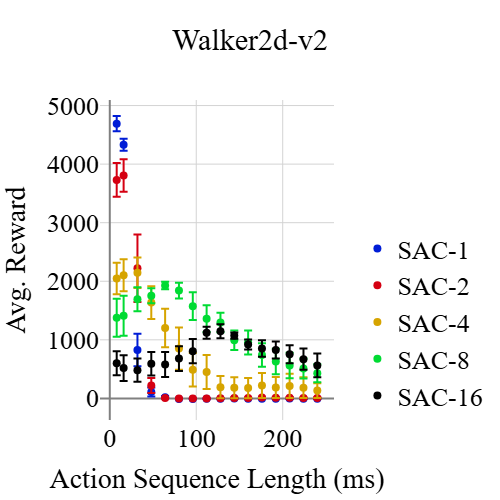}
        \label{fig:image4}
    }
    \hfill
    \subfloat{
        \includegraphics[width=0.3\textwidth]{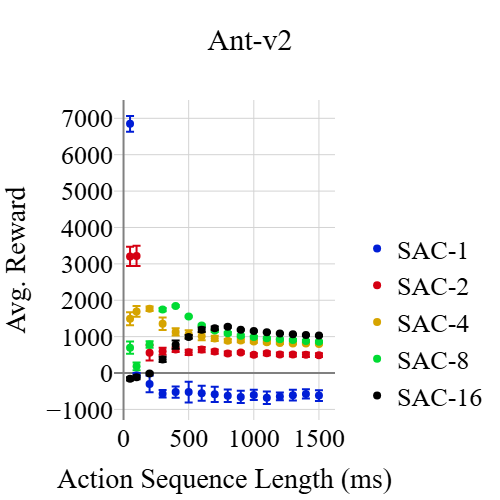}
        \label{fig:image5}
    }
    \hfill
    \subfloat{
        \includegraphics[width=0.3\textwidth]{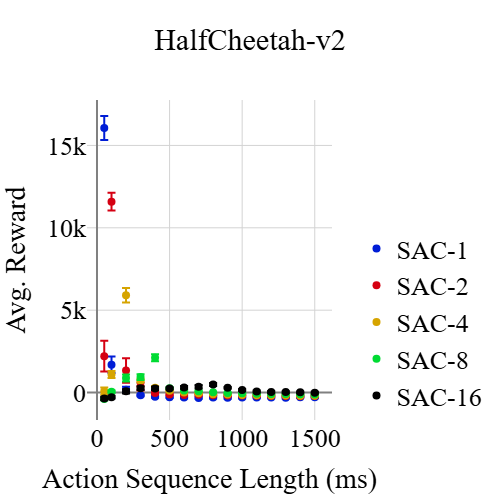}
        \label{fig:image6}
    }

    \subfloat{
        \includegraphics[width=0.3\textwidth]{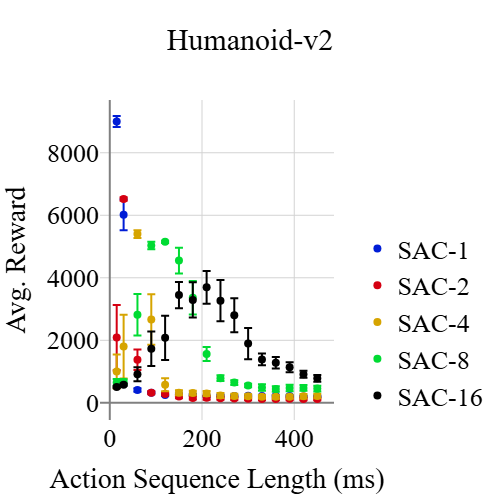}
        \label{fig:image6}
    }
    \hfill
    \subfloat{
        \includegraphics[width=0.3\textwidth]{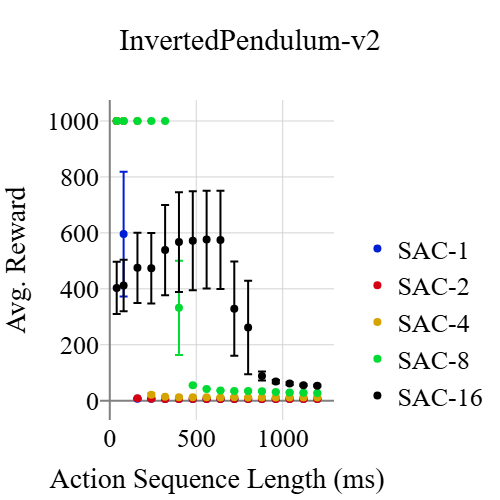}
        \label{fig:image4}
    }
    \hfill
    \subfloat{
        \includegraphics[width=0.3\textwidth]{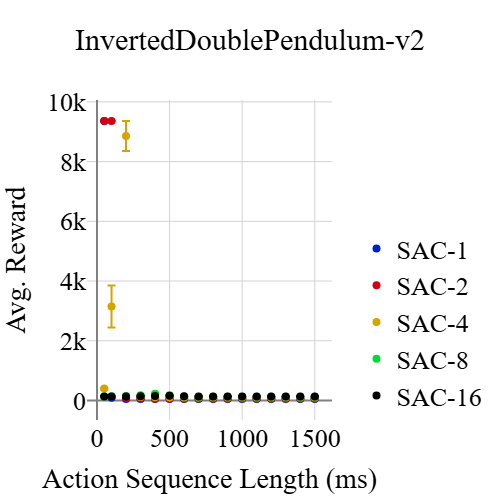}
        \label{fig:image5}
    }
 
    \subfloat{
        \includegraphics[width=0.3\textwidth]{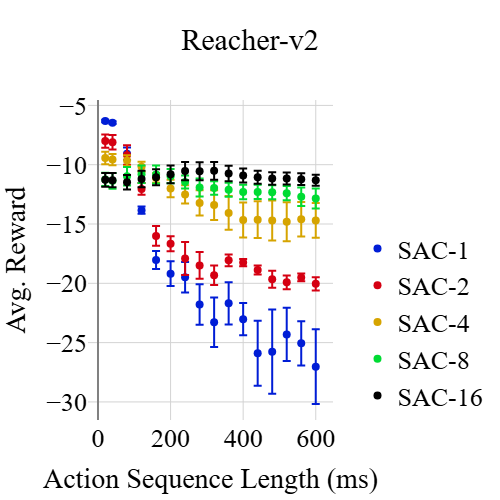}
        \label{fig:image4}
    }
    \subfloat{
        \includegraphics[width=0.3\textwidth]{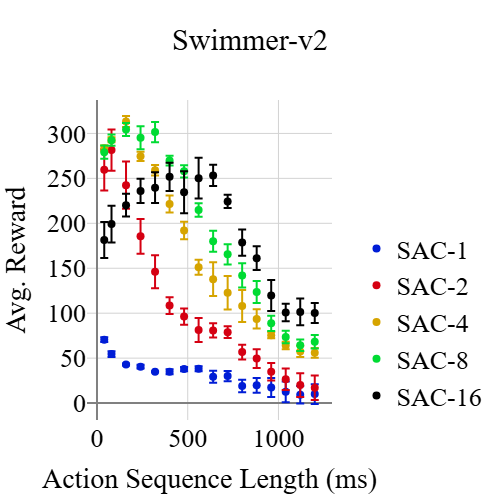}
        \label{fig:image4}
    }
    \caption{Performance of SAC-$J$ at different Action Sequence Lengths (ASL). SAC repeats the same action for the duration. All policies were tested on ASL of 1, 2, 4, 8 ... 30. All markers are averaged over 5 trials, with the error bars representing standard error.}
    \label{fig:SAC_ASL}
\end{figure}

\begin{figure}[htbp]
    \centering
    \subfloat{
        \includegraphics[width=0.3\textwidth]{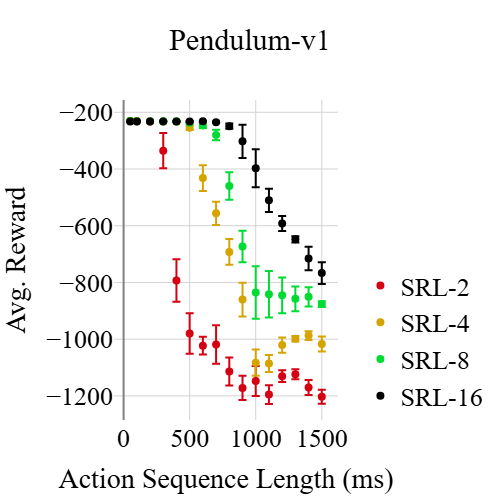}
        \label{fig:image1}
    }
    \hfill
    \subfloat{
        \includegraphics[width=0.3\textwidth]{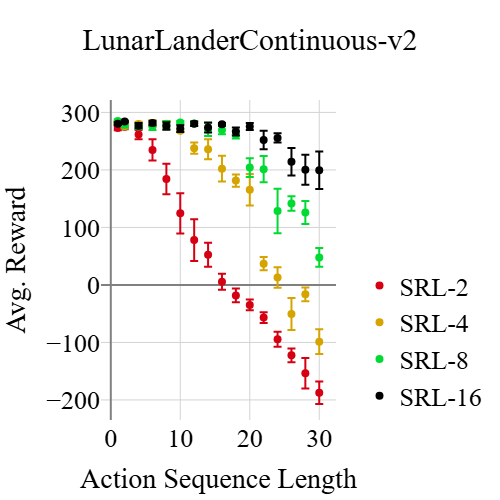}
        \label{fig:image2}
    }
    \hfill
    \subfloat{
        \includegraphics[width=0.3\textwidth]{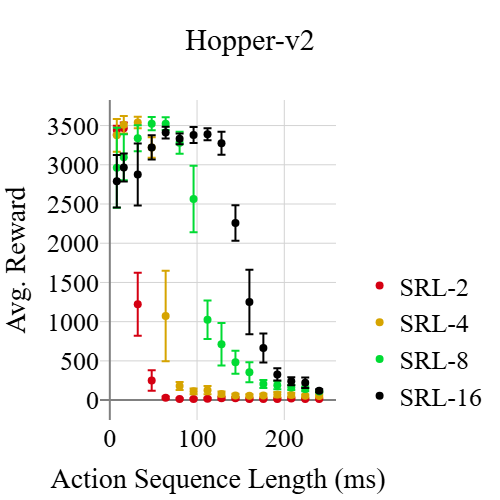}
        \label{fig:image3}
    } \\
    \subfloat{
        \includegraphics[width=0.3\textwidth]{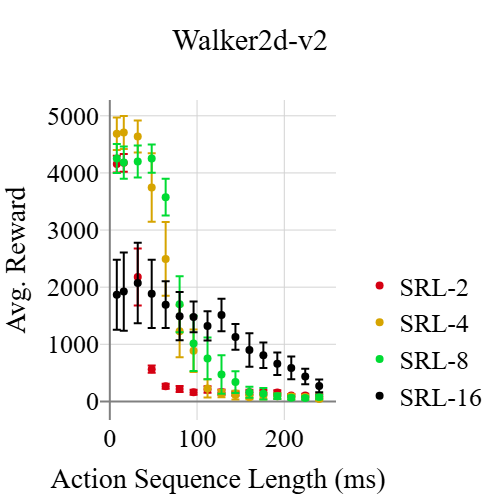}
        \label{fig:image4}
    }
    \hfill
    \subfloat{
        \includegraphics[width=0.3\textwidth]{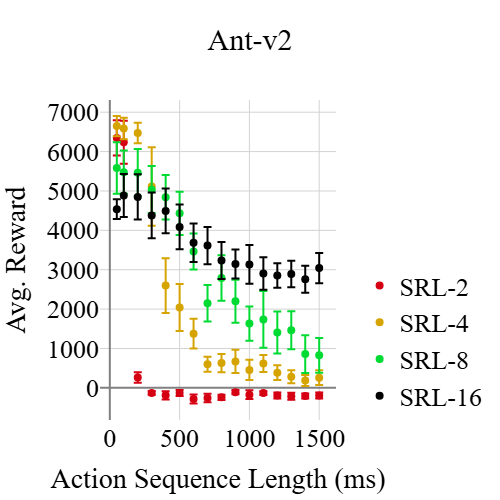}
        \label{fig:image5}
    }
    \hfill
    \subfloat{
        \includegraphics[width=0.3\textwidth]{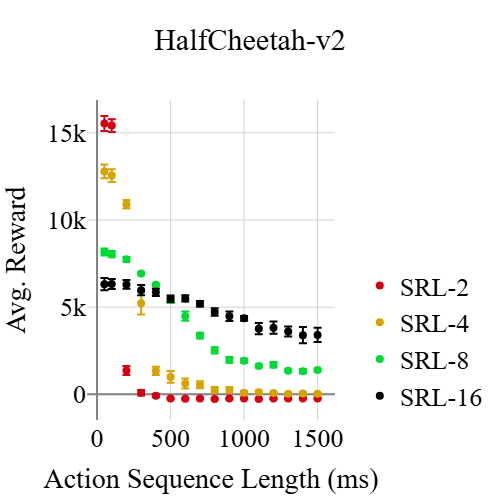}
        \label{fig:image6}
    }

    \subfloat{
        \includegraphics[width=0.3\textwidth]{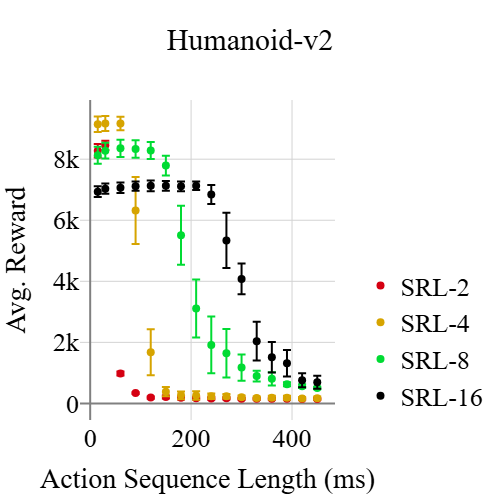}
        \label{fig:image4}
    }
    \hfill
    \subfloat{
        \includegraphics[width=0.3\textwidth]{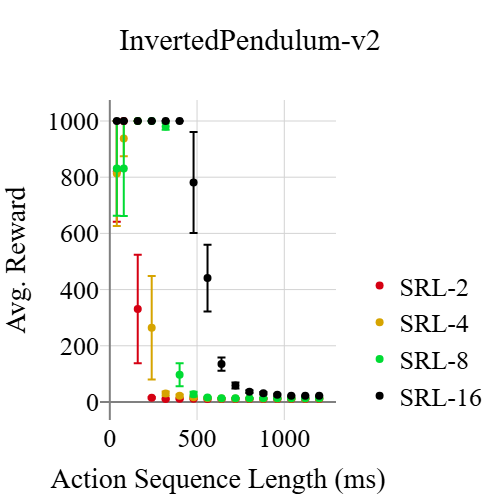}
        \label{fig:image5}
    }
    \hfill
    \subfloat{
        \includegraphics[width=0.3\textwidth]{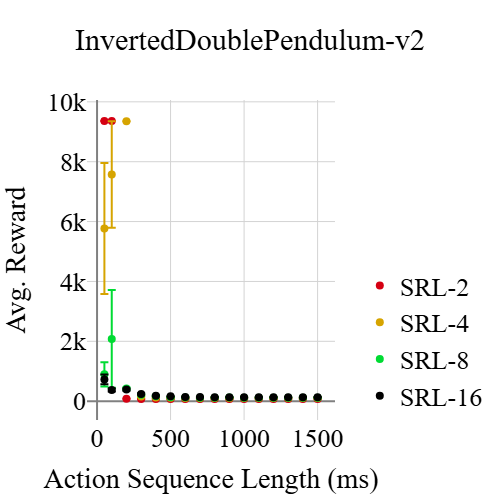}
        \label{fig:image6}
    }

    \subfloat{
        \includegraphics[width=0.3\textwidth]{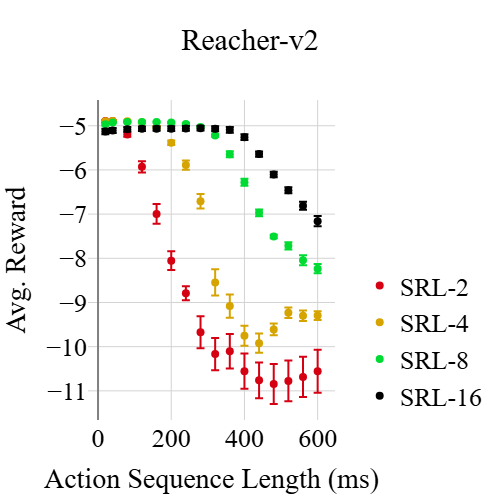}
        \label{fig:image4}
    }
       \subfloat{
        \includegraphics[width=0.3\textwidth]{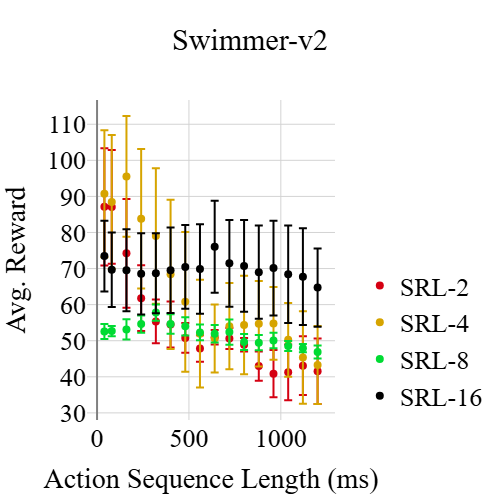}
        \label{fig:image4}
    }
    \caption{Performance of SRL-$J$ at different Action Sequence Lengths (ASL). All policies were tested on ASL of 1, 2, 4, 8 ... 30. All markers are averaged over 5 trials, with the error bars representing standard error.}
    \label{fig:ASL}
\end{figure}

\newpage
\subsection{Plots for FAS vs. Performance for Stochastic Timestep}\label{A7}

\begin{figure}[htbp]
    \centering
    \subfloat{
        \includegraphics[width=0.3\textwidth]{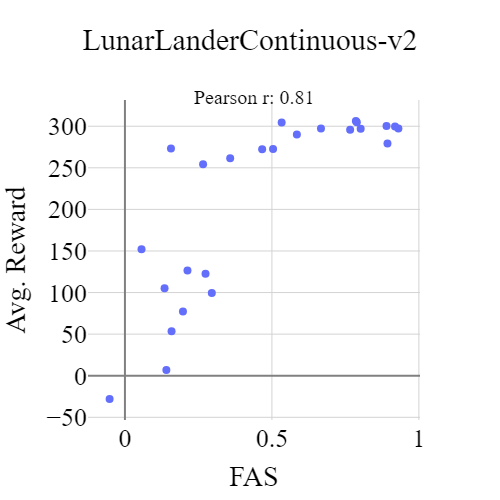}
        \label{fig:image1}
    }
    \hfill
    \subfloat{
        \includegraphics[width=0.3\textwidth]{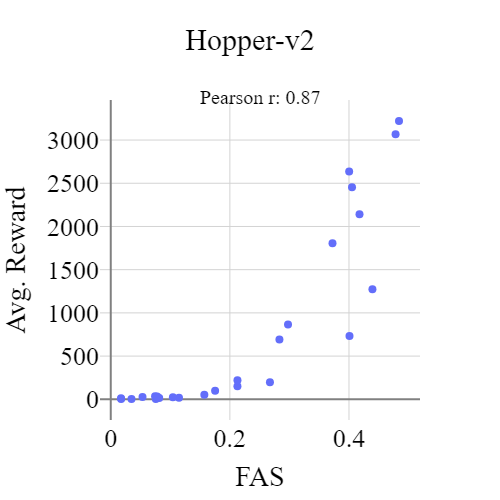}
        \label{fig:image2}
    }
    \hfill
    \subfloat{
        \includegraphics[width=0.3\textwidth]{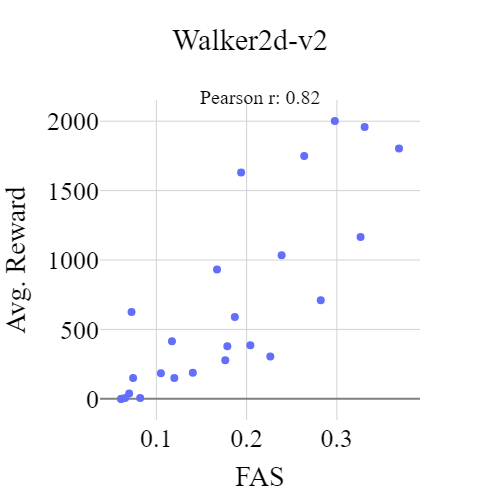}
        \label{fig:image3}
    } \\
    \subfloat{
        \includegraphics[width=0.3\textwidth]{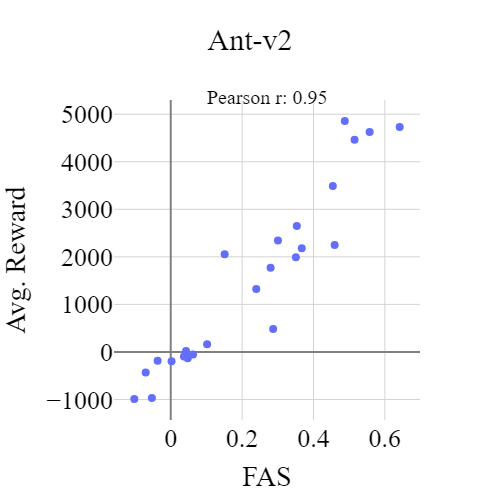}
        \label{fig:image4}
    }
    \hfill
    \subfloat{
        \includegraphics[width=0.3\textwidth]{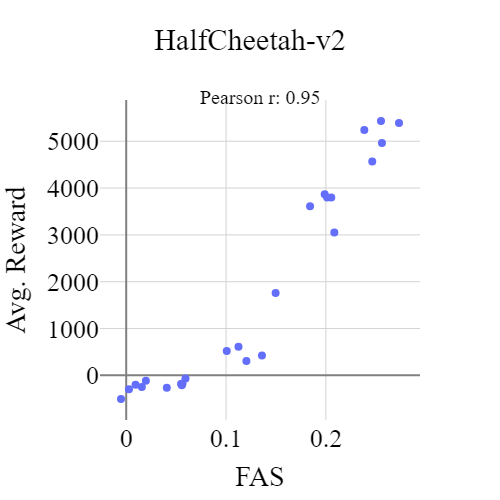}
        \label{fig:image5}
    }
    \hfill
    \subfloat{
        \includegraphics[width=0.3\textwidth]{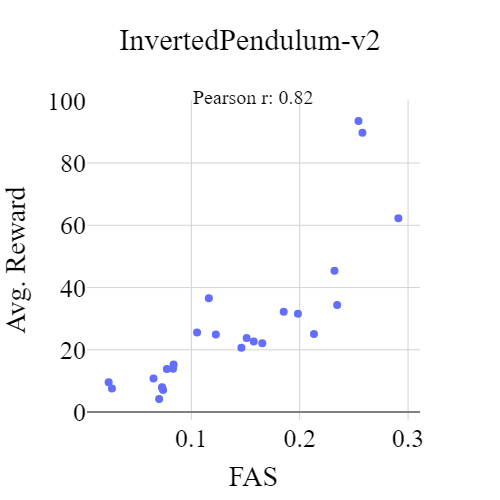}
        \label{fig:image4}
    }\\
   
    \subfloat{
        \includegraphics[width=0.3\textwidth]{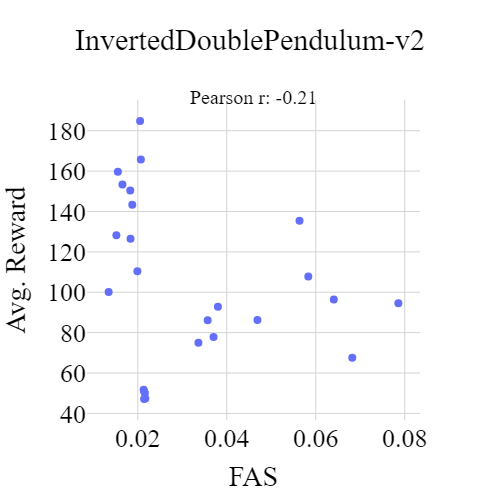}
        \label{fig:image5}
    }
    \hfill
    \subfloat{
        \includegraphics[width=0.3\textwidth]{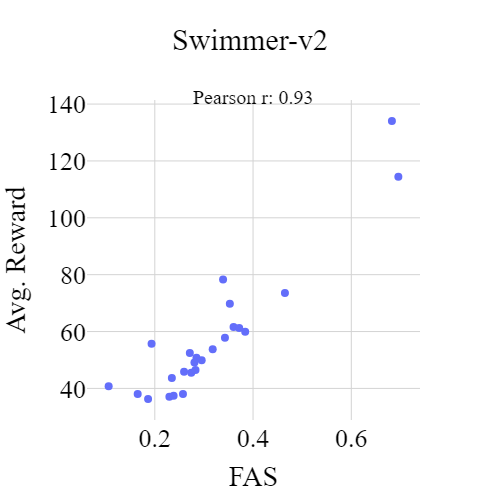}
        \label{fig:image6}
    }
    \hfill
    \subfloat{
        \includegraphics[width=0.3\textwidth]{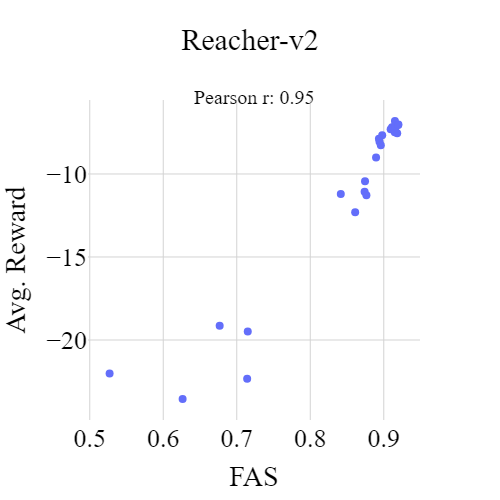}
        \label{fig:image4}
    }
    \caption{Performance vs. FAS of different policies (SAC, SRL-2, SRL-4, SRL-8, SRL-16). For each algorithm, we test 5 policies over 10 episodes.}
    \label{fig:Stochastic}
\end{figure}

In Figure \ref{fig:Stochastic}, we present the plots for FAS vs performance for all environments. For all environments except InvertedDoublePendulum-v2, we see a high correlation. InvertedDoublePendulum-v2 is a difficult problem at slow frequency and demonstrates poor performance of less than 200, thus it does not correlate to FAS. 

\subsection{Generative Replay in Latent Space}\label{A8}
\begin{figure}[htbp]
    \centering
    \subfloat{
        \includegraphics[width=0.4\textwidth]{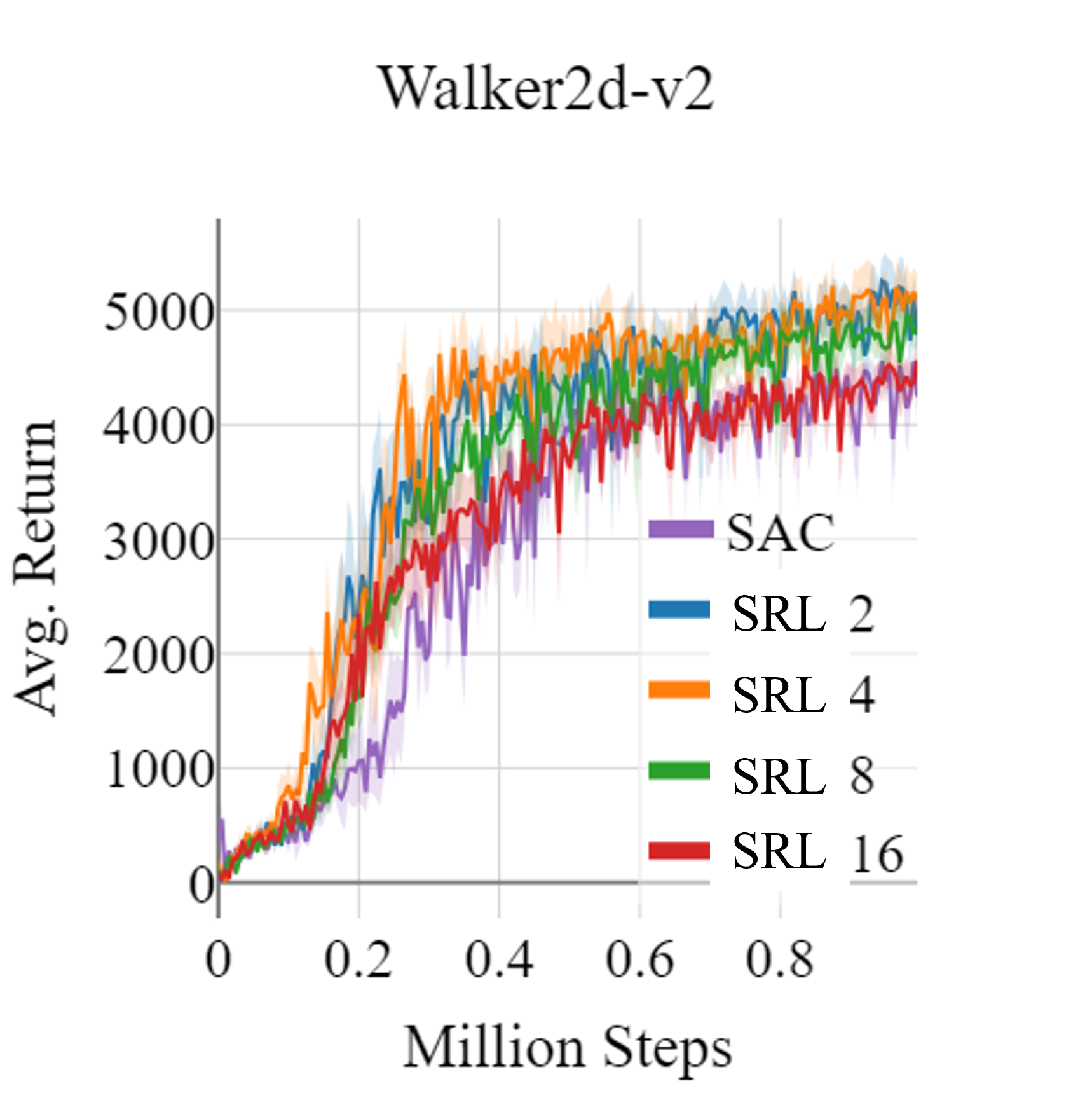}
        \label{fig:image1}
    }
    \subfloat{
        \includegraphics[width=0.4\textwidth]{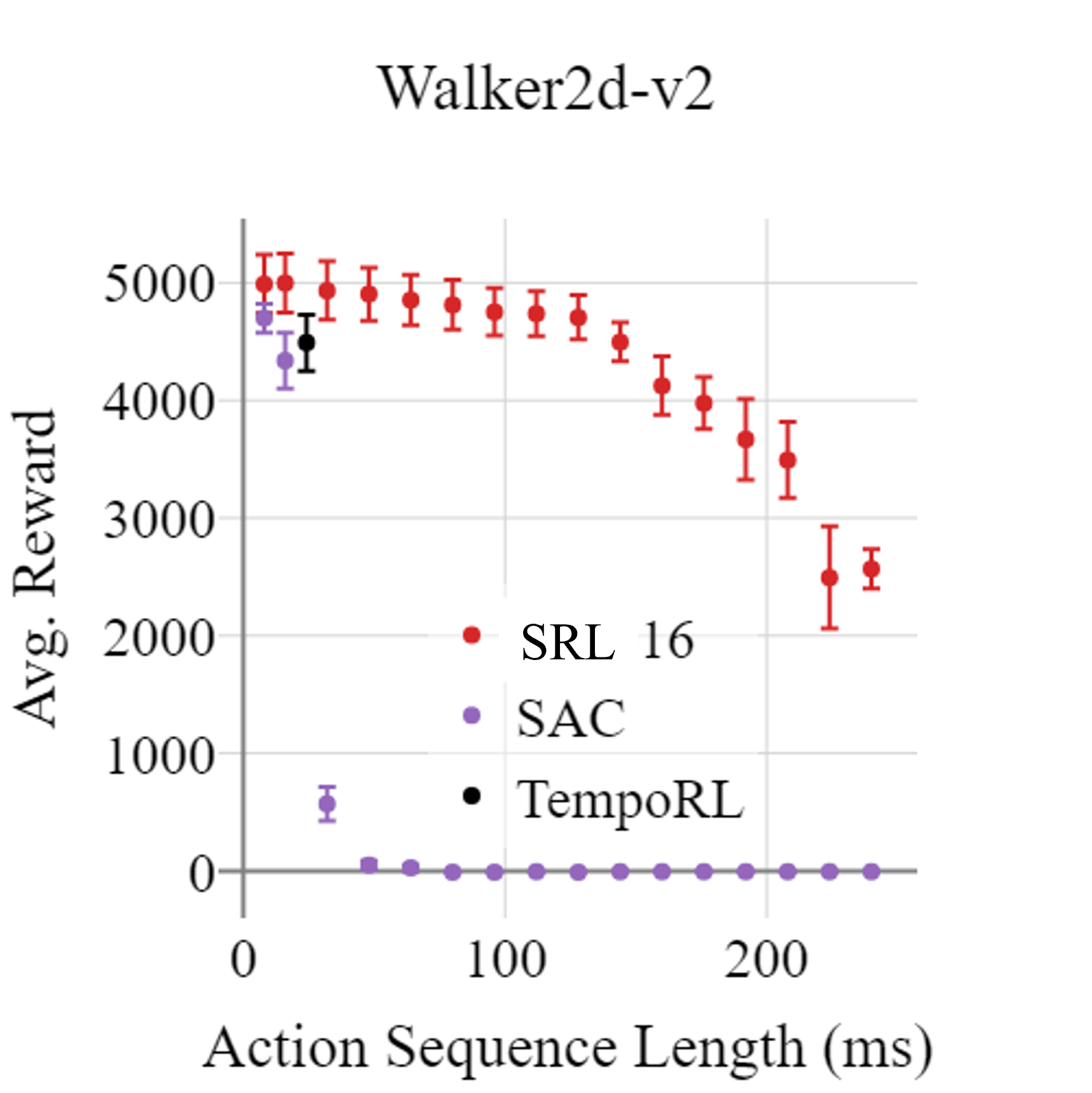}
        \label{fig:image2}
    }
    \caption{Left: Learning curve of SRL with latent state-space on the Walker2d-v2 environment. Right: Performance of latent SRL-16 on different ASL, compared to SAC and TempoRL. Utilizing a latent representation for state space is especially beneficial for the Walker2d environment so that it outperforms SAC even when training upto sequence lengths of $J=16$.}
    \label{fig:latent}
\end{figure}

Previous studies have shown that generative replay benefits greatly from latent representations \citep{van2020brain}. Recently, Simplified Temporal Consistency Reinforcement Learning (TCRL) \citep{zhao2023simplified} demonstrated that learning a latent state-space improves not only model-based planning but also model-free RL algorithms. Building on this insight, we introduced an encoder to encode the observations in our algorithm. 

Following the TCRL implementation, we use two encoders: an online encoder $\textbf{e}_\theta$ and a target encoder $\textbf{e}_{\theta^-}$, which is the exponential moving average of the online encoder:

\begin{equation}
\begin{split}
    \text{Encoder}&: e_t = \textbf{e}_\theta(s_t) 
\end{split}
\end{equation}

Thus, the model predicts the next state in the latent space. Additionally, we introduce multi-step model prediction for temporal consistency. Following the TCRL work, we use a cosine loss for model prediction. The model itself predicts only a single step forward, but we enforce temporal consistency by rolling out the model $H$-steps forward to predict $\Tilde{e}_{t+1:t+1+H}$.

 Specifically, for an $H$-step trajectory $\tau = (z_t, a_t, z_{t+1})_{t:t+H}$ drawn from the replay buffer $\mathcal{D}$, we use the online encoder to get the first latent state $e_{t}=\textbf{e}_\theta(o_t)$. Then conditioning on the sequence of actions $a_{t:t+H}$, the model is applied iteratively to predict the latent states $\Tilde{e}_{t+1}=\textbf{m}_\phi(\Tilde{e}_t, a_t)$. Finally, we use the target encoder to calculate the target latent states $\hat{e}_{t+1:t+H+1} = \textbf{e}_{\theta^-}(o_{t+1:t+1+H})$. 
The Loss function is defined as:
\begin{equation}
    \mathcal{L}_{\theta, \phi} = \mathbb{E}_{\tau\sim\mathcal{D}}\bigg[\sum_{h=0}^{H} - \gamma^h \bigg(\frac{\Tilde{e}_{t+h}}{||\Tilde{e}_{t+h}||_2}\bigg)^T \bigg(\frac{\hat{e}_{t+h}}{||\hat{e}_{t+h}||_2}\bigg)\bigg]
\end{equation}

We set $H = 5$ for our experiments. Both the encoder and the model are feed-forward neural networks with two hidden layers.

We provide preliminary results for the Walker environment. Utilizing the latent space for generative replay significantly improved performance, making it competitive even at 16 steps (128ms) (Figure \ref{fig:latent}).

We also provide the TempoRL \citep{Biedenkapp2021TempoRLLW} algorithm as a benchmark as it is an algorithm that successfully reduces the number of decisions per episodes. TempoRL is designed to dynamically pick the best frameskip (for performance), therefore we report the avg. action sequence length for TempoRL.

\subsection{Neural Basis for Sequence Learning} \label{Neural}

Unlike artificial RL agents, learning in the brain does not stop once an optimal solution has been found. During initial task learning, brain activity increases as expected, reflecting neural recruitment. However, after training and repetition, activity decreases as the brain develops more efficient representations of the action sequence, commonly referred to as muscle memory \citep{wiestler2013skill}. This phenomenon is further supported by findings that sequence-specific activity in motor regions evolves based on the amount of training, demonstrating skill-specific efficiency and specialization over time \citep{wymbs2015human}.

The neural basis for action sequence learning involves a sophisticated interconnection of different brain regions, each making a distinct contribution:

\begin{enumerate}

    \item \textbf{Basal ganglia} (BG): Action chunking is a cognitive process by which individual actions are grouped into larger, more manageable units or "chunks," facilitating more efficient storage, retrieval, and execution with reduced cognitive load \citep{favila2024role}. Importantly, this mechanism allows the brain to perform extremely fast and precise sequences of actions that would be impossible if produced individually. The BG plays a crucial role in chunking, encoding entire behavioral action sequences as a single action \citep{jin2014basal, favila2024role, jin2015shaping, berns1996basal, berns1998computational, garr2019contributions}. Dysfunction in the BG is associated with deficits in action sequences and chunking in both animals \citep{doupe2005birdbrains, jin2010start, matamales2017corticostriatal} and humans \citep{phillips1995impaired, boyd2009motor, favila2024role}. However, the neural basis for the compression of individual actions into sequences remains poorly understood.

    \item \textbf{Prefrontal cortex} (PFC): The PFC is critical for the active unbinding and dismantling of action sequences to ensure behavioral flexibility and adaptability \citep{geissler2021illuminating}. This suggests that action sequences are not merely learned through repetition; the PFC modifies these sequences based on context and task requirements. Recent research indicates that the PFC supports memory elaboration \citep{immink2021prefrontal} and maintains temporal context information \citep{shahnazian2022neural} in action sequences. The prefrontal cortex receives inputs from the hippocampus.  
    
    \item \textbf{ Hippocampus} (HC) replays neuronal activations of tasks during subsequent sleep at speeds six to seven times faster. This memory replay may explain the compression of slow actions into fast chunks. The replayed trajectories from the HC are consolidated into long-term cortical memories \citep{zielinski2020role, malerba2018learning}.  This phenomenon extends to the motor cortex, which replays motor patterns at accelerated speeds during sleep \citep{rubin2022learned}.

\end{enumerate}

\newpage

\subsection{Clarification Figure}\label{A10}
\begin{figure}[htbp]
    \centering
    \includegraphics[width=\linewidth]{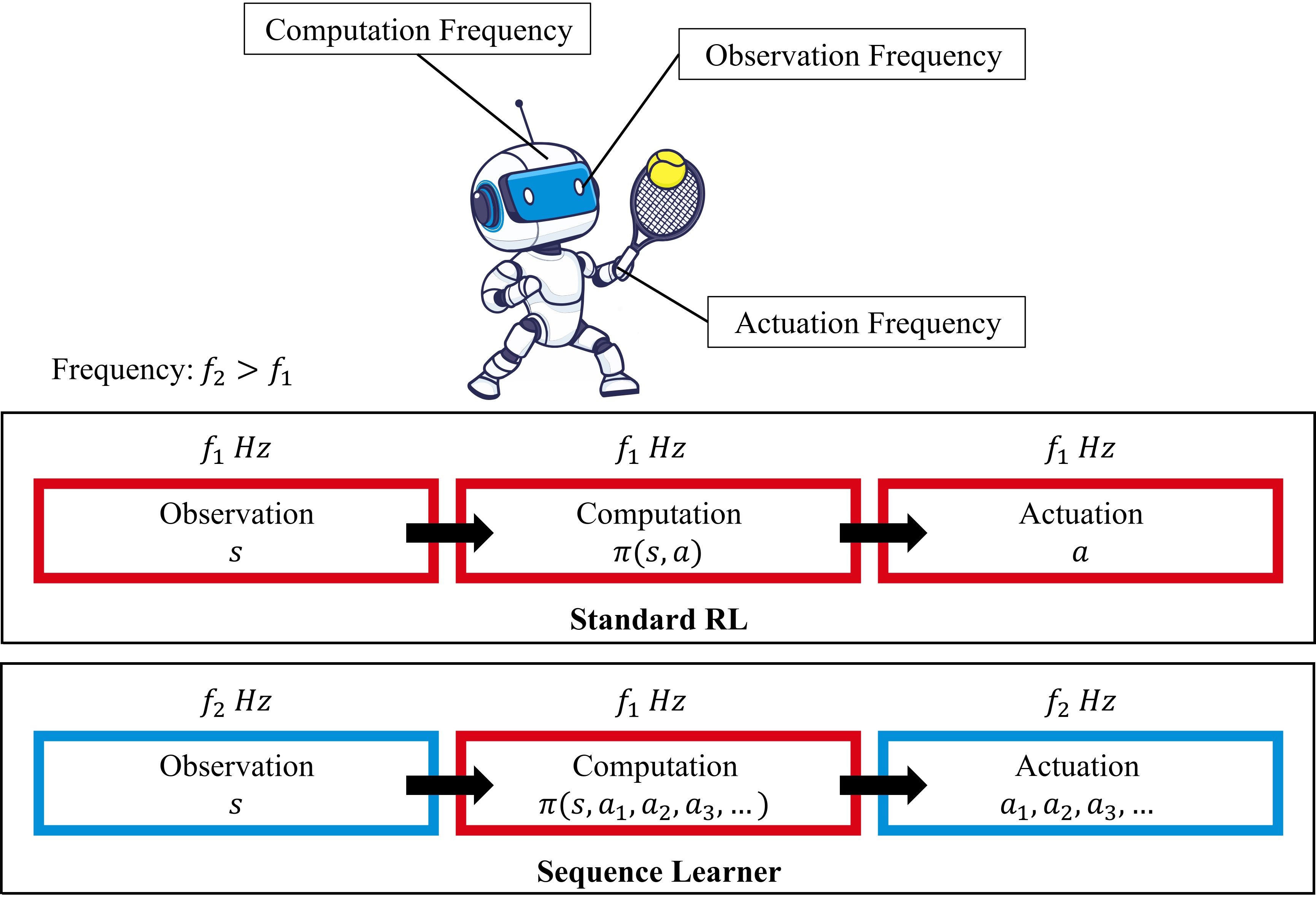}
    \caption{Illustration of the control process in an RL agent, comprising three key components: observation, computation, and actuation. In a standard RL framework, these components typically operate at the same frequency, with each observation leading to a single action after a computation pass. However, the sequence learner can achieve faster actuation by generating multiple primitive actions per observation. It's important to note that during training, the observation frequency must be at least equal to the actuation frequency and, after training, must match the computation frequency.}
    \label{fig:combined_image}
\end{figure}

\newpage
\subsection{Learning Curves by J}\label{A11}

\begin{figure}[htbp]
    \centering
    \subfloat{
        \includegraphics[width=0.16\textwidth]{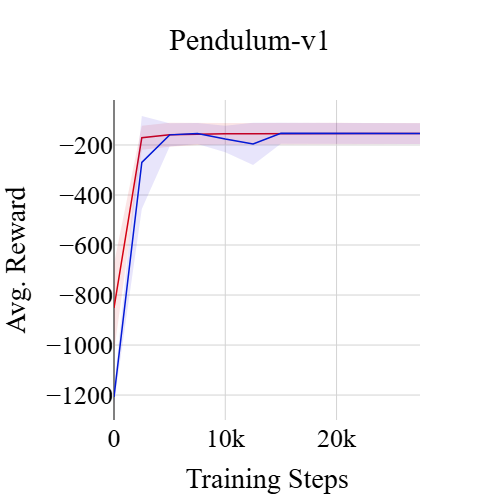}
        \label{fig:image1}
    }
    \subfloat{
        \includegraphics[width=0.16\textwidth]{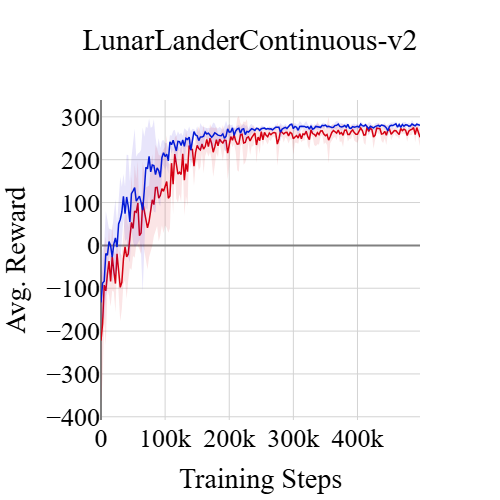}
        \label{fig:image2}
    }
    \subfloat{
        \includegraphics[width=0.16\textwidth]{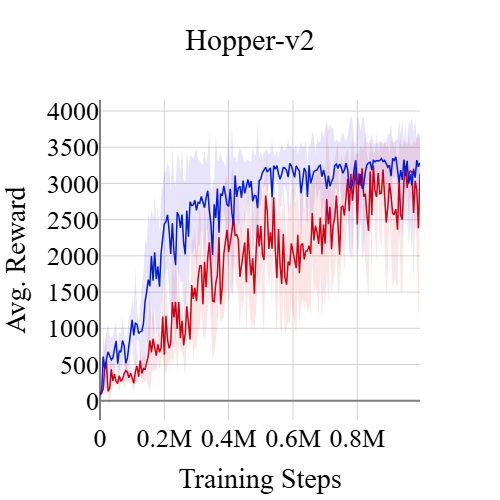}
        \label{fig:image3}
    } 
    \subfloat{
        \includegraphics[width=0.16\textwidth]{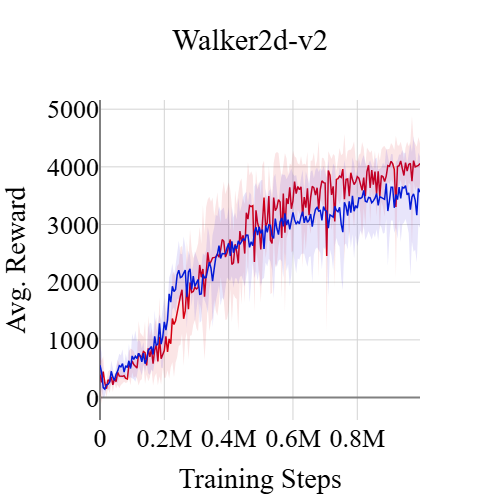}
        \label{fig:image4}
    }
    \subfloat{
        \includegraphics[width=0.16\textwidth]{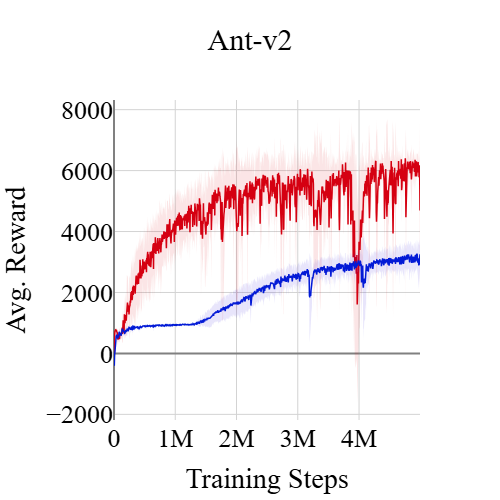}
        \label{fig:image5}
    }
    \subfloat{
        \includegraphics[width=0.16\textwidth]{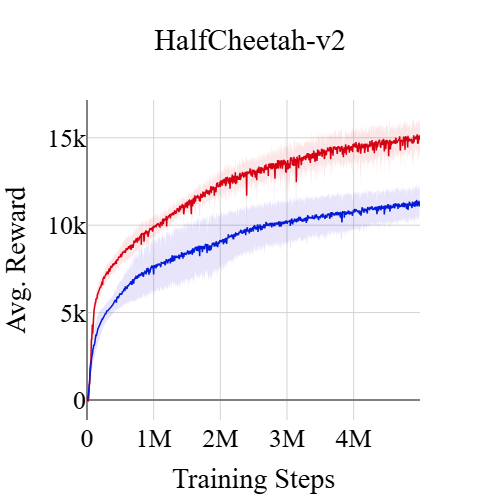}
        \label{fig:image6}
    }

    \subfloat{
        \includegraphics[width=0.16\textwidth]{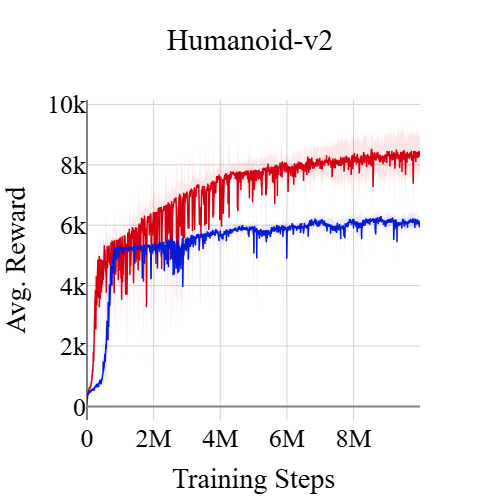}
        \label{fig:image4}
    }
    \subfloat{
        \includegraphics[width=0.16\textwidth]{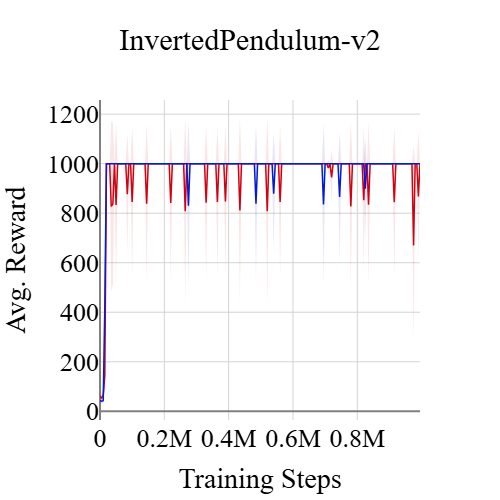}
        \label{fig:image5}
    }
    \subfloat{
        \includegraphics[width=0.16\textwidth]{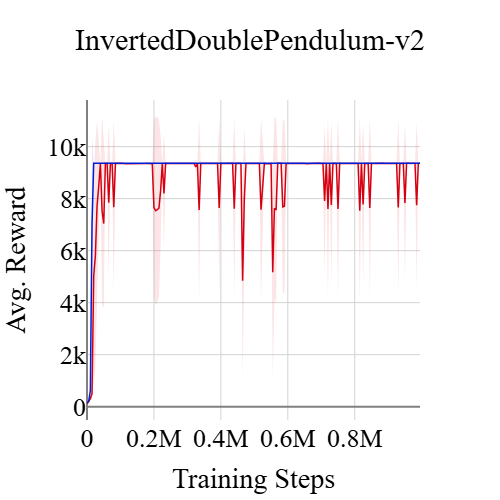}
        \label{fig:image6}
    }
    \subfloat{
        \includegraphics[width=0.16\textwidth]{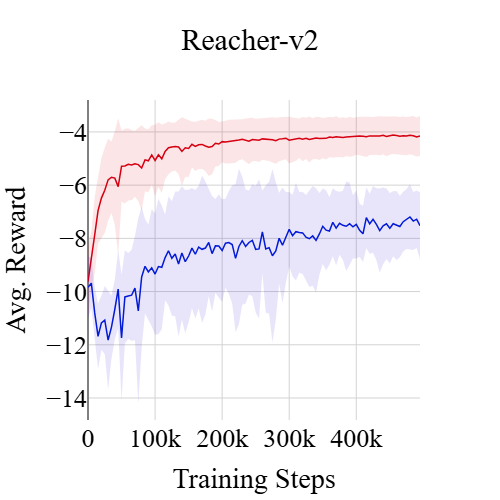}
        \label{fig:image4}
    }
   \subfloat{
    \includegraphics[width=0.16\textwidth]{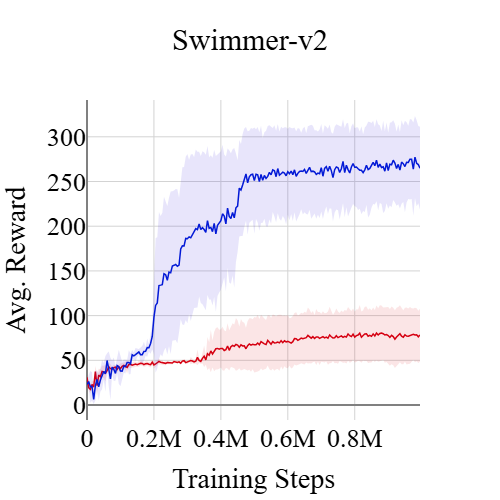}
    \label{fig:image4}
    }
    \subfloat{
    \includegraphics[width=0.16\textwidth]{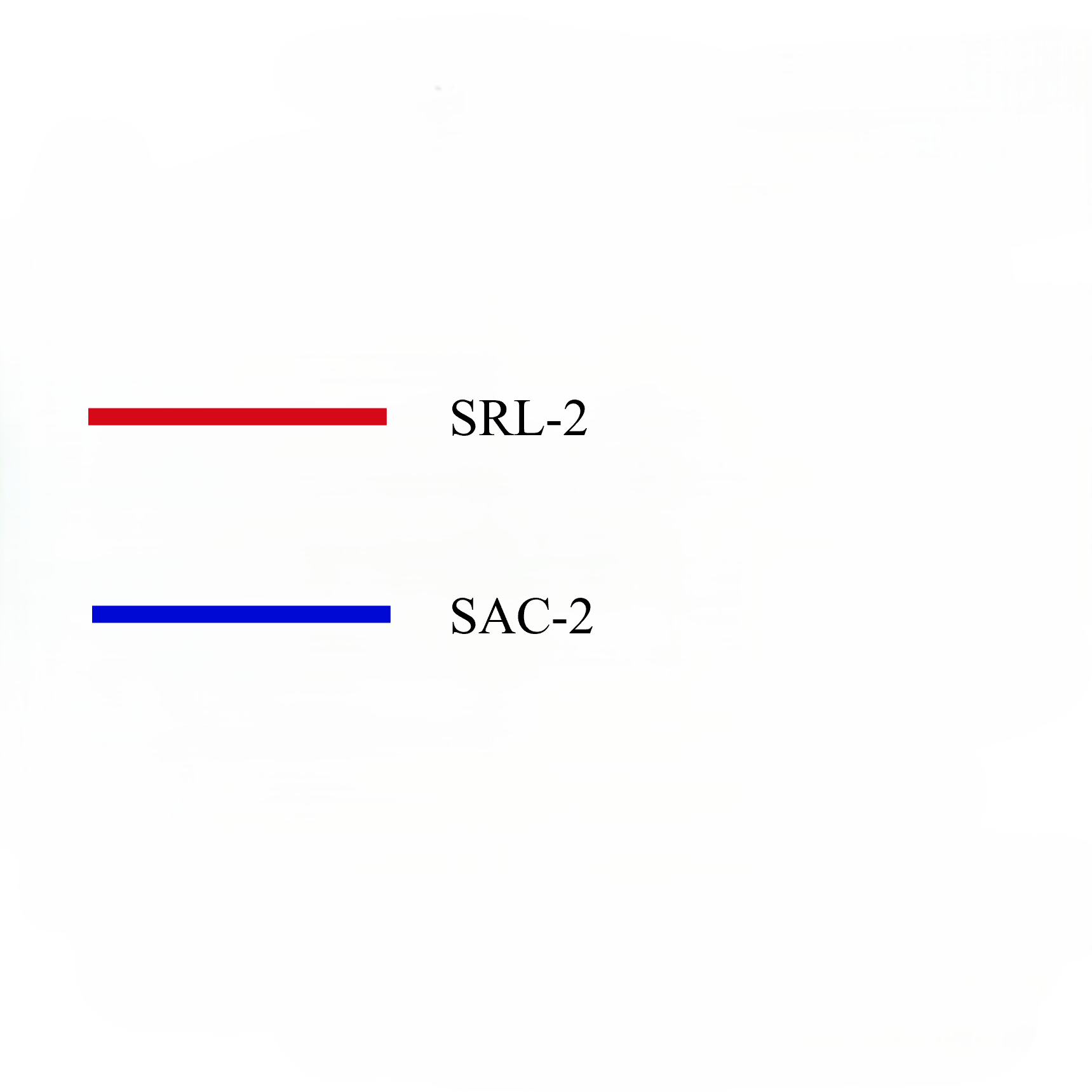}
    \label{fig:image4}
    }
    \caption{Learning curve of SRL-$2$ and SAC-$2$.}
    \label{fig:lc2}
\end{figure}

\begin{figure}[htbp]
    \centering
    \subfloat{
        \includegraphics[width=0.16\textwidth]{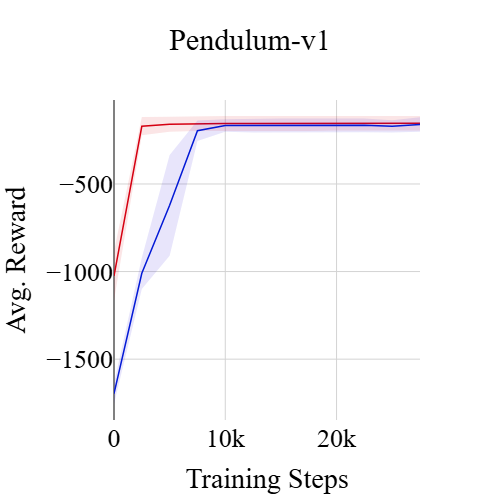}
        \label{fig:image1}
    }
    \subfloat{
        \includegraphics[width=0.16\textwidth]{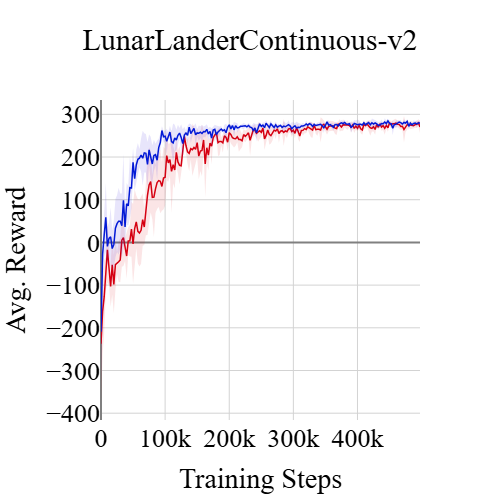}
        \label{fig:image2}
    }
    \subfloat{
        \includegraphics[width=0.16\textwidth]{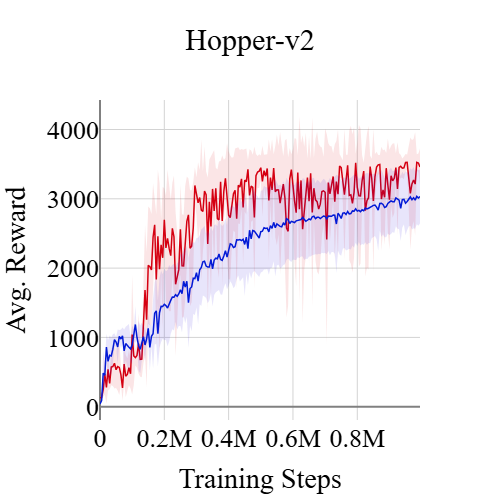}
        \label{fig:image3}
    } 
    \subfloat{
        \includegraphics[width=0.16\textwidth]{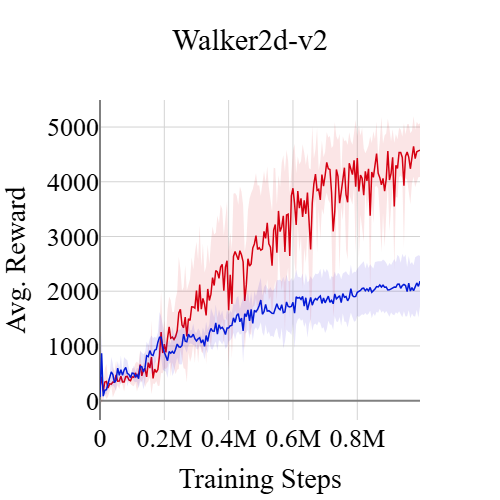}
        \label{fig:image4}
    }
    \subfloat{
        \includegraphics[width=0.16\textwidth]{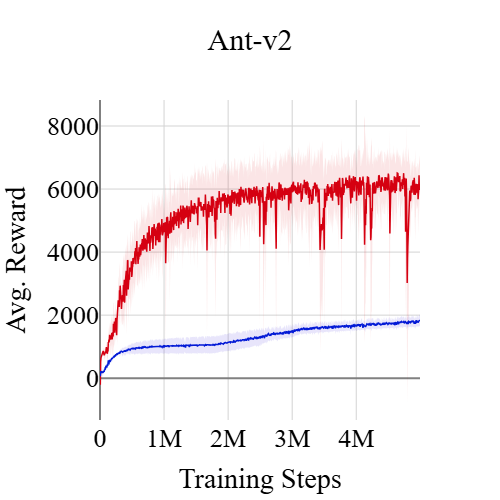}
        \label{fig:image5}
    }
    \subfloat{
        \includegraphics[width=0.16\textwidth]{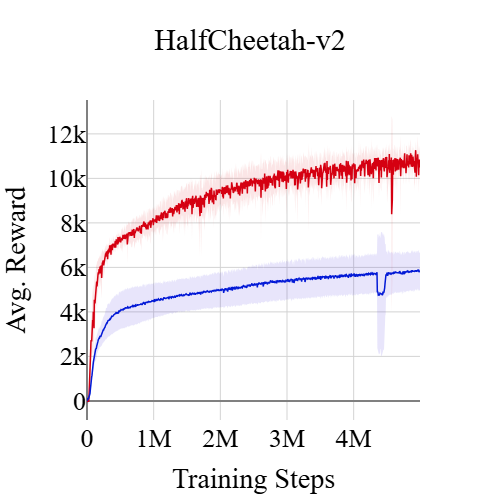}
        \label{fig:image6}
    }

    \subfloat{
        \includegraphics[width=0.16\textwidth]{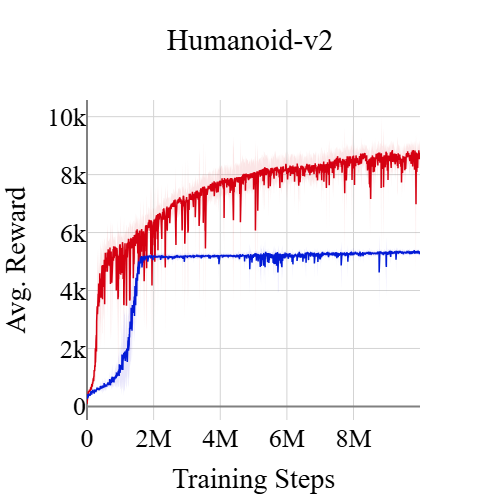}
        \label{fig:image4}
    }
    \subfloat{
        \includegraphics[width=0.16\textwidth]{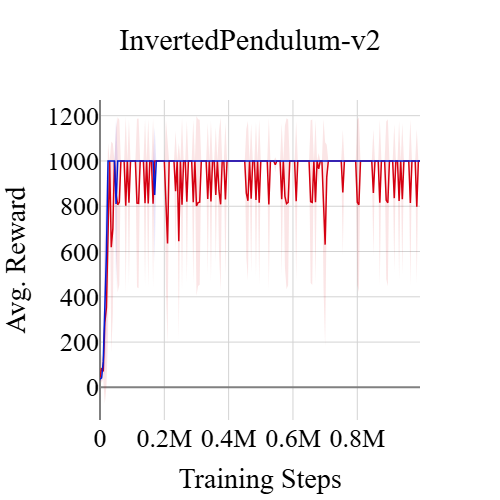}
        \label{fig:image5}
    }
    \subfloat{
        \includegraphics[width=0.16\textwidth]{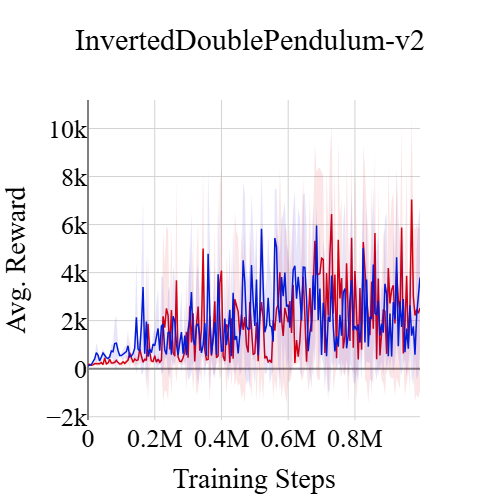}
        \label{fig:image6}
    }
    \subfloat{
        \includegraphics[width=0.16\textwidth]{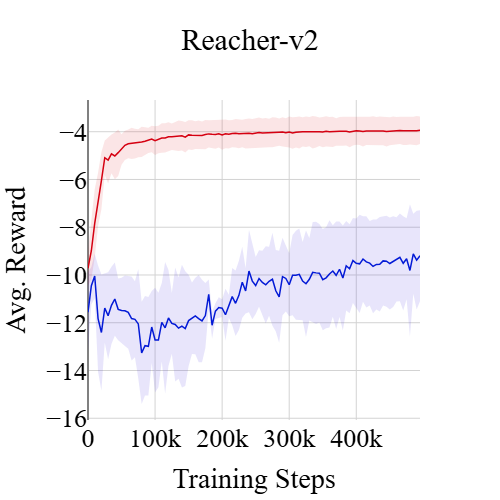}
        \label{fig:image4}
    }
   \subfloat{
    \includegraphics[width=0.16\textwidth]{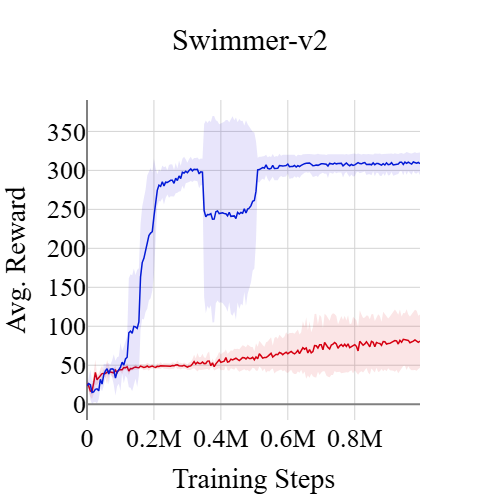}
    \label{fig:image4}
    }
    \subfloat{
    \includegraphics[width=0.16\textwidth]{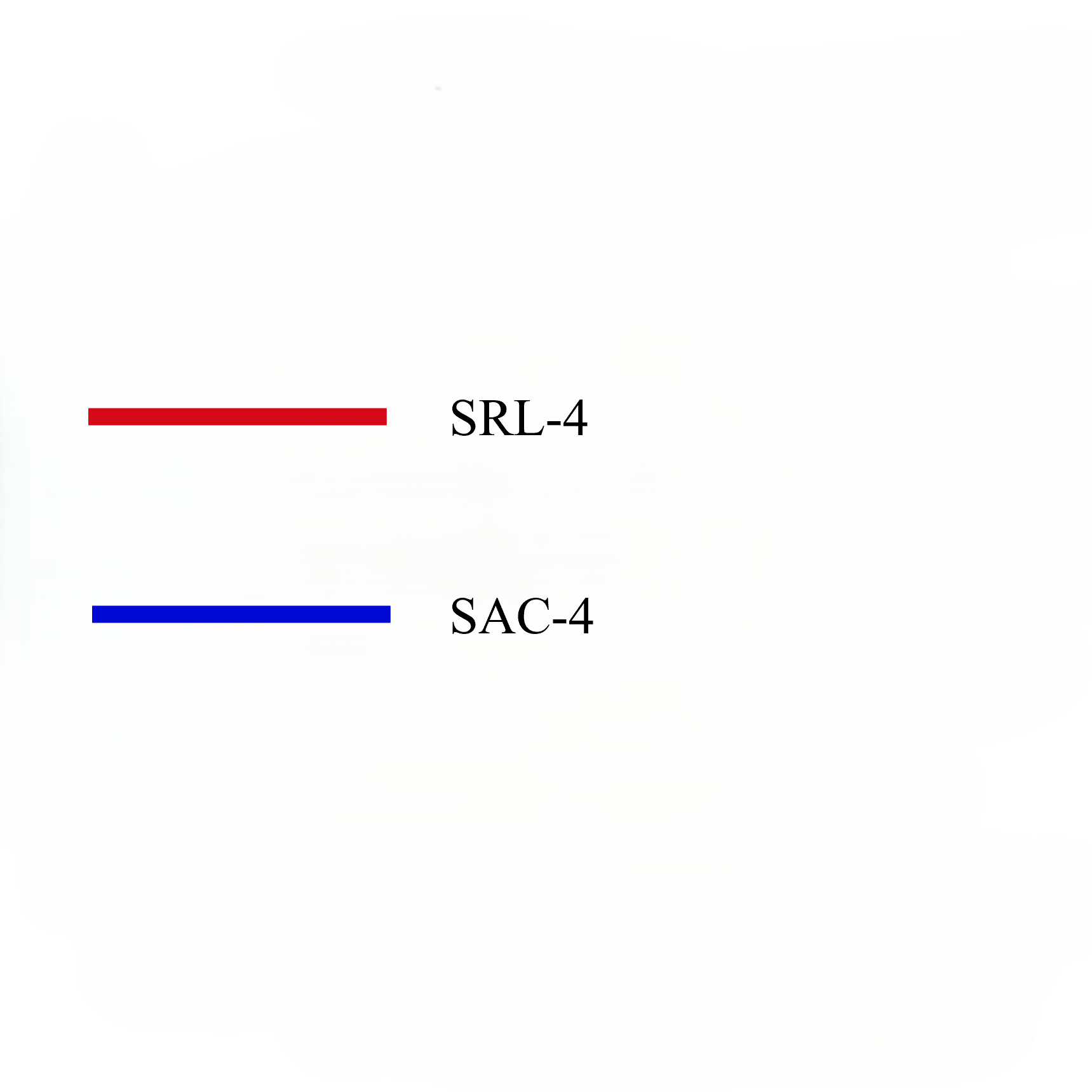}
    \label{fig:image4}
    }
    \caption{Learning curve of SRL-$4$ and SAC-$4$.}
    \label{fig:lc2}
\end{figure}

\begin{figure}[htbp]
    \centering
    \subfloat{
        \includegraphics[width=0.16\textwidth]{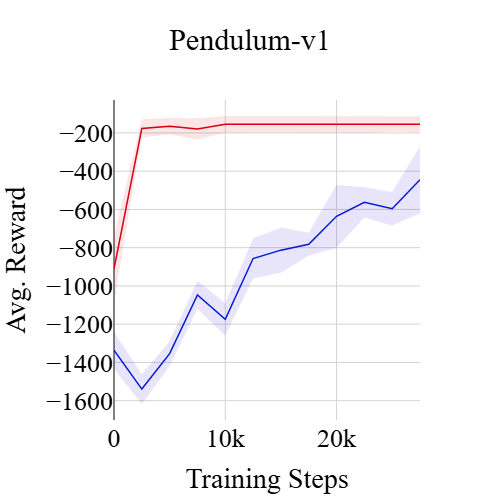}
        \label{fig:image1}
    }
    \subfloat{
        \includegraphics[width=0.16\textwidth]{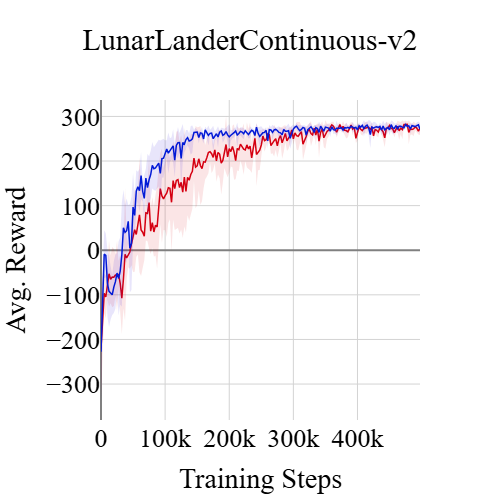}
        \label{fig:image2}
    }
    \subfloat{
        \includegraphics[width=0.16\textwidth]{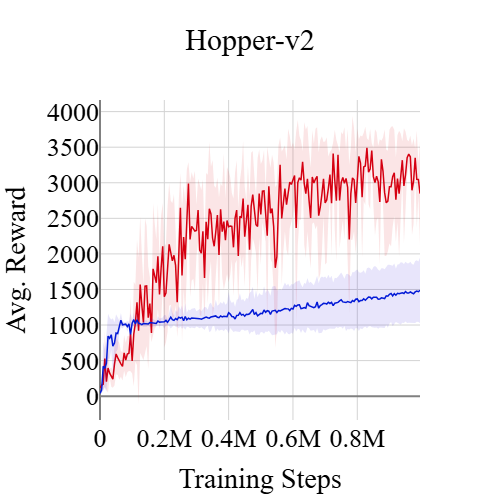}
        \label{fig:image3}
    } 
    \subfloat{
        \includegraphics[width=0.16\textwidth]{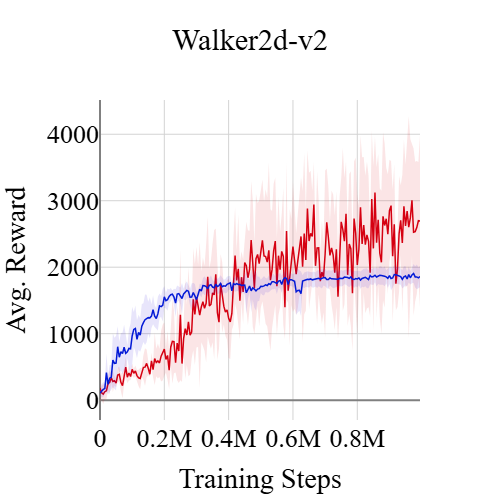}
        \label{fig:image4}
    }
    \subfloat{
        \includegraphics[width=0.16\textwidth]{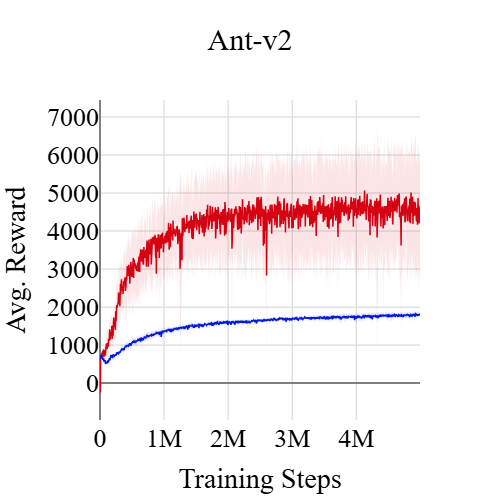}
        \label{fig:image5}
    }
    \subfloat{
        \includegraphics[width=0.16\textwidth]{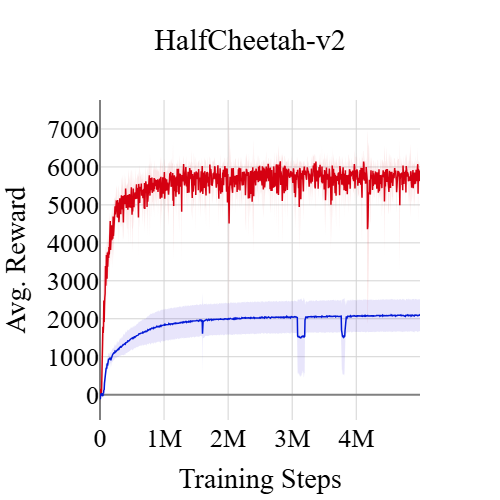}
        \label{fig:image6}
    }

    \subfloat{
        \includegraphics[width=0.16\textwidth]{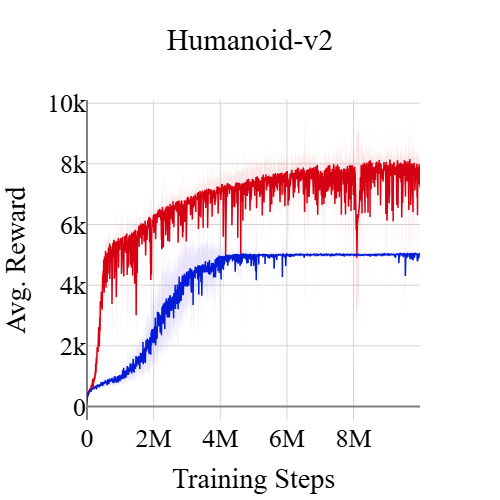}
        \label{fig:image4}
    }
    \subfloat{
        \includegraphics[width=0.16\textwidth]{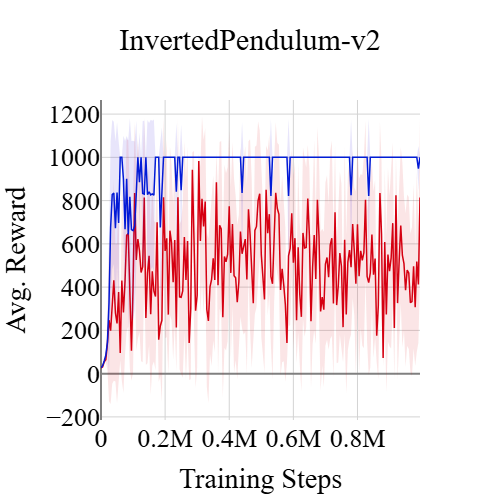}
        \label{fig:image5}
    }
    \subfloat{
        \includegraphics[width=0.16\textwidth]{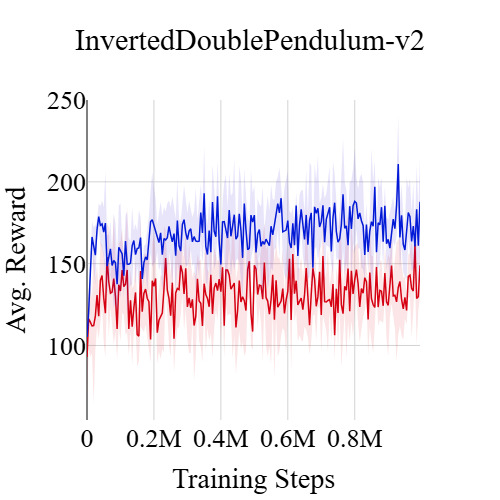}
        \label{fig:image6}
    }
    \subfloat{
        \includegraphics[width=0.16\textwidth]{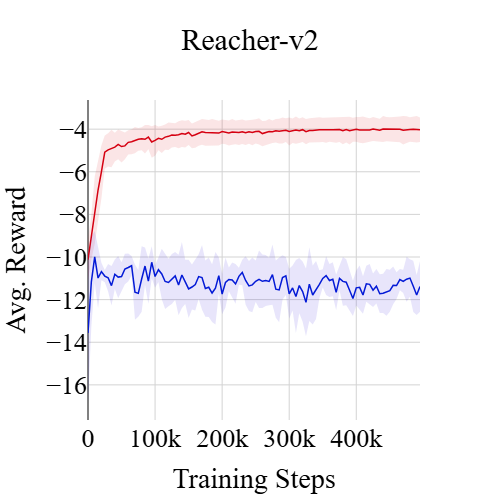}
        \label{fig:image4}
    }
   \subfloat{
    \includegraphics[width=0.16\textwidth]{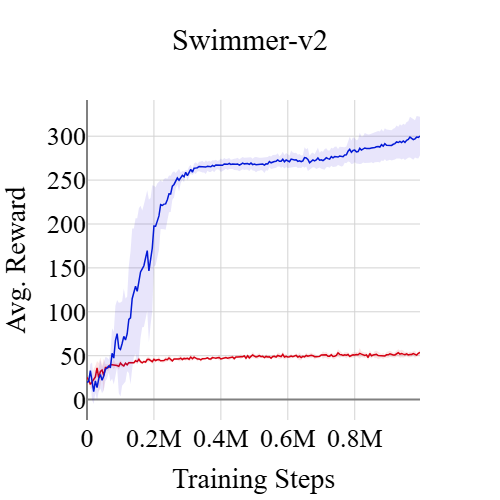}
    \label{fig:image4}
    }
    \subfloat{
    \includegraphics[width=0.16\textwidth]{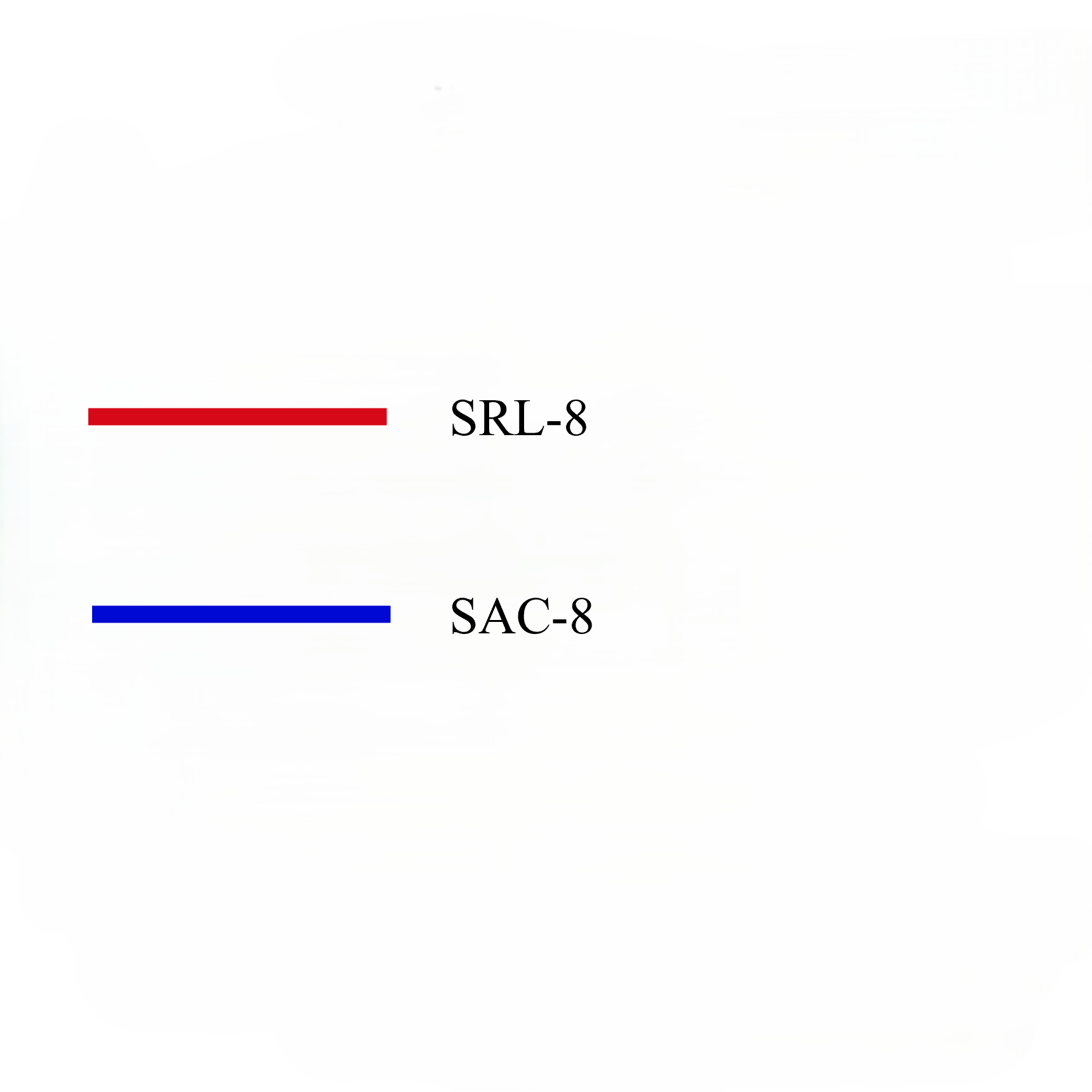}
    \label{fig:image4}
    }
    \caption{Learning curve of SRL-$8$ and SAC-$8$.}
    \label{fig:lc8}
\end{figure}

\begin{figure}[htbp]
    \centering
    \subfloat{
        \includegraphics[width=0.16\textwidth]{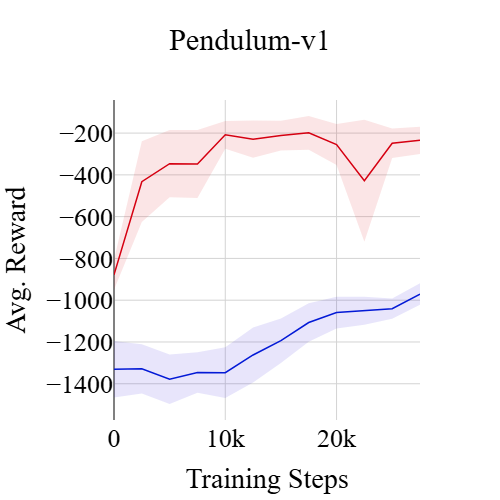}
        \label{fig:image1}
    }
    \subfloat{
        \includegraphics[width=0.16\textwidth]{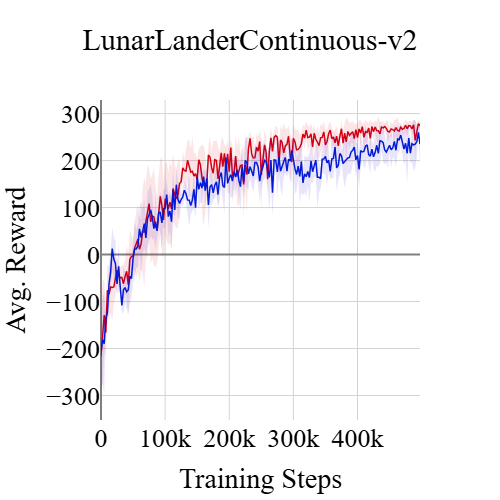}
        \label{fig:image2}
    }
    \subfloat{
        \includegraphics[width=0.16\textwidth]{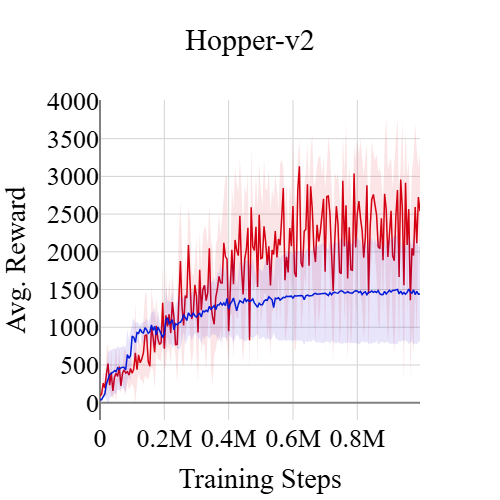}
        \label{fig:image3}
    } 
    \subfloat{
        \includegraphics[width=0.16\textwidth]{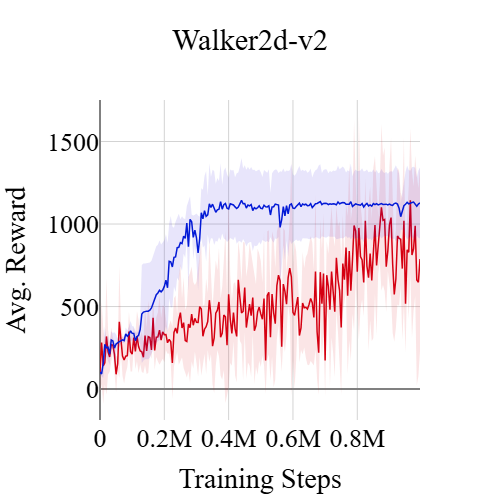}
        \label{fig:image4}
    }
    \subfloat{
        \includegraphics[width=0.16\textwidth]{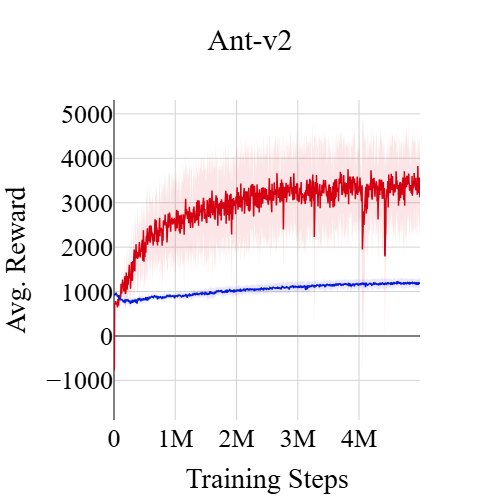}
        \label{fig:image5}
    }
    \subfloat{
        \includegraphics[width=0.16\textwidth]{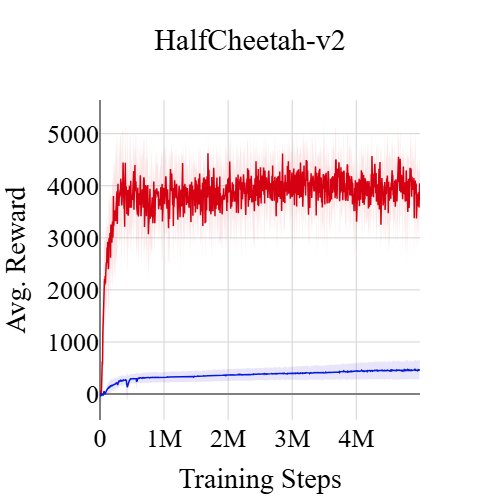}
        \label{fig:image6}
    }

    \subfloat{
        \includegraphics[width=0.16\textwidth]{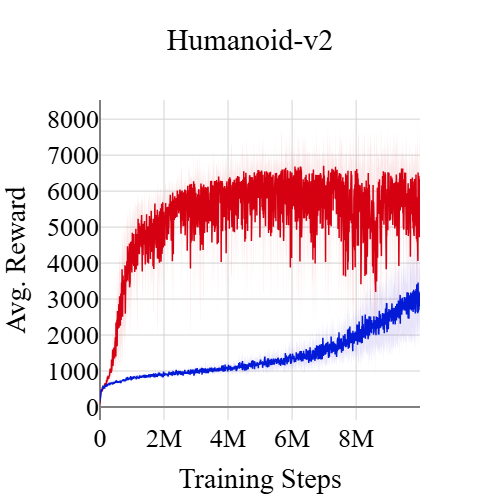}
        \label{fig:image4}
    }
    \subfloat{
        \includegraphics[width=0.16\textwidth]{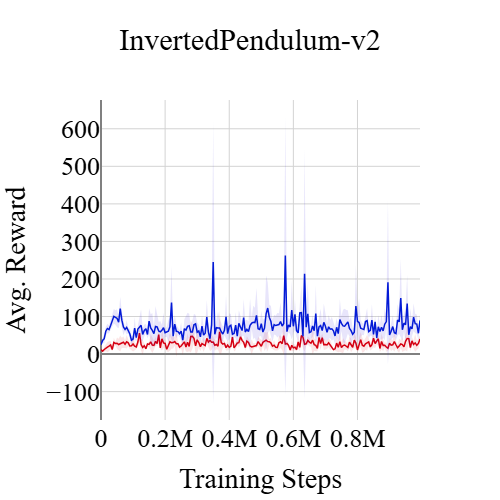}
        \label{fig:image5}
    }
    \subfloat{
        \includegraphics[width=0.16\textwidth]{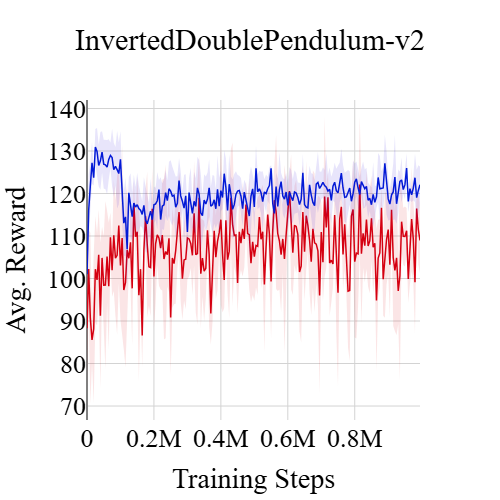}
        \label{fig:image6}
    }
    \subfloat{
        \includegraphics[width=0.16\textwidth]{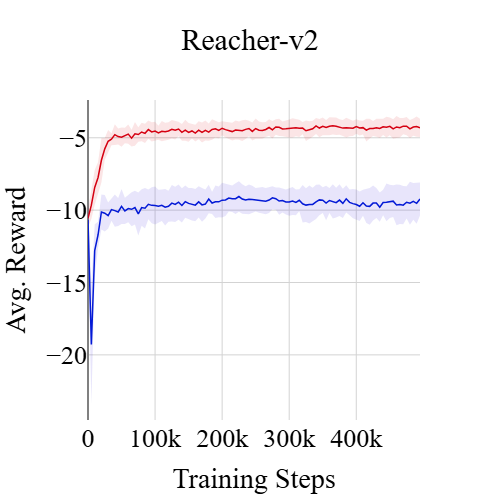}
        \label{fig:image4}
    }
   \subfloat{
    \includegraphics[width=0.16\textwidth]{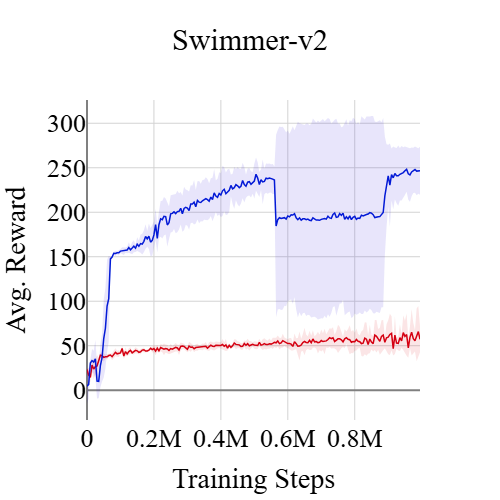}
    \label{fig:image4}
    }
    \subfloat{
    \includegraphics[width=0.16\textwidth]{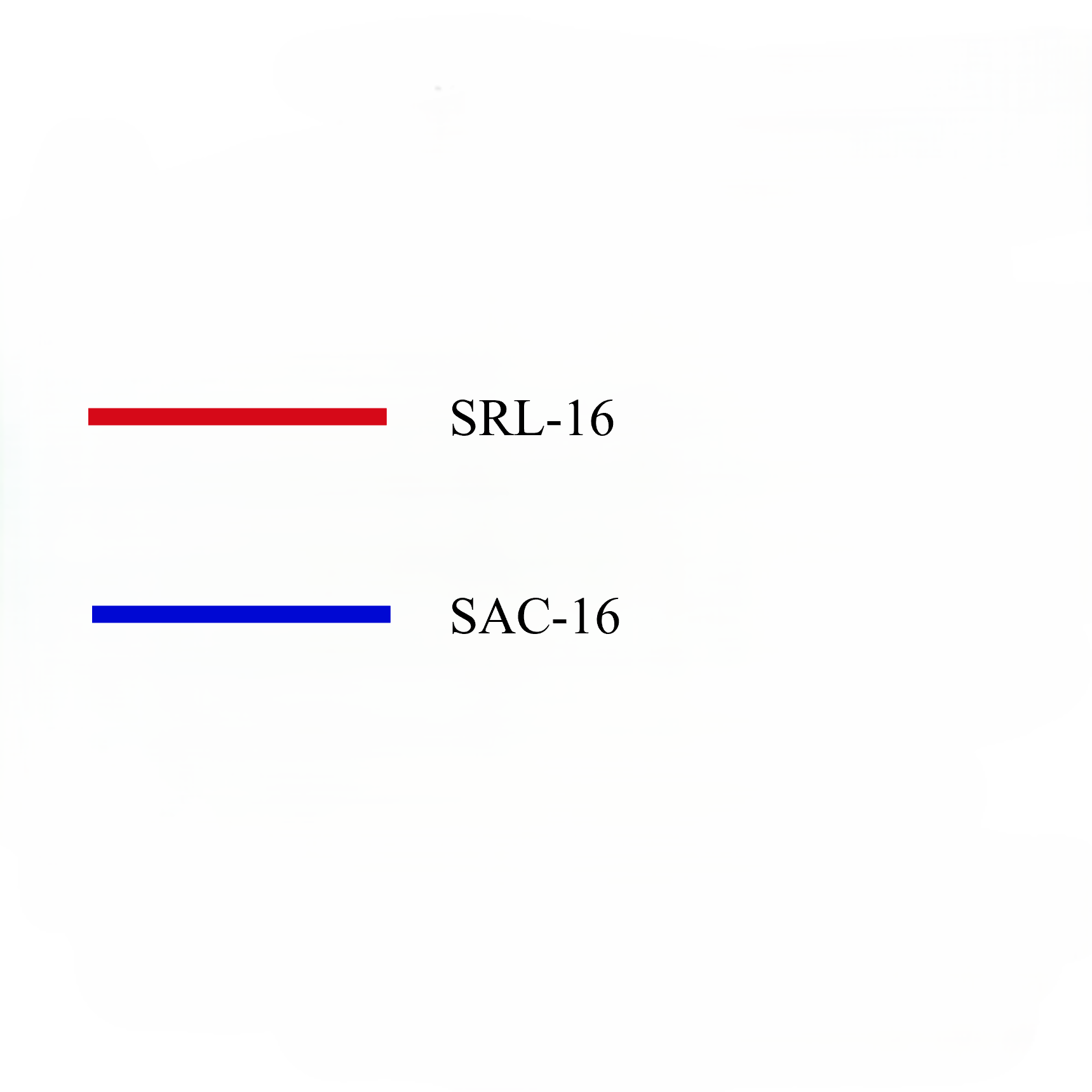}
    \label{fig:image4}
    }
    \caption{Learning curve of SRL-$16$ and SAC-$16$.}
    \label{fig:lc16}
\end{figure}

\newpage

\subsection{Randomized frame-skipping}\label{A12}
As shown, SAC trained on a constant timestep cannot adapt to different timesteps. For a fairer comparison, we also present results on randomized frame-skipping implemented on SAC during training. 

\begin{figure}[htbp]
    \centering
    \subfloat{
        \includegraphics[width=0.3\textwidth]{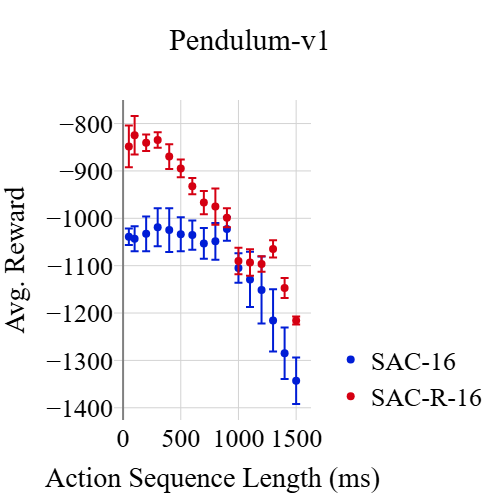}
        \label{fig:image1}
    }
    \subfloat{
        \includegraphics[width=0.3\textwidth]{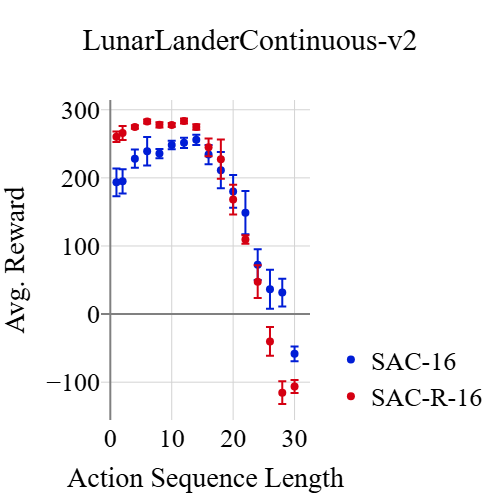}
        \label{fig:image2}
    }
    \subfloat{
        \includegraphics[width=0.3\textwidth]{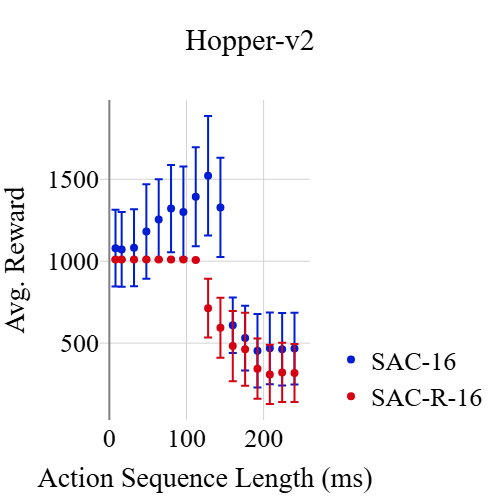}
        \label{fig:image2}
    }
    
    \caption{Performance of SAC and randomized SAC (SAC-R).}
    \label{fig:randomizedSAC}
\end{figure}

Figure \ref{fig:randomizedSAC} compares the performance of randomized SAC (SAC-R) to SAC at $J=16$. Surprisingly, we find that randomized frame-skipping during training improves the performance at shorter action sequence lengths (ASL) for simple environments like pendulum and lunar lander. However, for Hopper, SAC-R performs worse than SAC. This is most probably due to the stochasticity introduced due to the randomized frame-skipping. Even with randomized frame-skipping, SAC fails to achieve performance similar to SRL on simple environments, thus further reinforcing the results presented in this paper.

\subsection{Results for TempoRL Algorithm}\label{A13}

To further provide provide context for the contribution of this work in comparison to previous work, we provide further comparison to TempoRL \citep{Biedenkapp2021TempoRLLW} and also discuss performance compared to recent work on observational dropout.

\begin{table}[h]
    \centering
    \begin{tabular}{|l|c|c|c|}
    \hline
        \textbf{Environment} & \textbf{Avg. Reward} & \textbf{Avg. Sequence Length} & \textbf{Max sequence Length}\\
    \hline
    Pendulum & $-149.38 \pm 31.26$ & 71.74ms & 6\\
    Hopper & $2607.86 \pm 342.23$ & 22.4ms & 9 \\
    Walker2d & $4581.69 \pm 561.95$ & 25.54ms & 7 \\
    Ant & $3507.85 \pm 579.95$ & 62.66ms & 3 \\
    HalfCheetah & $6627.73 \pm 2500.77$ & 56.20ms & 3 \\
    Inv Pendulum & $984.21 \pm 47.37$ &  73.92ms & 10 \\ 
    InvD Pendulum & $9352.61 \pm 2.2$ & 58.76ms & 5 \\
    \hline
    \end{tabular}
    \caption{Results of running TempoRL on Mujoco Tasks. All results are averaged over 10 seeds.}
    \label{tab:TempoRL}
\end{table}

TempoRL cannot be adapted to the FAS setting since after each action is picked, it further picks the duration for the amount of time the action will be performed. Yet, since it promotes action repetiton, it results in lower decision frequency and longer action sequence lengths than standard algorithms like TD3 and SAC.

Table \ref{tab:TempoRL} demonstrates the results of training TempoRL algorithm on some of the benchmarks presented in this paper. We did a quick hyperparameter search over the max sequence length parameter and pick the highest number over 3 that did not result in a significant drop in performance. We find that while TempoRL achieve optimal performance on environments with single dimensions like pendulums, it demonstrates significant drop in performance on environments with multiple dimensions like Ant and HalfCheetah. Furthermore, on all environments, it maintains a relatively short action sequence length and even though it is given the option of picking long action sequences, it rarely does so. This result further demonstrates the contribution of SRL at maintaining performance at really long sequence lengths in environments with high action dimensions.



\end{document}